\documentclass[10pt,a4paper,twoside,onecolumn,openright,nomarginpar]{book}

\usepackage[textheight=620pt,voffset=25pt]{geometry}

\usepackage{etex}
\reserveinserts{100}

\usepackage{layout}

\usepackage[margin=10pt,font=small,labelfont=bf]{caption}

\usepackage{chngcntr}
\counterwithout{footnote}{chapter}

\usepackage[all]{xy}

\usepackage{makeidx}  
\usepackage{times}
\usepackage{graphicx}
\usepackage{latexsym,footnote}
\usepackage{amsmath}
\DeclareFontFamily{U}{MnSymbolC}{}
\DeclareSymbolFont{MnSyC}{U}{MnSymbolC}{m}{n}
\DeclareFontShape{U}{MnSymbolC}{m}{n}{
    <-6>  MnSymbolC5
   <6-7>  MnSymbolC6
   <7-8>  MnSymbolC7
   <8-9>  MnSymbolC8
   <9-10> MnSymbolC9
  <10-12> MnSymbolC10
  <12->   MnSymbolC12%
}{}
\DeclareMathSymbol{\powerset}{\mathord}{MnSyC}{180}\usepackage{amsthm}
\usepackage{algorithm}
\usepackage[noend]{algpseudocode}
\algrenewcommand\algorithmicrequire{\textbf{Input:}}
\algrenewcommand\algorithmicensure{\textbf{Output:}}

\usepackage{mathnotation}
\usepackage{array}
\usepackage{dashrule}

\usepackage[T1]{fontenc} 

\usepackage[draft]{fixme}

\usepackage{subcaption}
\usepackage{multirow}
\usepackage{tabularx}
\usepackage{cuted}

\usepackage{amssymb}

\usepackage{color}															 
\usepackage[table]{xcolor}
\usepackage{booktabs}
\usepackage{arydshln}
\usepackage{cancel}
\usepackage{rotating}
\usepackage{framed}
\usepackage{float}
\usepackage{ marvosym }

\usepackage[nottoc,notlot,notlof]{tocbibind}

\usepackage{xcolor}

\usepackage{url}
\usepackage{tikz}

\usepackage{amsfonts}
\usepackage{wasysym}

\usepackage{pdfpages}
\usepackage[nottoc]{tocbibind}
\usepackage{fancyhdr}
\usepackage{emptypage}
\usepackage{enumitem}
\usepackage{rotating}
\newcommand\tabrotate[1]{\rotatebox{90}{#1\hspace{\tabcolsep}}}

\usepackage{picinpar}

\definecolor{light-gray1}{gray}{0.95}

\fxsetup{
nomargin,inline,index,
theme=color
}

\newcounter{examplecounter}
\newenvironment{example}[1][\unskip]{
    \refstepcounter{examplecounter}%
  
	\vspace{7pt}
	\noindent\textbf{Example \arabic{chapter}.\arabic{examplecounter}\;\textbf{(#1)}}%
  \quad
}{

\vspace{7pt}
}
\numberwithin{examplecounter}{chapter}

\newcounter{remarkcounter}

\numberwithin{remarkcounter}{chapter}

\definecolor{darkgray}{rgb}{0.8,0.8,0.8}
\definecolor{lightgray}{rgb}{0.95,0.95,0.95}

\newcommand{\md}{\mathcal{D}}
\newcommand{\mc}{\mathcal{C}}

\newcommand{\mC}{{\bf{C}}}
\newcommand{\mD}{{\bf{D}}}

\newcommand{\ok}{{\mathit{ok}}}
\newcommand{\uu}[1]{\underline{\underline{#1}}}

{}
{}
{}
{}
{}

\usepackage{tocloft}
\begin{document}
\includepdf{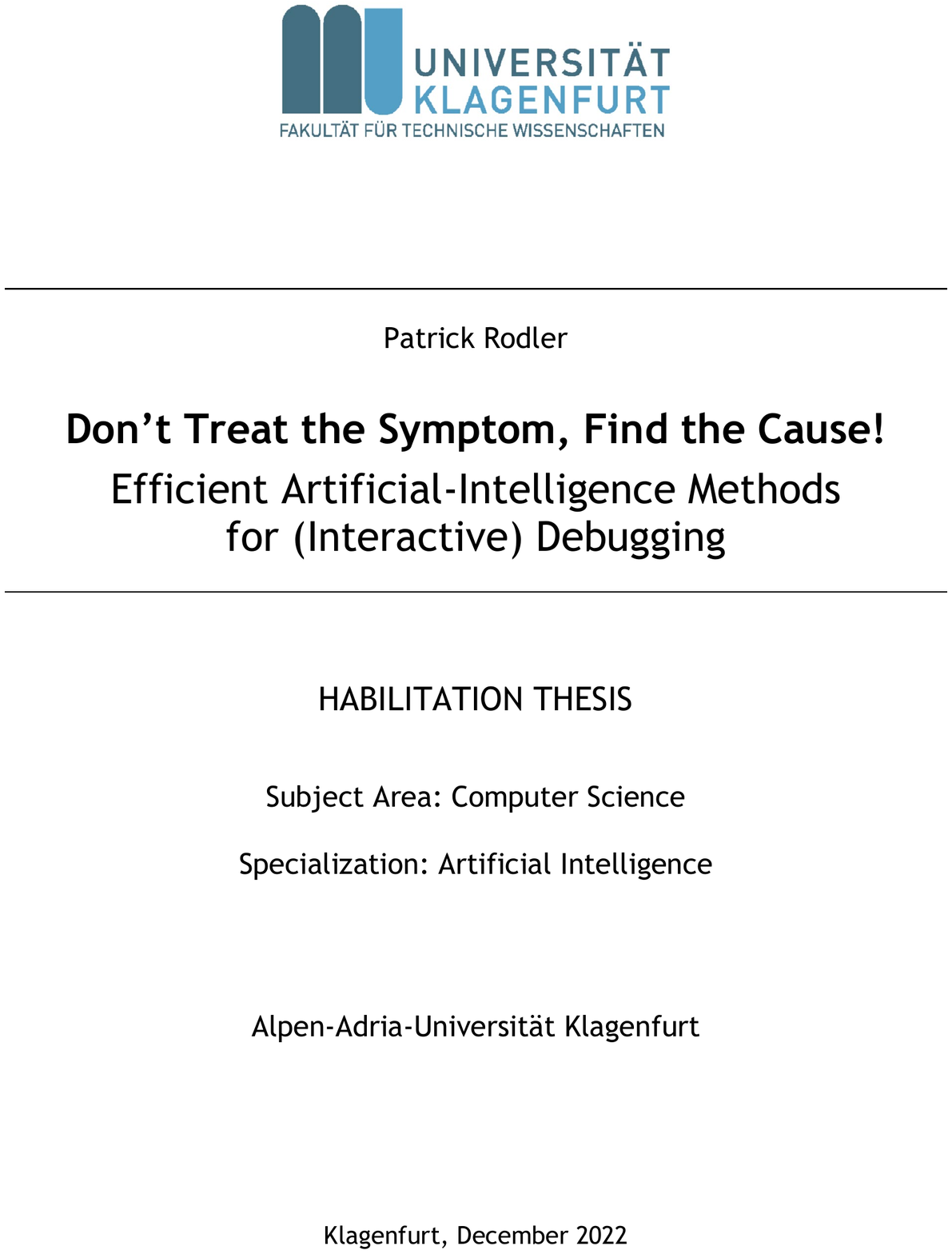}
\includepdf{empty_page.pdf}
\chapter*{Abstract}
In the modern world, we are permanently using, leveraging, interacting with, and relying upon systems of ever higher sophistication, ranging from our cars, recommender systems in e-commerce, and networks when we go online, to integrated circuits when using our PCs and smartphones, the power grid to ensure our energy supply,
security-critical software when accessing our bank accounts, and spreadsheets for financial planning and decision making. The complexity of these systems coupled with our high dependency on them implies both a non-negligible likelihood of system failures, and a high potential that such failures have significant negative effects on our everyday life. For that reason, it is a vital requirement to keep the harm of emerging failures to a minimum, which means minimizing the system downtime as well as the cost of system repair. This is where model-based diagnosis comes into play.

Model-based diagnosis is a principled, domain-independent approach that can be generally applied to troubleshoot systems of a wide variety of types, including all the ones mentioned above, and many more. It exploits and orchestrates i.a.\ techniques for knowledge representation, automated reasoning, heuristic problem solving, intelligent search, optimization, stochastics, statistics, decision making under uncertainty, machine learning, as well as calculus, combinatorics and set theory to detect, localize, and fix faults in abnormally behaving systems.    

In this thesis, we will give an introduction to the topic of model-based diagnosis, point out the major challenges in the field, and discuss a selection of approaches from our research addressing these issues. For instance, we will 
\begin{itemize}[noitemsep,topsep=0pt]
	\item present methods for the optimization of the time and memory performance of diagnosis systems,
	\item tackle the improvement of the usefulness and usability of debuggers,
	\item introduce efficient techniques for a semi-automatic debugging by interacting with a user or expert,
	\item reveal results of profound theoretical and empirical analyses,
	\item address how to conquer particularly challenging diagnostic use cases,
	\item make recommendations 
	as to the best use, configuration, and combination 
	of diagnostic algorithms,
	\item provide assistance for other researchers and practitioners by a survey and didactic treatises,
	\item critically call into question and scrutinize techniques and practices adopted in the field,
	\item describe our full-fledged debugging tool that incorporates more than a decade of research, and
	\item demonstrate how our methods can be effectively harnessed in important application domains such as 
	optimal scheduling, spreadsheet debugging, or the quality assurance of the Semantic Web.
\end{itemize}


\thispagestyle{empty}
\includepdf{empty_page.pdf}
{\pagestyle{empty}
	\tableofcontents
\pagestyle{empty}}
\thispagestyle{empty}
\addtocontents{toc}{\protect\thispagestyle{empty}}
\mainmatter
\begingroup
\let\clearpage\relax
\addcontentsline{toc}{chapter}{\listtablename}
\listoftables
\endgroup
\begingroup
\let\clearpage\relax
\vspace{-0pt}
\begingroup
\renewcommand*{\addvspace}[1]{}
\addcontentsline{toc}{chapter}{\listfigurename}
\listoffigures
\endgroup
\endgroup
\newpage\null\thispagestyle{empty}\newpage

\chapter{Introduction and Preliminaries}
\label{chap:intro}
We first provide a short and example-driven introduction to model-based diagnosis, the overarching topic of this thesis. Then, we outline the main ideas behind sequential diagnosis, a research area dedicated to efficient interactive diagnosis. Next, we describe the principal diagnostic tasks of interest, shed light on application areas, and discuss links to related research domains. Finally, we expound the structure, modules and way of functioning of a generic (interactive) diagnosis system, based on which we explain the main research challenges and goals in the field tackled by our works described in later chapters. 

Please note that we intentionally keep the discussion of the preliminaries on a higher level of abstraction and abstain from a very formal and detailed treatment in this chapter, whose main aim is to give the reader an intuition and overview of the topic. More in-depth accounts of technicalities can be found in the papers included in this thesis (cf. Appendix~\ref{apx:papers_incl_in_thesis}).  

\section{Model-Based Diagnosis}
\label{sec:MBD}
\textbf{Model-based diagnosis} \cite{reiter_theory_1987,de_kleer_diagnosing_1987} is a popular, well-understood, domain-independent and principled paradigm for troubleshooting malfunctioning systems of a wide variety of types, such as physical devices, software, knowledge bases, or vehicles. 
The theory of model-based diagnosis 
\cite{reiter_theory_1987} assumes a \emph{system} (e.g., a circuit)
that is composed of a set of \emph{components} (e.g., gates), and a 
\emph{system description} (e.g., a logical knowledge base characterizing the circuit). The latter is expressed in a formal (logical) language and can be used to derive the expected behavior of the system by means of automated deduction systems. If the predicted system behavior, under the assumption that all components are functioning normally, 
is not in line with \emph{observations} made or \emph{measurements} conducted about the system,
the goal is to locate the abnormal system components that are responsible for this discrepancy.   

\begin{figure}
	\centering
	\includegraphics[trim={0cm 0 0 0},clip,width=0.9\linewidth]{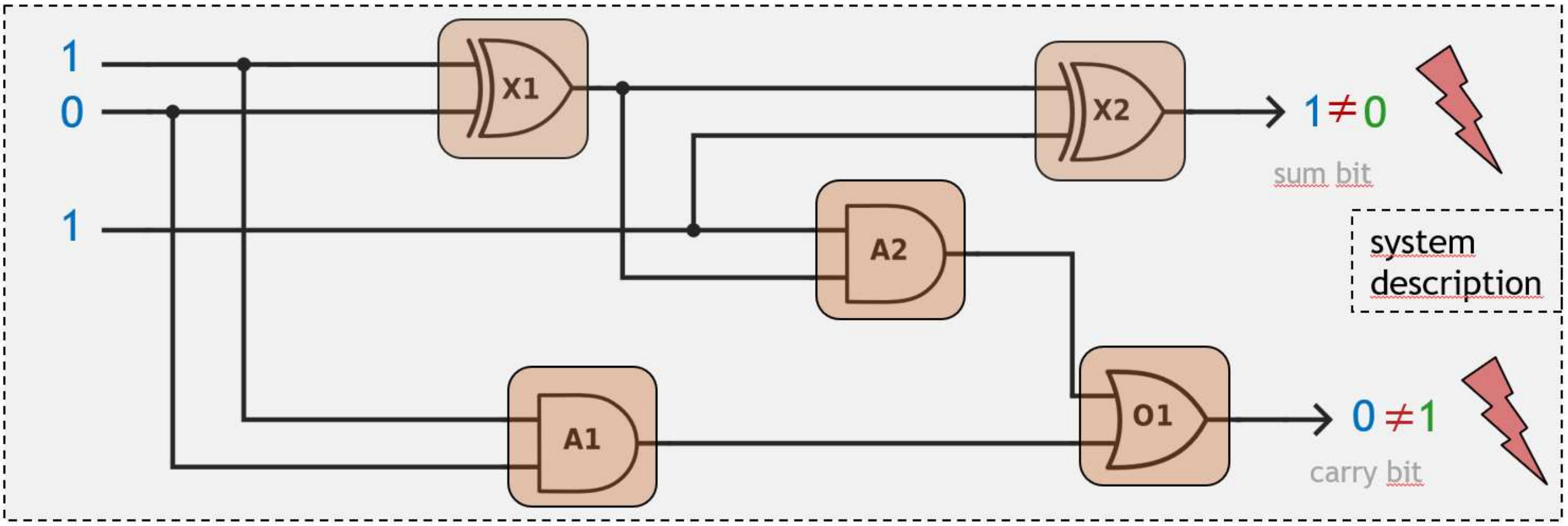}
	\caption[Full adder example: Components, system description, observations, predictions]{Full adder example: Components (\emph{orange boxes}), observations (\emph{blue numbers}), predictions under the assumption of normally behaving components (\emph{green numbers}), and consistency check between observed and predicted values (\emph{red inequality signs}; \emph{red flashes} indicate inconsistency). The system description is a formal characterization of the depicted circuit, e.g., a set of logical sentences expressing the (normal) behavior of the components, their interrelations, as well as other relevant domain knowledge.}
	\label{fig:ex:circuit_MBD}
\end{figure}

\begin{example}[System description, components, observations, predictions]\label{ex:circuit_MBD}
	Consider the faulty full adder (cf.\ \cite{reiter_theory_1987}) depicted in Fig.~\ref{fig:ex:circuit_MBD}. We have five system components, comprising two xor-gates ($X_1,X_2$), two and-gates ($A_1,A_2$), and one or-gate ($O_1$). Moreover, we have observations of the circuit's inputs ($1$,$0$,$1$) and outputs (sum bit equal to $1$ and carry bit equal to $0$). No measurements are given. The system description could, e.g., be formulated as a logical knowledge base, specifying the (normal) behavior of all gates, which gates are connected by wires, as well as relevant general domain knowledge. E.g., for the xor-gate $X_1$, we would have the logical axiom $\ok(X_1) \to \mathit{out}(X_1) = \mathit{xor}(\mathit{in}_1(X_1),\mathit{in}_2(X_1))$ where $\mathit{out}(G)$ describes the value at the output terminal of gate $G$, $\mathit{in}_1(G), \mathit{in}_2(G)$ denote the two input values of gate $G$, $\mathit{xor}$ is the standard xor-function, and $\ok$ is a distinguished predicate used to state that a component is functioning normally. Further, the connection between, e.g., the xor-gates $X_1$ and $X_2$ is specified by the axiom $\mathit{out}(X_1) = \mathit{in}_1(X_2)$. An example of a general domain axiom is $\lnot(1 = 0)$, which expresses that the constants $1$ and $0$ are different, i.e., if we derive that the value at some gate terminal is both $1$ and $0$, then we obtain a contradiction. Now, using a theorem prover to predict the circuit's outputs given its inputs, while assuming all five gates to be normal, i.e., $\{\ok(X_1),\ok(X_2),\ok(A_1),\ok(A_2),\ok(O_1)\}$, we obtain a contradiction. In particular, we deduce that the sum bit must be $0$, and the carry bit must be $1$, which however is inconsistent with the given observations of the two outputs. As a conclusion, we know that some component must be abnormal, i.e., the $\ok(G)$ assumption for some gate(s) $G$ must be retracted to restore consistency between predictions and observations. \qed      
\end{example}
\noindent\textbf{Different Interpretations of the Diagnosis Problem.} In general, the are two main interpretations of the system description on the one hand, and the observations and measurements on the other. First, as illustrated in Example~\ref{ex:circuit_MBD}, the system description might constitute the correct specification of a system (e.g., a physical device) in which case the observations and measurements represent evidence about the actual system (the real device) implementing the specification. We term this interpretation of system model vs.\ evidence as \emph{correct vs.\ actual}.
Second, the system description might correspond to a model of a faulty artifact (e.g., a buggy program) in which case the observations and measurements are specifications of the intended system behavior (e.g., giving evidence what the correct program should do), and the ``normal'' behavior of components (e.g., lines of code) is the description of the components just as they are given in the faulty artifact. We refer to this interpretation as \emph{actual vs.\ intended}.\vspace{4pt}

\noindent\textbf{Diagnosis Problem Instance and Diagnoses.} Given a \emph{diagnosis problem instance (DPI)}---consisting of the system description, the system components, the observations and the measurements---a \emph{(minimal) diagnosis} is a (minimal\footnote{Whenever we speak of \emph{minimality}, we mean minimality wrt.\ set inclusion.}) set of components that, when assumed abnormal, while all other components are assumed normal, leads to consistency between the system description 
(predicted behavior) and the observations and measurements (real/intended behavior). 
Simply put, a diagnosis is an explanation for the observed or measured system misbehavior, and a minimal diagnosis is an explanation that assumes a minimal set of components to be abnormal. 
In practice, for reason of computational efficiency and obeying the principle of parsimony \cite{reiter_theory_1987}, the focus of diagnosis research and tools is often restricted to minimal diagnoses. 
The set of minimal diagnoses is representative of all diagnoses given the common \emph{weak fault model} \cite{de_kleer_characterizing_1992} setting where the system description characterizes only the normal, and no abnormal, behavior of components. This is also the diagnostic setting we consider throughout this work. 

\noindent\textbf{Conflicts and Hitting-Set Property.} Computing (minimal) diagnoses requires checking the consistency of different component (ab)normality assumptions 
with the observations and measurements. Basically, 
the relevant search space spans all subsets of the set of system components, which makes an exhaustive exploration infeasible in all but the simplest cases. However, a more focused generation of diagnoses is enabled by the notion of a conflict. Intuitively, when the windshield wipers of our car do not work, the attention of the diagnostic analysis can concentrate on a small subset of the car's components, e.g., the motor of the wipers, or the connecting cables, but not, say, on parts of the car's engine or the braking system, because they do not influence the observed behavior of the car \cite[Chap.~10]{van_harmelen_handbook_2008}. That is, observations are usually not only inconsistent with the system description under the assumption that \emph{all} components are normal, but already when an (often relatively) small subset of the components is assumed fault-free. In more technical terms, a \emph{(minimal) conflict} is a (minimal) set of system components that, when assumed normal, leads to an inconsistency between the system description and the observations and measurements. So, at least one component in each (minimal) conflict must be assumed abnormal in order to explain the system misbehavior. This directly leads us to the hitting-set relationship between conflicts and diagnoses \cite{reiter_theory_1987}: A (minimal) diagnosis is a (minimal) hitting set of all minimal conflict sets. Given a collection of sets $\mathbf{S}$, $X$ is a \emph{(minimal) hitting set} of $\mathbf{S}$ iff ($X$ is a minimal set such that) 
$X$ includes only elements occurring in sets of $\mathbf{S}$ and $X$ has a non-empty intersection with each set in $\mathbf{S}$.


%

\begin{figure}
	\centering
	\includegraphics[trim={0cm 0 0 0},clip,width=0.9\linewidth]{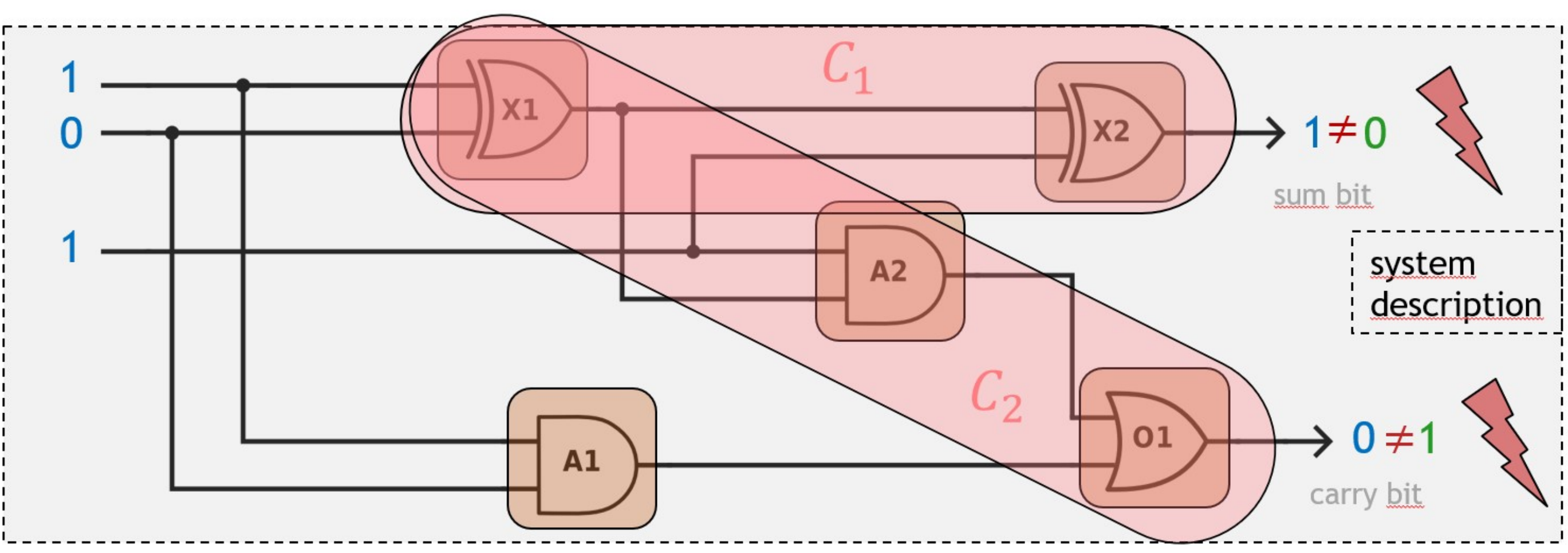}
	\caption[Full adder example: Minimal conflicts]{Full adder example: Minimal conflicts.}
	\label{fig:ex:circuit_conflicts}
\end{figure}

\begin{figure}
	\centering
	\includegraphics[trim={0cm 0 0 0},clip,width=0.9\linewidth]{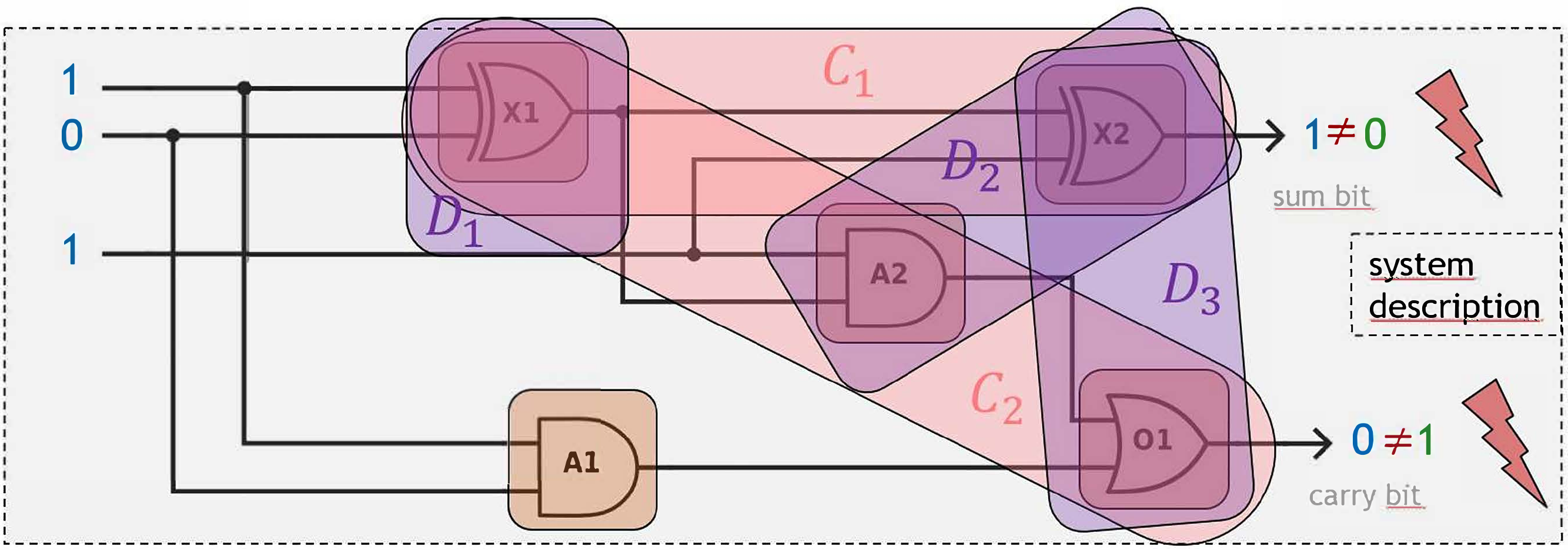}
	\caption[Full adder example: Minimal diagnoses]{Full adder example: Minimal conflicts and minimal diagnoses.}
	\label{fig:ex:circuit_conflicts_diags}
\end{figure}

\begin{example}[Conflicts and diagnoses]\label{ex:circuit_conflicts+diagnoses}
	For our full-adder problem in Fig.~\ref{fig:ex:circuit_MBD}, we find two minimal conflicts, i.e., $\mC = \{\mc_1,\mc_2\} = \{\tuple{X_1,X_2},\tuple{X_1,A_2,O_1}\}$, as illustrated in Fig.~\ref{fig:ex:circuit_conflicts}.\footnote{Throughout this work, we denote conflicts by angle brackets and diagnoses by squared brackets.} E.g., $\mc_1$ is a conflict because, assuming the two xor-gates $X_1,X_2$ to be functioning normally, \emph{(1)}~we can derive from the two inputs $(\mathit{in}_1(X_1),\mathit{in}_2(X_1)) = (1,0)$ of $X_1$ that the output $\mathit{out}(X_1)$ of $X_1$ must be equal to $1$, and \emph{(2)}~from $\mathit{out}(X_1) = \mathit{in}_1(X_2) = 1$ as well as $\mathit{in}_2(X_2) = 1$, we can deduce that $\mathit{out}(X_2) = 0$, which means a contradiction to the observed value $\mathit{out}(X_2) = 1$. The minimality of the conflicts follows from the fact that, once we omit one of the normality assumptions for $X_1$ or $X_2$, we cannot derive $\mathit{out}(X_2) = 0$ anymore. An analogous argumentation is applicable for the second conflict $\mc_2$.
	
	By computing all minimal hitting sets for $\mC$, we can determine all minimal diagnoses $\mD = \{\md_1,\md_2$, $\md_3\} = \{[X_1],[A_2,X_2],[O_1,X_2]\}$,
	as shown by Fig.~\ref{fig:ex:circuit_conflicts_diags}. E.g., diagnosis $\md_1$ corresponds to the assumption $\{\lnot\ok(X_1),\ok(X_2),\ok(A_1),\ok(A_2),\ok(O_1)\}$. This assumption explains the faulty behavior of the full-adder because, when assuming $X_1$ to be abnormal (i.e., $\lnot \ok(X_1)$), which is equivalent to dropping the restriction $\mathit{out}(X_1) = \mathit{xor}(\mathit{in}_1(X_1),\mathit{in}_2(X_1))$ imposed on the behavior of $X_1$ (cf.\ Example~\ref{ex:circuit_MBD}), none of the two predictions $\mathit{out}(X_2) = 0, \mathit{out}(O_1) = 1$ which contradict the observations can be inferred. In other words, the system description under this assumption is consistent with the observations. This is exactly the definition of a diagnosis. Minimality of the diagnosis $\md_1$ can be directly seen by observing that the only proper subset $[\,]$ (the empty diagnosis which assumes none of the components faulty) of $[X_1]$ is not a diagnosis (cf.\ the argumentation in Example~\ref{ex:circuit_MBD}).
	\qed  
\end{example}

\noindent\textbf{Classical Algorithms for Diagnosis and Conflict Computation.} 
As a consequence of the hitting-set property discussed above, conflicts can help to largely restrict the search space for diagnoses, and also permit the computation of diagnoses in a particular order of preference or plausibility. The two most common such notions of diagnostic preferability are minimum-cardinality (i.e., those diagnoses assuming the fewest faulty components are favored) and maximal probability (i.e., the most likely diagnoses based on, e.g., known component failure rates are deemed most plausible). Notably, it is often not necessary to generate \emph{all} (minimal) conflicts if the goal is to find only a set of the most-preferred minimal diagnoses. 

\begin{figure}
	\centering
	\includegraphics[trim={0cm 0 0 0},clip,width=0.6\linewidth]{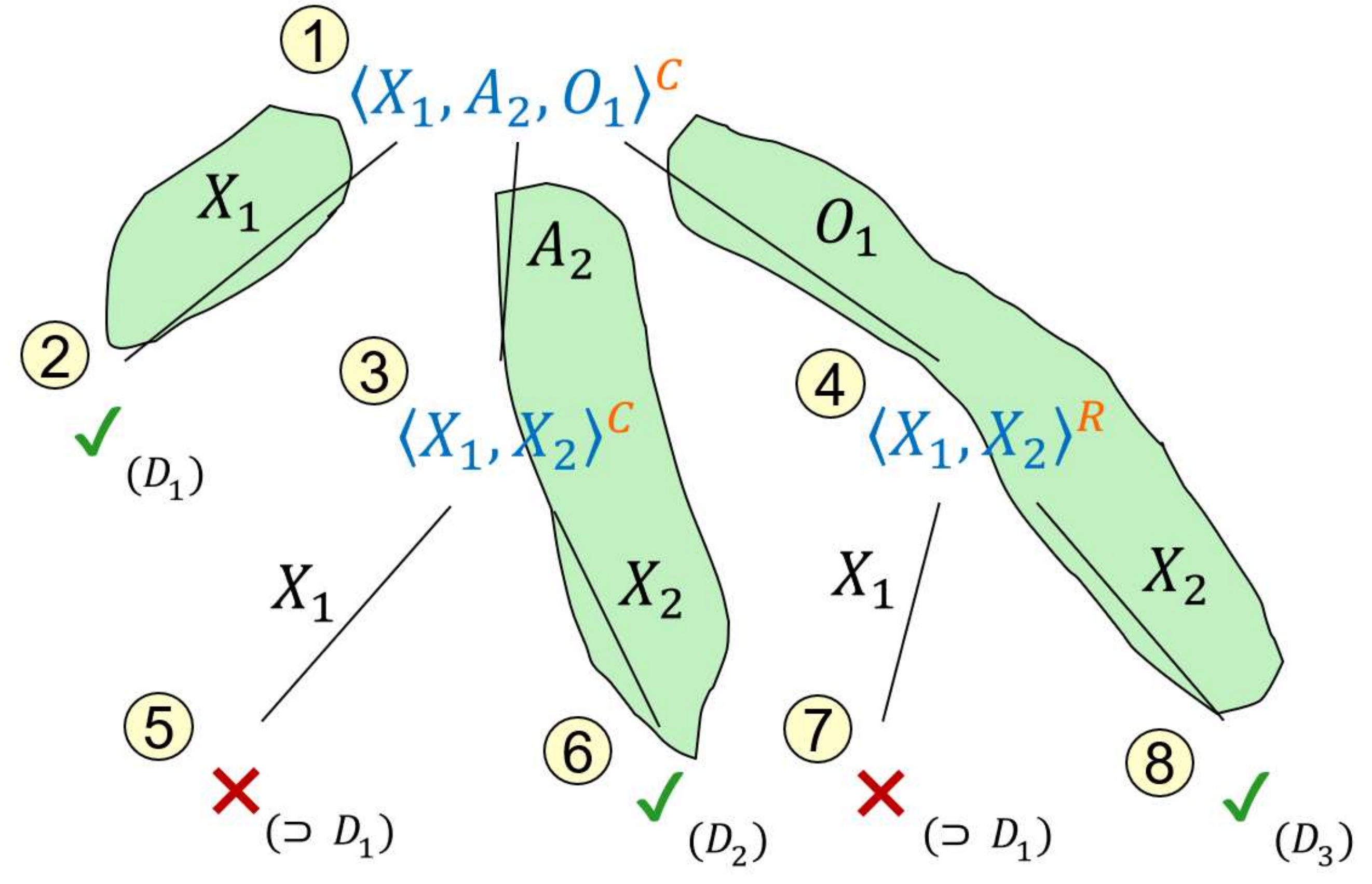}
	\caption[Full adder example: Diagnosis computation using HS-Tree]{Full adder example: Diagnosis computation using HS-Tree. Blue font...minimal conflicts; green shaded areas...paths that are minimal diagnoses; green font...node labels for minimal diagnoses; red font...node labels for non-minimal diagnoses; yellow circles...order of node processing (breadth-first); orange font...label indicating if conflict is freshly computed (C) or reused from stored conflicts (R).}
	\label{fig:ex:circuit_diagnosis_computation_using_HS-Tree}
\end{figure}

One prominent and commonly adopted algorithm for diagnosis computation based on conflicts is Reiter's HS-Tree \cite{reiter_theory_1987}. It constructs a node- and edge-labeled tree in breadth-first manner (cf.\ Fig.~\ref{fig:ex:circuit_diagnosis_computation_using_HS-Tree}). Each internal tree node $\mathsf{n}$ is labeled by a minimal conflict which has an empty intersection with $\mathit{path}(\mathsf{n})$, the set of edge labels along the path from the root to $\mathsf{n}$ (where $\mathit{path}(\mathsf{n}) = \emptyset$ for the root node $\mathsf{n}$). 
If there is no such conflict, $\mathit{path}(\mathsf{n})$ is a \emph{diagnosis} (as it must already hit all minimal conflicts) and is labeled by 
\begin{enumerate}[label=\emph{(\alph*)},noitemsep,topsep=0pt]
	\item $\times_{(\supset \md_i)}$
	if there is some already found minimal diagnosis $\md_i$ with $\md_i \subset \mathit{path}(\mathsf{n})$ (i.e., $\mathit{path}(\mathsf{n})$ is a \emph{non-minimal diagnosis} and node $\mathsf{n}$ is closed, i.e., becomes a leaf node), or
	\item $\checkmark_{(\md_i)}$ else, where the index $(\md_i)$ indicates that $\mathit{path}(\mathsf{n})$ is the $i$-th found minimal diagnosis denoted 
	by $\md_i$ (i.e., $\mathit{path}(\mathsf{n})$ is a \emph{minimal diagnosis}\footnote{Observe that $\mathit{path}(\mathsf{n})$ must indeed be a \emph{minimal} diagnosis in this case since the check in (a) was negative and due to the breadth-first tree construction, which guarantees that all nodes $\mathsf{n}'$ with $\mathit{path}(\mathsf{n}') \subset \mathit{path}(\mathsf{n})$ must already have been processed (and all diagnoses of lower cardinality must already have been found) before $\mathsf{n}$ is labeled.} 
	and node $\mathsf{n}$ is closed, i.e., becomes a leaf node).
\end{enumerate}
Each node labeled by a minimal conflict $\mc = \{c_1,\dots,c_k\}$ has $k$ outgoing edges, each labeled by a different element $c_i$ from $\mc$ and pointing to a newly generated node. Before labeling a node $\mathsf{n}$, the algorithm tests if there is another node $\mathsf{n}'$ with $\mathit{path}(\mathsf{n}') = \mathit{path}(\mathsf{n})$, and if so, $\mathsf{n}$ is labeled by $\times$, i.e., node $\mathsf{n}$ is closed and becomes a leaf node (because in terms of its path it is a duplicate and hence can be discarded).
Finally, when the tree $T$ is completed, i.e., there are no more unlabeled nodes, the set of minimal diagnoses is exactly given by $\mathit{Sol}_{\mathit{HS}} := \{\mathit{path}(\mathsf{n})\,|\,\mathsf{n} \text{ is labeled by }\checkmark\}$.

In fact, the execution of HS-Tree can be stopped at any point in time; in this case, $\mathit{Sol}_{\mathit{HS}}$ will comprise a set of most preferred diagnoses (those with minimal cardinality). Since conflict computation, taking into account the involved theorem prover calls, is usually the most expensive operation in the HS-Tree algorithm, all minimal conflicts computed throughout the tree construction process are stored, and before a conflict is freshly computed for a node, the collection of stored conflicts is scanned for a suitable conflict with an empty intersection with the current node's path.\footnote{Please refer to, e.g., \cite{rodler_dynamichs_2022} for a more elaborate discussion and a precise algorithmic description of HS-Tree.}

Literature offers a variety of algorithms for conflict computation (cf.\ e.g., \cite{rodler_formal_2022}). One of the most commonly used in model-based diagnosis is the seminal QuickXplain algorithm \cite{junker_quickxplain_2004,rodler_formal_2022} (cf.\ Example~\ref{ex:conflict_computation} and Sec.~\ref{sec:qx_proof}). It performs a recursive divide-and-conquer analysis of the set of system components 
and requires per computed minimal conflict a number of calls to a consistency checker that is linear in the number of system components. Hence, HS-Tree requires $O(|\mathit{Sol}_{\mathit{HS}}| + |\textsc{comps}|*n_{C})$ consistency checks, one for each diagnosis $\md_i$ ($\checkmark$ label) 
to verify that it hits every conflict,
plus $O(|\textsc{comps}|)$ calls per computed minimal conflict, where $\textsc{comps}$ denotes the set of system components and $n_C$ the number of computed minimal conflicts throughout the execution of HS-Tree. Note, no theorem prover calls are necessary to determine non-minimal diagnoses or duplicate paths ($\times$ labels), cf.\ \cite{rodler_dynamichs_2022}.

\begin{example}[Diagnosis computation using HS-Tree]\label{ex:diagnosis_computation}
Reconsider our full-adder example from Fig.~\ref{fig:ex:circuit_MBD}. The tree built by the HS-Tree algorithm for this problem is displayed by Fig.~\ref{fig:ex:circuit_diagnosis_computation_using_HS-Tree}. 
The set of computed diagnoses is indeed equal to all minimal diagnoses (cf.\ the set $\mD$ discussed in Example~\ref{ex:circuit_conflicts+diagnoses}). 
The algorithm requires two conflict computations (eight plus five consistency checks, cf.\ Example~\ref{ex:conflict_computation}) and three additional consistency checks (to prove that the branches marked with $\checkmark$ are diagnoses). For more complex systems with a large set of components, this conflict-based computation can make a huge difference as compared to a brute-force exploration of the diagnosis search space, as the size and number of minimal conflicts is often relatively small.
Note, to compute only one preferred diagnosis, say, a minimum-cardinality diagnosis assuming the fewest abnormal components (in this case $[X_1]$), the computation of a single minimal conflict would suffice in this particular example.\qed 
\end{example}

\begin{example}[Conflict computation using QuickXplain]\label{ex:conflict_computation}
The idea underlying conflict computation by means of QuickXplain can be illustrated as follows. Consider, e.g., the node labeled at time point $\textcircled{\footnotesize 3}$ in HS-Tree (cf.\ Fig.~\ref{fig:ex:circuit_diagnosis_computation_using_HS-Tree}). To determine the minimal conflict $\langle X_1,X_2 \rangle$ to label this node, QuickXplain will execute the Steps~{\small (1)}--{\small (5)} shown below. Each step involves one consistency check using a suitable (sound and complete) theorem prover. First, note that the goal when labeling the node with the number $\textcircled{\footnotesize 3}$ in the HS-Tree is finding a minimal conflict that has an empty intersection with the node's path $\{A_2\}$. In other words, we need to determine a minimal conflict that is a subset of the set $S$ including all components minus the node's path, i.e., $S = \{A_1,O_1,X_1,X_2\} = \{A_1,A_2,O_1,X_1,X_2\} \setminus \{A_2\}$. So, in Step~{\small (1)}, QuickXplain checks whether a conflict is contained in $S$ by testing whether the assumption that all the four components in $S$ are normal is consistent with the observations (below we highlight the components assumed normal and taking part in the consistency check by a single underlining). The result is negative (i.e., ``inconsistent'', shown by a $\textcircled{\scriptsize -}$) which means that there is at least one (minimal) conflict among the subsets of $S$. Let us refer to the minimal conflict that will be found by QuickXplain by $\mc$. In Step~{\small (2)}, the algorithm partitions $S$ into two equal-sized subsets and tests whether the left subset, $\{A_1,O_1\}$, is already a conflict. Since the normality assumption of these two components does not contradict the observations (i.e., we obtain ``consistent'', signified by the $\textcircled{\scriptsize +}$), it can be inferred that some component from $\{X_1,X_2\}$ must be in $\mc$. Hence, QuickXplain starts to analyze the right subset $\{X_1,X_2\}$ and recursively splits this set in two equally-sized subsets, $\{X_1\}$ and $\{X_2\}$, and adds the left set $\{X_1\}$ to the normally assumed components for the next consistency check performed in Step~{\small (3)}. We find that this check is positive ($\textcircled{\scriptsize +}$), which implies that $X_2$ is in $\mc$ (denoted by a double underlining in the illustration below). Intuitively, $X_2$ must be in $\mc$ as it is known from Step~{\small (1)} that $ \{A_1,O_1,X_1,X_2\}$ contains a conflict, and from Step~{\small (3)} that, once $X_2$ is removed, the set does not comprise any conflicts. In Step~{\small (4)}, a consistency test for (the normality assumption of) $\{A_1,O_1,X_2\}$ is executed to figure out whether the left subset, $\{X_1\}$, is not needed to build a conflict given the knowledge that the right subset, $\{X_2\}$, is in the conflict. However, we observe that consistency is given ($\textcircled{\scriptsize +}$), which entails that $X_1$ must be in $\mc$ as well (due to an analogue argumentation as given for $X_2$). Now, as it has been derived that both $X_1$ and $X_2$ are in $\mc$, the algorithm tests in Step~{\small (5)} if the left subset of $S$, i.e., $\{A_1,O_1\}$, is still relevant for the conflict in the light of this knowledge. Since the consistency check returns negatively ($\textcircled{\scriptsize -}$), QuickXplain deduces that both components $A_1,O_1$ are irrelevant to $\mc$ and returns the minimal conflict $\mc = \langle X_1,X_2\rangle$ as a final result.

We observe that the computation of
$\mc$ takes five consistency checks. Likewise, we find that the computation of the conflict $\mc' := \langle X_1,A_2,O_1\rangle$ at node $\textcircled{\footnotesize 1}$ in Fig.~\ref{fig:ex:circuit_diagnosis_computation_using_HS-Tree}, given the list of components $[A_1,A_2,O_1,X_1,X_2]$, requires eight theorem prover calls.\qed 
\begin{align*}
	\mbox{\small (1)} \qquad [\underline{A_1, O_1, X_1, X_2}] & \;\;\;\textcircled{\scriptsize -} \quad \rightarrow \text{\quad the set $\{A_1,O_1,X_1,X_2\}$ includes a conflict}\\
	\mbox{\small (2)} \qquad [\underline{A_1, O_1}, X_1, X_2] & \;\;\;\textcircled{\scriptsize +} \quad \rightarrow \text{\quad some element from $\{X_1,X_2\}$ is in $\mc$}\\
	\mbox{\small (3)} \qquad[\underline{A_1, O_1, X_1}, X_2] & \;\;\;\textcircled{\scriptsize +} \quad \rightarrow \text{\quad $X_2$ is in $\mc$ }\\
	\mbox{\small (4)} \qquad[\underline{A_1, O_1}, X_1,  \uu{X_2}] & \;\;\;\textcircled{\scriptsize +} \quad \rightarrow \text{\quad $X_1$ is in $\mc$}\\ 
	\mbox{\small (5)} \qquad[A_1, O_1, \uu{X_1, X_2}] & \;\;\;\textcircled{\scriptsize -} \quad \rightarrow \text{\quad $A_1, O_1$ are not in $\mc$}
\end{align*}  
\end{example}
%

\section{Sequential Diagnosis}
\label{sec:sequential_diagnosis}
When a system malfunction is detected, there is in many cases not enough information 
available to locate the real cause unambiguously. Rather, one often faces a substantial number of different (minimal) diagnoses given the initial system observations and measurements. However, only one of the diagnoses, which we refer to as the \emph{actual diagnosis}, pinpoints the (unknown) actually faulty components. All other diagnoses are spurious in that they assume at least one actually normal component to be faulty, or at least one actually faulty component to be normal, or both.

One strategy to handle multiple diagnoses is to adopt techniques to rank or restrict the computed diagnoses based on some informative criterion such as maximal probability or minimal cardinality \cite{kleer_focusing_1991}. Whereas such 
techniques can be very powerful if the actual diagnosis appears (early) in the solution list, they cannot give the user a guarantee that the actually faulty components will be located (efficiently).

This is where sequential diagnosis techniques \cite{de_kleer_diagnosing_1987} enter the stage.
Their aim is to collect additional information about the system in order to gradually rule out spurious diagnoses and thereby reduce the diagnostic uncertainty. More specifically, sequential diagnosis approaches iteratively pose \emph{queries} to an oracle (e.g., an electrical engineer in case of a malfunctioning circuit, or a domain expert in case of a faulty knowledge base). At this, queries are selected in a way that each query answer elicits some new information about the diagnosed system that allows to discriminate between diagnoses. Depending on the system type, queries can capture, e.g., system tests (given specific inputs of the system, determine the resulting outputs), probes (make a measurement at a particular location in the system), or expert questions (answer a question about the intended behavior of the system) \cite{rodler_sequential_2022}.  
%
 
The basic technical idea behind queries is to exploit the fact that different diagnoses predict different system behaviors or properties (e.g., intermediate values or outputs). Observing a system aspect for which the predictions of multiple diagnoses disagree then leads to the invalidation of those diagnoses whose predictions are inconsistent with the query answer. 
Since the determination of optimal queries (which minimize the overall effort for the interacting oracle) is NP-hard \cite{pattipati_application_1990}, sequential diagnosis methods often have to rely on myopic strategies to estimate the favorability
of different query candidates. One approach is to evaluate the quality of queries in terms of the utility of the expected situation after knowing their answer, taking into account features such as the number of invalidated diagnoses, or the estimated additional oracle effort. 
Such \emph{one-step-lookahead analysis} \cite{de_kleer_one_1992} based on query selection heuristics has proven to constitute a particularly favorable trade-off between computational efficiency and diagnostic effectivity, and is today state-of-the-art in sequential diagnosis. Accordingly, a range of 
\emph{heuristics} 
that quantify the goodness of queries based on various information-theoretic considerations have been proposed in literature \cite{moret_decision_1982,de_kleer_diagnosing_1987,pattipati_application_1990,shchekotykhin_interactive_2012,rodler_rio_2013,rodler_towards_2016,rodler_active_2017,rodler_impact_2018}.

A generic \emph{sequential diagnosis process} 
can 
be roughly thought of as a recurring execution of 
\begin{enumerate}[label=\emph{(\arabic{*})},noitemsep,topsep=0pt]
	\item \label{enum:seq_diag_process_step1} the computation of a set of (preferred) minimal diagnoses,
	\item the selection of the most informative query according to some query selection heuristic based on the 
	computed diagnoses,
	\item the presentation of the computed query to the oracle which provides a query answer, and
	\item \label{enum:seq_diag_process_step4} the exploitation of the query outcome to refine the system knowledge.
\end{enumerate}
This iterative process continues until sufficient diagnostic certainty is obtained, e.g., only one minimal diagnosis (with overwhelming probability) remains.

\begin{figure}[t]
	\centering
	\begin{subfigure}{\linewidth}
		\centering
		\includegraphics[width=0.75\textwidth]{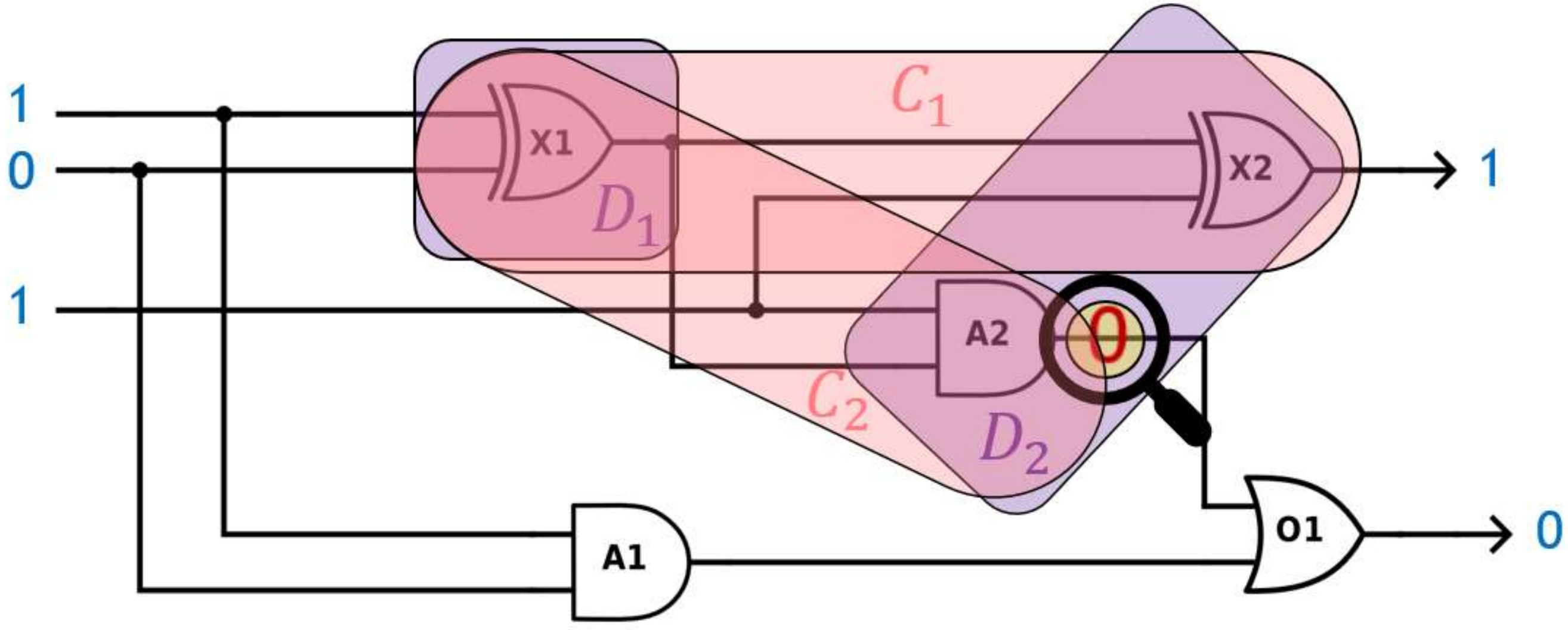}
		\caption{First measurement is $\mathit{out}(A_2) = 0$.}
		\label{fig:ex:circuit_seq_diag_after_meas_0}
		\vspace{15pt}
	\end{subfigure}
	\begin{subfigure}{\linewidth}
		\centering
		\includegraphics[width=0.75\textwidth]{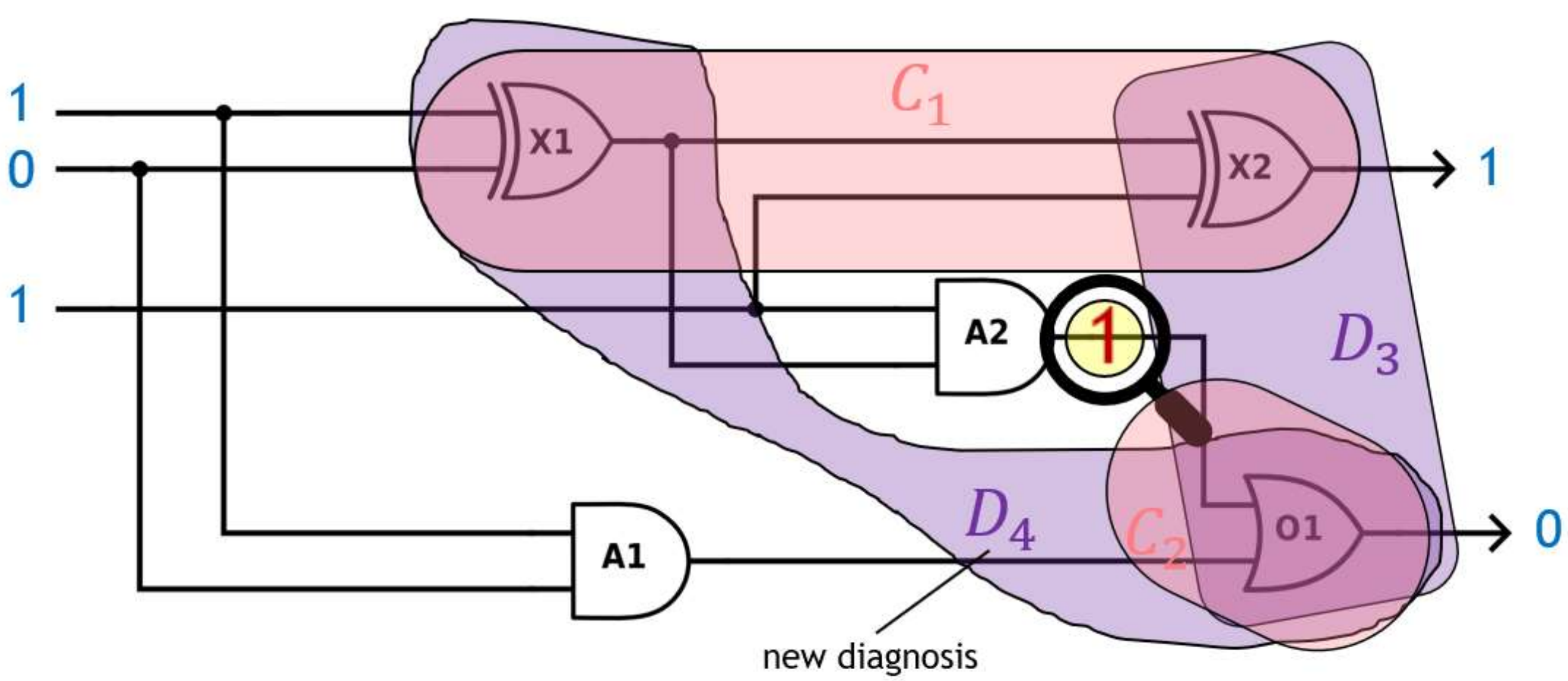}
		\caption{First measurement is $\mathit{out}(A_2) = 1$.}
		\label{fig:ex:circuit_seq_diag_after_meas_1}
	\end{subfigure}
	\label{fig:ex:circuit_seq_diag_after_meas}
	\caption[Full adder example: Sequential diagnosis, situation after the first measurement]{Full adder example: Sequential diagnosis, situation after the first measurement.}
\end{figure}

%
%

\begin{example}[Sequential diagnosis]\label{ex:circuit_seqDiag}
	Reconsider our full-adder from Fig.~\ref{fig:ex:circuit_MBD}. As expounded in Example~\ref{ex:circuit_conflicts+diagnoses}, the initial system knowledge in terms of observations and measurements leaves the user (e.g., an electrical engineer) with three minimal diagnoses. Let us assume that in the past the failure rate of the \mbox{or-gates} has been registered to be higher than for other types of gates. In the light of this meta information, diagnosis $\md_3 = [X_2,O_1]$ would appear to be the most likely and plausible. 
	However, let us also suppose that the intention is to avoid the potential cost involved in the unnecessary replacement of healthy gates as well as the cost of missing the replacement of faulty gates as a consequence of choosing a diagnosis that turns out not to be the actual fault explanation. 
	To this end, 
	the user might decide to use a sequential diagnosis engine as an aid to efficiently locate the real fault with certainty. 
	
	Initiating the sequential diagnosis process given the known diagnoses $\mD = \{\md_1,\md_2,\md_3\}$ (cf.\ Example~\ref{ex:circuit_conflicts+diagnoses}), a query selection engine might propose a query asking the engineer to measure the voltage at the wire $\mathit{out}(A_2)$ that connects the gates $A_2$ and $O_1$.  The rationale behind this query could be to differentiate between the most probable diagnosis $\md_3$ (being the only diagnosis including the or-gate) and the remaining diagnoses $\{\md_1,\md_2\}$. Note that this query has the property of eliminating one of the known diagnoses irrespective of the oracle's answer; that is, already before the query outcome is known, it is guaranteed that the query will bring relevant information for diagnosis discrimination. In various diagnostic scenarios, it is demanded of 
	query selection engines to generate only such discriminating queries. 
	
	Let us now analyze the effects the two possible query answers, $\mathit{out}(A_2) = 0$ and $\mathit{out}(A_2) = 1$, have on the minimal conflicts and diagnoses. Given $\mathit{out}(A_2) = 0$, we 
	learn that at least one of $\{X_1,A_2\}$ must be at fault, since assuming both of them to work nominally, the output of $X_1$ (inputs $1$ and $0$) would be $1$, and the output of $A_2$ (inputs $1$ and $1$) would be $1$ as well. Fig.~\ref{fig:ex:circuit_seq_diag_after_meas_0} shows the implication of this reasoning, i.e., the ``shrinkage'' of the minimal conflict $\mc_2$ to $\langle X_1,A_2\rangle$, which causes the invalidation of diagnosis $\md_2$ (as can be easily seen by means of the hitting-set relationship between diagnoses and conflicts). If, on the other hand, the oracle measures $\mathit{out}(A_2) = 1$, the conflict $\mc_2$ reduces to $\langle O_1\rangle$ because we now know that one of its inputs is $1$ (query answer) and its output is $0$ (observation), which entails that $O_1$ must be abnormal. 
	This means that, of the original minimal diagnoses, only $\md_3$ remains valid. However, notably, in this case we detect that a new diagnosis, $\md_4 = [X_1,O_1]$, has emerged as a superset of $\md_1$. In general, it holds that each minimal diagnosis after the query answer is incorporated into the system knowledge is either one of the original minimal diagnoses or a proper superset of some original minimal diagnosis.
	
	For any outcome we might have got for the first query, the debugger would suggest a measurement at $\mathit{out}(X_1)$ as a second query. Regardless of its outcome, diagnostic certainty will be obtained. To see this, let us consider all four cases: First, for the situation $\mathit{out}(A_2) = 0$ (Fig.~\ref{fig:ex:circuit_seq_diag_after_meas_0}), we have:
	\begin{itemize}[noitemsep,topsep=0pt]
		\item If $\mathit{out}(X_1) = 0$, the new minimal conflicts are given by $\mC = \{\langle X_1\rangle\}$. I.e., $\md_1 = [X_1]$ is the only remaining minimal diagnosis.
		\item If $\mathit{out}(X_1) = 1$, the new minimal conflicts are given by $\mC = \{\langle X_2\rangle, \langle A_2\rangle\}$. I.e., $\md_2 = [X_2,A_2]$ is the only remaining minimal diagnosis.
	\end{itemize} 
	Second, for the situation $\mathit{out}(A_2) = 1$ (Fig.~\ref{fig:ex:circuit_seq_diag_after_meas_1}), we get: 
	\begin{itemize}[noitemsep,topsep=0pt]
		\item If $\mathit{out}(X_1) = 0$, the new minimal conflicts are given by $\mC = \{\langle X_1\rangle, \langle A_2\rangle, \langle O_1\rangle\}$. I.e., $\md_4 = [X_1,A_2,O_1]$ is the only remaining minimal diagnosis.
		\item If $\mathit{out}(X_1) = 1$, the new minimal conflicts are given by $\mC = \{\langle X_2\rangle, \langle O_1\rangle\}$. I.e., $\md_3 = [X_2,O_1]$ is the only remaining minimal diagnosis.
	\end{itemize}
	%
	Hence, the user is overall required to make two measurements in the circuit in order to determine the actual diagnosis with full certainty.\qed
\end{example}

\section{Diagnostic Tasks}
\label{sec:diagnostic_tasks}
In model-based diagnosis, there are three main tasks \cite[Chap.~10]{van_harmelen_handbook_2008}: 
\begin{itemize}[noitemsep,topsep=0pt]
	\item \emph{fault detection} (is there a fault? / is some component faulty?),
	\item \emph{fault localization} (where is the fault? / which components are faulty?), and
	\item \emph{fault identification and repair} (what is faulty with each component? / how to fix it?).
\end{itemize}
Except for the case where faulty components are simply replaced by new ones, all these three steps are usually necessary for the successful debugging of an abnormal system, and they are executed in the order they are listed. Fault localization can be seen as a core phase in model-based diagnosis as it incorporates a fault detection step (localization is only relevant once it has been ascertained that faults are indeed present in the system), and its outcome represents one
solution to the repair problem (in terms of a simple, but often still reasonable and efficient component replacement). Given a strong fault model, where the system description, besides the normal behavior, also specifies abnormal component behavior(s) called fault modes, the fault identification task (find a fault mode assignment for each component that is consistent with the observations and measurements) might be directly tackled as a second step in the process, without first locating the fault. A more detailed discussion of this is beyond the scope of this work and can be found in \cite[Chap.~10]{van_harmelen_handbook_2008}. In any case, the fault identification and repair step will usually require domain expertise in terms of a more detailed insight into the functioning, structure, construction, etc.\ of the components. The works included in this thesis cover all three above-mentioned tasks, however with a focus on the fault localization problem.

%
%

\begin{figure}[!t]
	\centering
	\includegraphics[trim={0cm 0 0 0},clip,width=0.85\linewidth]{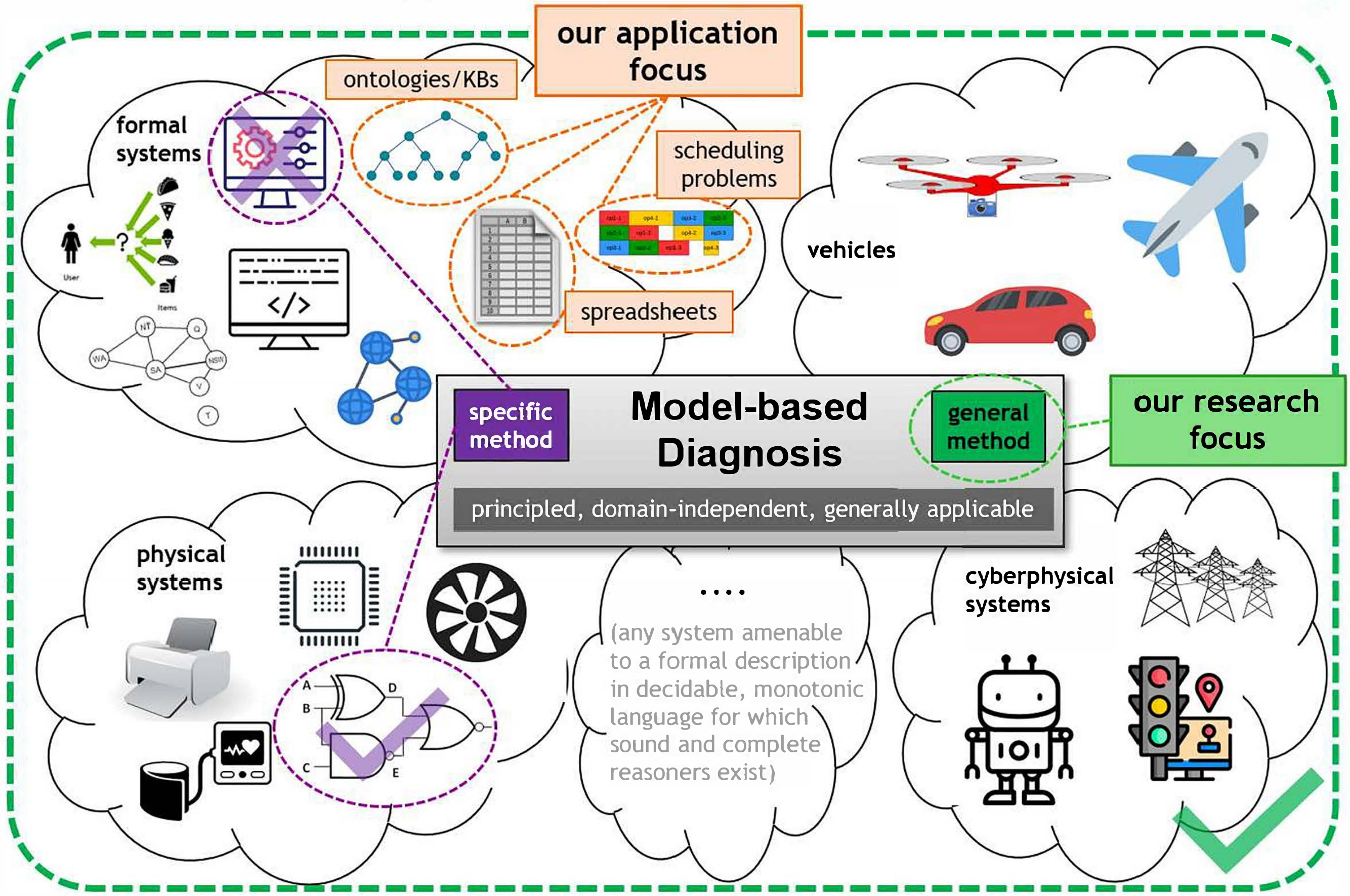}
	\caption[Model-based diagnosis application domains]{Model-based diagnosis application domains grouped into categories (\emph{clouds}), differentiation between specific (\emph{violet}) and general (\emph{green}) methods, and illustration of research and application focus (\emph{light green} and \emph{light orange}) of the works included in this thesis. The \emph{bold green dashed line and checkmark sign} indicate that general methods aim at preserving maximal generality in order to be (at least in principle) applicable to any model-based diagnosis application, which is not the case for specific methods (as exemplarily visualized by the \emph{violet dashed line}).}
	\label{fig:applications}
\end{figure}

\section{Application Areas}
\label{sec:application_areas}
Model-based diagnosis can be and has been applied to a myriad of different application areas. As suggested by Fig.~\ref{fig:applications}, we might categorize some of the prominent use cases into 
\begin{itemize}[noitemsep,topsep=0pt]
	\item \emph{formal systems} (e.g., software, spreadsheets, knowledge bases, ontologies, networks, scheduling problems, recommender systems, constraint satisfaction problems, or configuration systems),
	\item \emph{vehicles} (e.g., drones, aircrafts, or cars),
	\item \emph{cyberphysical systems} (e.g., power grids, traffic control systems, or robots), and 
	\item \emph{physical systems} (e.g., office machines, medical devices, circuits, hardware systems, or turbines). 
\end{itemize}
While the 
provision of a suitable system model might sometimes be one of the most challenging steps towards diagnosing (e.g., physical) systems when relying on the ``correct vs.\ actual'' interpretation of the diagnosis problem (cf.\ Sec.~\ref{sec:MBD}), a readily usable system model can often be directly extracted from the buggy system in an automatic way when considering the ``actual vs.\ intended'' perspective on the problem. The latter strategy is often applicable to formal systems, which 
constitute the primary focus of our empirical evaluations conducted in the course of the works included in this thesis.    

In general, any system that is decomposable into a set of components and amenable to a formal description in a decidable monotonic knowledge representation language, for which sound and complete theorem provers (for consistency checking) exist, is a potential use case for model-based diagnosis.

Whereas model-based diagnosis is a generally applicable and domain-independent framework, this does not automatically mean that all implementations of the framework, i.e., model-based debugging approaches and tools, are built so as to maintain the maximal generality. As suggested in \cite{rodler_how_2022}, we might discern specific and general diagnosis methods (cf.\ Fig.~\ref{fig:applications}). The former are designed to harness specifics of a particular (class of) system type(s) in order to streamline the diagnosis process while accepting a somewhat narrowed down application scope of the approach. The latter, by contrast, aim at maintaining a maximally broad applicability by making no assumptions about the (structure, type, domain, etc.\ of the) diagnosed system, the language used to describe the system, or the inference mechanism used. The largest part of the works included in this thesis belong to the latter category and can in principle be generally applied to any model-based diagnosis problem as per Reiter's theory \cite{reiter_theory_1987}. As regards applications,
our research works mainly target ontology and knowledge base debugging, spreadsheet diagnosis, as well as over-constrained scheduling problems (cf.\ Fig.~\ref{fig:applications}).

\section{Related Research Fields}
\label{sec:related_fields}

\begin{figure}[!t]
	\centering
	\includegraphics[trim={0cm 0 0 0},clip,width=0.7\linewidth]{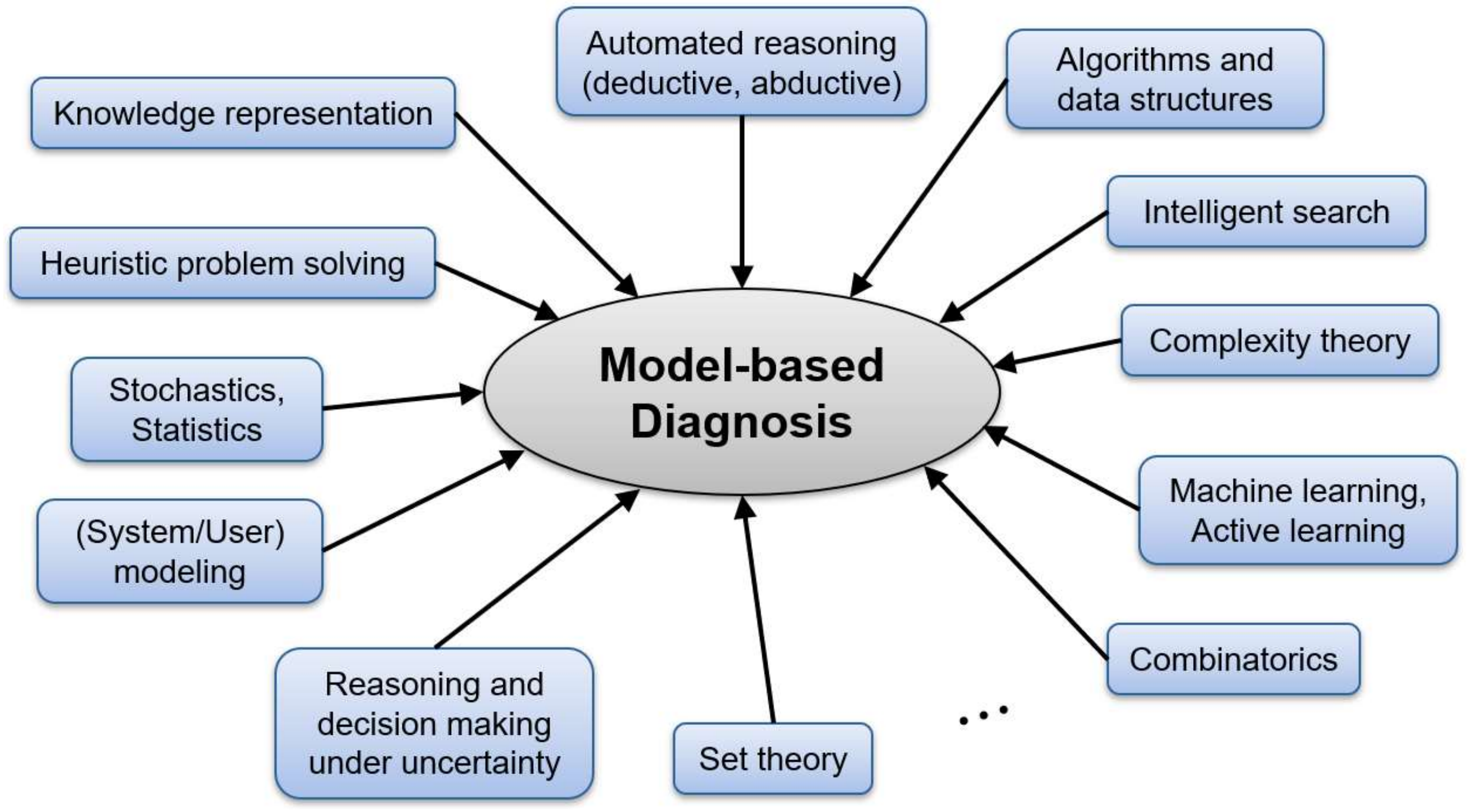}
	\caption[Related fields that influence model-based diagnosis]{(Some) related subjects and fields that influence model-based diagnosis research.}
	\label{fig:related_influencing_areas}
\end{figure}

\begin{figure}[!htp]
	\centering
	\includegraphics[trim={0cm 0 0 0},clip,width=0.7\linewidth]{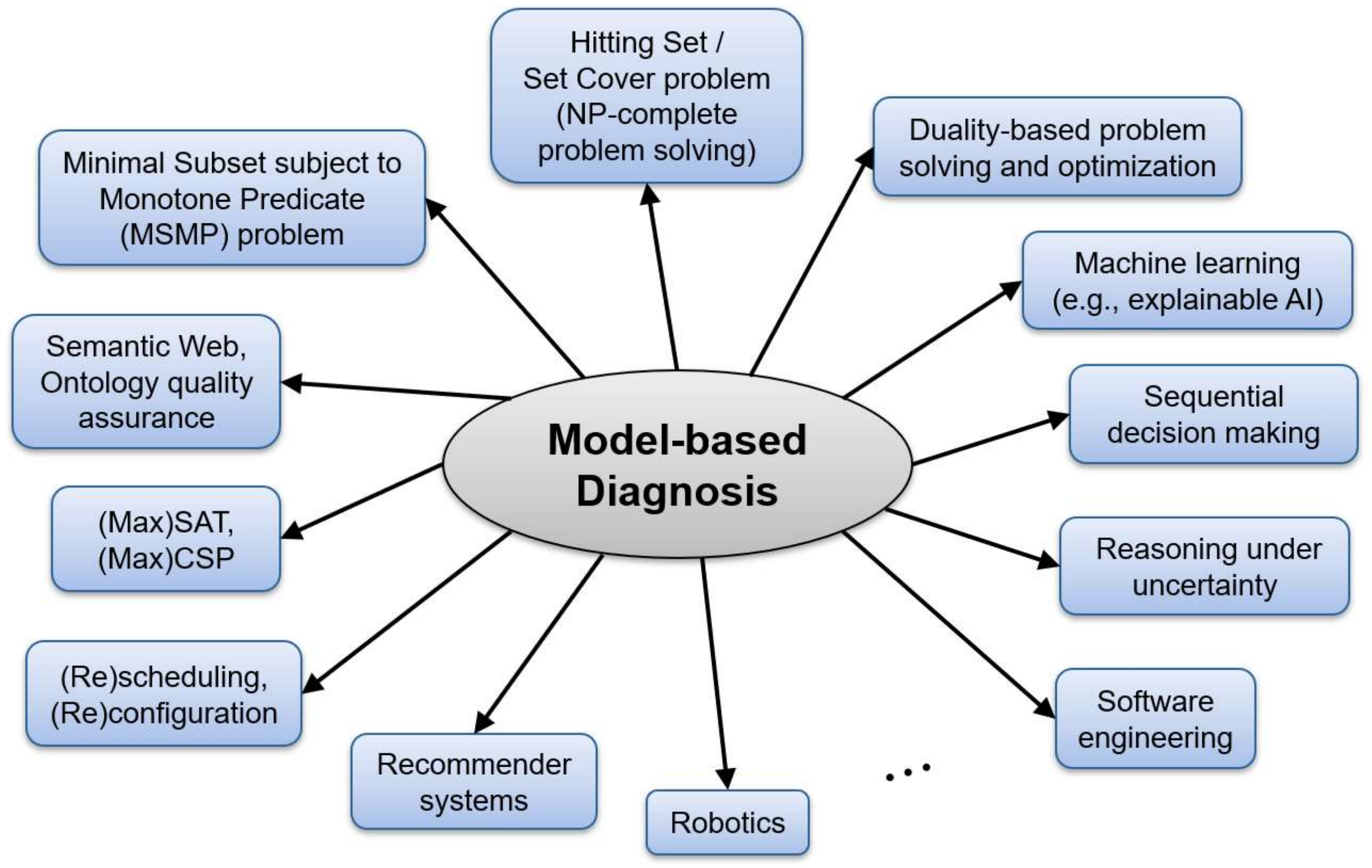}
	\caption[Related fields influenced by model-based diagnosis]{(Some) related subjects and fields that are influenced by model-based diagnosis research.}
	\label{fig:related_influenced_areas}
\end{figure}

Model-based diagnosis is a core discipline in model-based problem solving, which is embedded in the field of knowledge representation and reasoning, a large and important sub-area of artificial intelligence. By reason of 
its profound mathematical and theoretical foundations, 
the great variety of its applications,
the consideration of hard combinatorial and set theoretic problems,
the broad range of proposed algorithms, 
the diversity of possible diagnostic scenarios,
the partial observability and uncertainty usually involved in the process,
the usefulness of machine learning techniques to boost diagnostic performance,
and many others, model-based diagnosis has intersections with a multitude of other research areas. It is both influenced by and does itself influence a range of other fields, where Fig.~\ref{fig:related_influencing_areas} and Fig.~\ref{fig:related_influenced_areas} give an overview of some of them. 

Prominent influencing fields are, by the very nature of model-based diagnosis, of course knowledge representation, reasoning, and modeling techniques, as these fields deal with languages and ways to express system descriptions and with efficient automated deduction techniques on top of these formalisms for model-based consistency checking. Discrete-mathematical subjects such as combinatorics and set theory, as well as complexity theory, stochastics, and algorithms and data structures are central to theoretical analyses in the field. Moreover, given the NP-hardness and thus general intractability of the problems addressed in model-based diagnosis (e.g., diagnosis computation or query optimization), heuristic problem solving and intelligent search techniques as well as machine learning approaches can in some settings be vital to the development of practical and successful diagnosis methods. Finally, statistics and uncertain reasoning
play a pivotal role in dealing with the problem of intelligent and reasonable decision making given only partial information.

On the other hand, also model-based diagnosis research has contributed to advancements in several fields. First, there are
the various domains it has been successfully applied in, such as robotics, recommender systems, software engineering, the Semantic Web, ontology engineering, as well as scheduling and configuration problems. 
Apart from its application-related impact, model-based diagnosis has brought up substantial theoretical findings and algorithmic solutions that (can) provide valuable ideas and insights for, or propel the state-of-the-art of important research domains. Examples are the hitting-set (and the closely related set-cover) problem, which is one of the key problems studied in model-based diagnosis due to the hitting-set relationship between diagnoses and conflicts (cf.\ Sec.~\ref{sec:MBD}). Worth mentioning in this context is also the field of duality-based problem solving and optimization, as characterized in \cite{slaney_set-theoretic_2014}. Here, the problem of finding a (cost-)optimal set from a collection $\mathbf{X} = \{S_1,\dots,S_k\}$ of subsets $S_i$ of some universe $U$ is investigated, and suggested to be tackled by using best-first hitting set computation (e.g., \cite{reiter_theory_1987,rodler_interactive_2015}) over the dual collection $\mathbf{X}^*$ of $\mathbf{X}$. At this, $\mathbf{X}^*$ is defined as the collection including all subsets $S^*_j$ of $U$ whose complement $U \setminus S^*_j$ is not in $\mathbf{X}$. Since the collection of diagnoses is dual to the collection of conflicts, and vice versa, progress in conflict-based diagnosis computation can often be directly exploited for arbitrary duality-based computation or optimization problems. Other prominent notions linked by such a duality relationship are, e.g., minimal unsatisfiable subsets vs.\ minimal correction subsets of clauses in a CNF \cite{bacchus_using_2015}, which are highly relevant to SAT research \cite{biere_handbook_2009}, or conflicts vs.\ relaxations in the context of constraint satisfaction problems \cite{junker_quickxplain_2004}. 
Another interesting observation is that  
the computation of both minimal diagnoses and minimal conflicts is an instance of the so-called Minimal Subset subject to a Monotone Predicate (MSMP) problem \cite{marques-silva_minimal_2013}. As a consequence, efficient methods for diagnosis and conflict calculation, if sufficiently general (cf.\ Sec.~\ref{sec:application_areas} and \cite{rodler_formal_2022}), also constitute solutions to this more general MSMP problem. Instances of the latter include, e.g., minimal unsatisfiable subsets, minimal unsatisfiable cores, minimal correction subsets, prime implicants and implicates, as well as justifications and most concise optimal queries to an oracle \cite{de_kleer_diagnosing_1987,marquis_knowledge_1995,birnbaum_consistent_2003,dershowitz_scalable_2006,horridge_justification_2011,rodler_sequential_2022}. Furthermore, techniques from model-based diagnosis are relevant to sequential decision making, reasoning under uncertainty, and machine learning. Particularly noteworthy in this regard is 
the more recent field of explainable AI which has been gaining increasing attention and interest throughout the last years and which is devoted to supplying users of artificial intelligence and especially machine learning tools with understandable and useful explanations for generated outputs.
Finally, the principles underlying model-based diagnosis approaches share several crucial parallels with other model-based problem solving techniques such as system design, (re)configuration, or failure-modes-and-effects analysis \cite[Chap.~10]{van_harmelen_handbook_2008}. Thus, ideas and approaches for one area can be fruitful and beneficial also for the others. 




%
%
%



\begin{figure}
	\centering
	\includegraphics[trim={0cm 0 0 0},clip,width=0.98\linewidth]{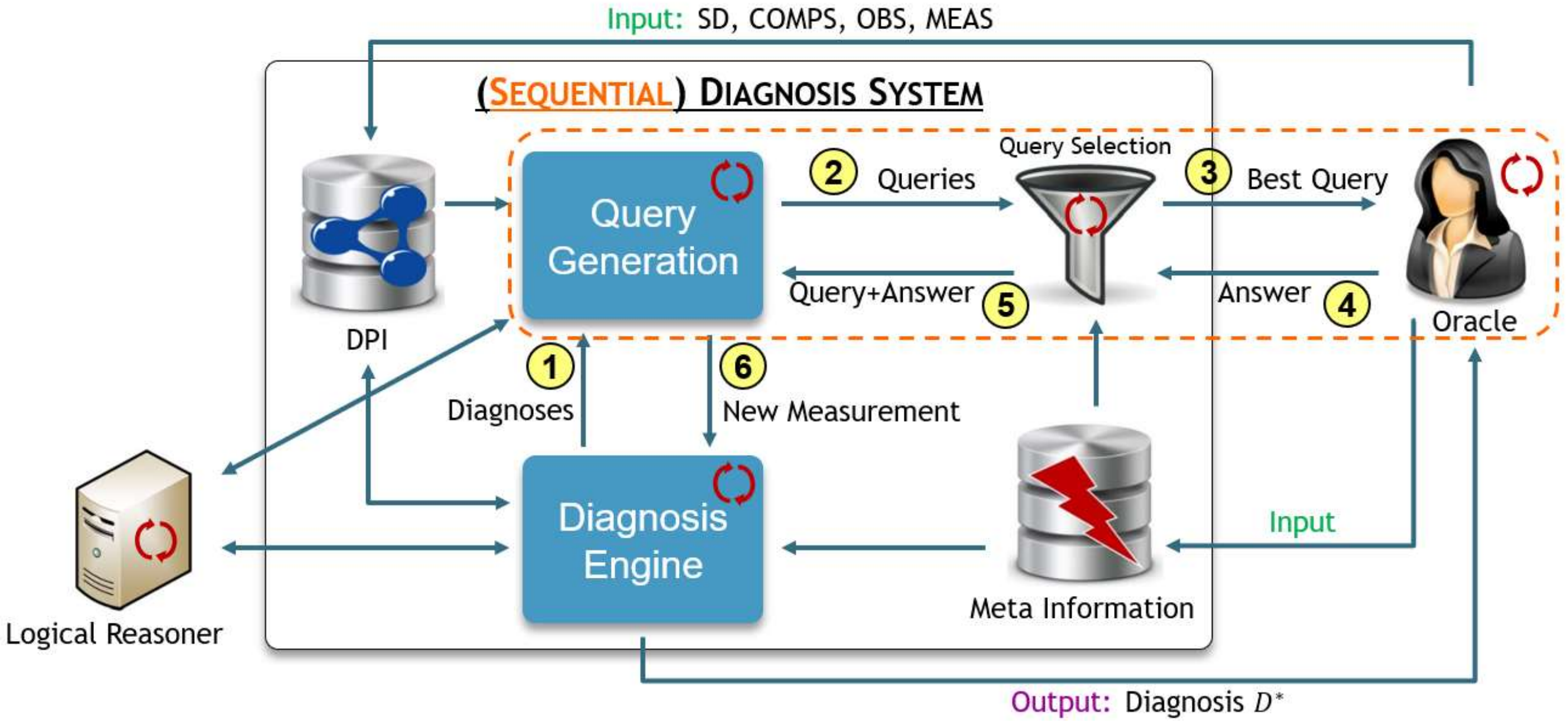}
	\caption[Generic (sequential) diagnosis system: Modules, external entities, functionality]{Generic (sequential) diagnosis system: Inputs, outputs, modules, stored information, external entities, and functionality. The \emph{orange dashed line} shows the elements additionally present if the debugging system targets sequential diagnosis. A \emph{circle of two red arrows} indicates that the operation executed by the respective entity is potentially expensive (e.g., wrt.\ time, memory, cost, or effort). The \emph{yellow circled numbers} visualize the process flow; the depicted loop is reiterated until a predefined stop criterion is satisfied.}
	\label{fig:generic_diagnosis_system_functionality}
\end{figure}

\section{A Generic (Interactive) Diagnosis System}
\label{sec:generic_diagnosis_system}
In order to 
get a feel for the problems, challenges, and research goals in the context of (sequential) model-based diagnosis (discussed in Sec.~\ref{sec:research_challenges+goals}), it is instructive to examine, on a sufficiently abstract level, the structure, modules, properties, and other aspects shared by most, if not all, (interactive) diagnosis tools and approaches. To this end, let us consider a generic diagnosis system, as depicted in Fig.~\ref{fig:generic_diagnosis_system_functionality}: \vspace{4pt}

\noindent\emph{Structure and external entities:} On the structural level, we have the diagnosis system and two entities outside of the system, viz the oracle (or: user) and the logical reasoner. Within the system, we discern two databases (in the most abstract conceptual sense), one storing the current diagnosis problem instance (DPI), and the other storing any meta information relevant to the modules in the system or to the present diagnostic scenario or application. Often, the meta information encompasses, e.g., fault information (such as component failure rates), parameter settings for algorithms (e.g., stopping or optimization criteria), heuristics (e.g., for query selection), or general preferences or configurations of the diagnosis system (e.g., which type of queries to ask \cite{rodler_one_2022}, or which type of problem to solve \cite{rodler_statichs_2018}). In addition, the system incorporates three modules, viz a diagnosis engine, a query generation unit, as well as a query selection component.
Note that, in case the diagnosis system is not intended for the purpose of sequential diagnosis, it does not need to include the query generation and query selection modules, and no communication with an oracle is necessary during the process (see 
the orange dashed line in Fig.~\ref{fig:generic_diagnosis_system_functionality}).\vspace{4pt}

\noindent\emph{Inputs and outputs (cf.\ green and violet font in Fig.~\ref{fig:generic_diagnosis_system_functionality}):} The inputs to a diagnosis system are, first, a diagnosis problem instance (DPI) capturing the initial state of knowledge about the diagnosed system in terms of the system description (\textsc{sd}), the components (\textsc{comps}), as well as the observations (\textsc{obs}) and measurements (\textsc{meas}), cf.\ Sec.~\ref{sec:MBD}. Technically, \textsc{sd}, \textsc{obs}, and \textsc{meas} will each be a set of sentences in a specific formal (often: logical) language, such as propositional logic, predicate calculus, description logics, finite-state machines, differential equations, or constraints, and \textsc{comps} a set of the system components of interest to the diagnostic task. Second, the user might (often: optionally) provide available meta information as described above. The output of a diagnosis system can in general take manifold shapes, depending on the type, application context, purpose, goal, way of deployment, etc.\ of the diagnosis system as well as of the user's preferences (cf.\ \cite{rodler_how_2022}). Usually, the output will be either one (often: minimal or preferred) diagnosis, or an (often: small) set of diagnoses (often: sorted by some preference criterion).\vspace{4pt}

\noindent\emph{Process flow (cf.\ yellow circled numbers in Fig.~\ref{fig:generic_diagnosis_system_functionality}, and enumeration \ref{enum:seq_diag_process_step1}--\ref{enum:seq_diag_process_step4} in Sec.~\ref{sec:sequential_diagnosis}):} Given the DPI and potentially some relevant meta information such as fault probabilities, a preference criterion, a number of diagnoses to compute, a timeout, or other algorithm-specific parameters, the first step, accomplished by the diagnosis engine, is the computation of a set of so-called \emph{leading} minimal diagnoses \cite{de_kleer_diagnosis_1989}, which are often the most preferred diagnoses, i.e., the most probable or minimum-cardinality ones. If the system is not sequential, then these diagnoses might be directly presented to the user (perhaps in the form of a ranking) for further inspection or processing. Otherwise, in a second step, the leading diagnoses are forwarded to the query generation unit, which usually generates a set of query candidates. The complexity of this step can range from trivial (e.g., if the possible queries, such as probing points in the form of wires in a circuit, are immediately extractable from the system description \cite{de_kleer_diagnosing_1987}) to very sophisticated (e.g., if queries, such as possible expert questions in ontology debugging, are implicit and computationally hard to extract \cite{rodler_sequential_2022}). In a third step, the query selection component filters the query candidates for the best (or a sufficiently good) query according to some (heuristic) measure given as a part of the meta information. The best query is then, in a fourth step, posed to the oracle, which performs according measurements, probes, tests, etc.\ in case of a tangible system under inspection, or thinks about the correct answer to a given expert question in case of a formal system to be debugged. The oracle is usually assumed to be capable or knowledgable enough to answer an adequate fraction of the possible queries such that the sequential diagnosis session can reasonably proceed (after an acceptable number of queries are perhaps skipped and replaced by alternative ones). The provided answer of the oracle along with the asked query is then used to formulate a (logical) sentence to be added to the DPI in terms of a new measurement in the course of the fifth and sixth steps. This loop is continued until some stop criterion is met, e.g., an adequate degree of diagnostic certainty is achieved (cf.\ Sec.~\ref{sec:sequential_diagnosis}).


\begin{figure}
	\centering
	\includegraphics[trim={0cm 0 0 0},clip,width=0.98\linewidth]{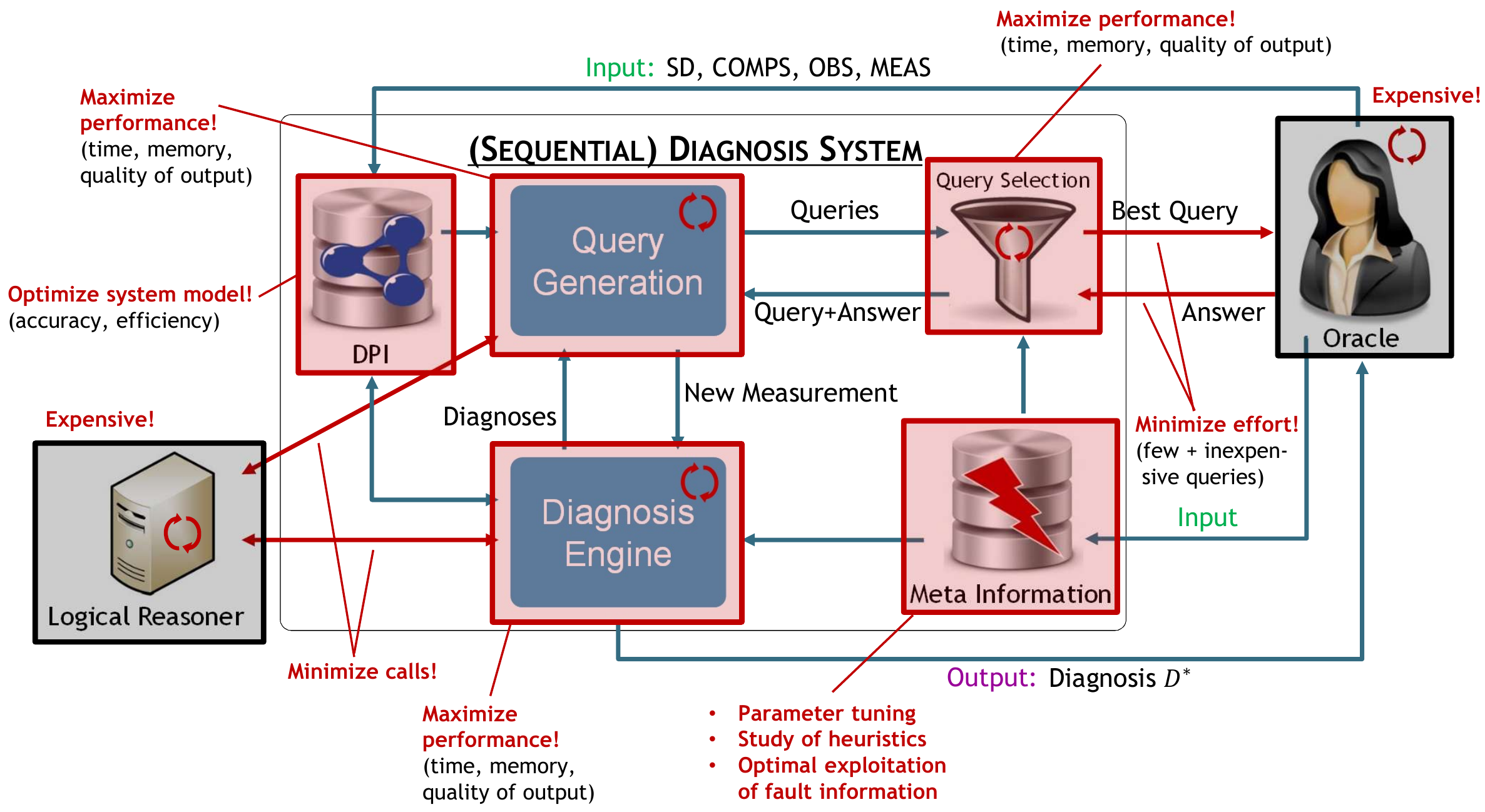}
	\caption[Generic (sequential) diagnosis system: Problems, research challenges, goals]{Generic (sequential) diagnosis system: Problems as well as research challenges and goals. \emph{Black boxes} illustrate that the respective entity can (often) not be directly influenced by the diagnosis system, i.e., is considered a black-box with a defined output given a defined input, but unknown internals. The meaning of \emph{red boxes} is that the respective entities can be actively shaped in that they are the result of design choices when building the diagnosis system, i.e., their internals can basically be implemented, organized and configured at will. \emph{Thick red arrows} indicate that the communication described by this arrow is generally expensive. The research challenges and goals are depicted by \emph{red-colored text}.}
	\label{fig:generic_diagnosis_system_goals}
\end{figure}

\section{Research Challenges and Goals}
\label{sec:research_challenges+goals}

Let us now take a look at the generic diagnosis system discussed in the previous section through the eyes of a researcher, by evaluating components and other aspects of the system regarding their influence on the \emph{diagnostic performance} in terms of time, memory or other costs, and on the \emph{diagnostic effectivity} in terms of the quality of the produced 
outputs. 
The result of this analysis is displayed in Fig.~\ref{fig:generic_diagnosis_system_goals}.\vspace{4pt}

\noindent\emph{External entities:} Both the calls to a logical reasoner and the consultations of an interacting oracle are usually among the most expensive ``operations'' throughout a diagnosis session. Since both these entities are frequently 
(and also in the works included in this thesis) viewed as black-boxes\footnote{\emph{Regarding the logical reasoner}, \cite{rodler_how_2022} gives an overview of some prominent diagnosis approaches, while also showing that most of them view the reasoner as a black-box. 
Alternatively, one can consider a reasoner as a so-called glass-box, where the idea is to streamline the reasoner's internals towards more efficient computations relevant to the diagnostic task (e.g., by extracting conflicts as a byproduct of negative consistency checks).
Studies contrasting black-box and glass-box techniques 
can be found, e.g., in \cite{horridge_justification_2011,kalyanpur_debugging_2006}. 
Roughly, their conclusion is that the latter usually outperform the former wrt.\ computation time. On the other hand, advantages of black-box methods
are their \emph{(1)}~robustness (no sophisticated, 
error-prone modifications of complex reasoning algorithms), 
\emph{(2)}~simplicity (internals of reasoner irrelevant), 
\emph{(3)}~flexibility (e.g., one can use a portfolio reasoning approach by switching to the most efficient reasoner in a simple plug-in fashion depending on the language used to describe the diagnosed system \cite{romero_more_2012}), and 
\emph{(4)}~up-to-dateness (which allows users to directly benefit from advances in general research on automated reasoning). 
\emph{Concerning the interacting oracle}, it was shown that making suggestions to the user wrt.\ their answering behavior (e.g., by soliciting additional inputs or explanations) can bring a significant boost in terms of fault localization efficiency \cite{rodler_one_2022}. Similar to the case of the logical reasoner, this could be seen as an interpretation of the oracle as a ``glass-box''. Nevertheless, presumably for simplicity, it appears that most methods in the literature regard the oracle as a black-box.}, i.e., as functions with unknown internals that yield a defined output given a defined input, an important goal towards efficient model-based diagnosis is to minimize the communication with both the reasoner and the query-answering oracle (cf.\ the thick red arrows in Fig.~\ref{fig:generic_diagnosis_system_goals}). \vspace{4pt}

\noindent\emph{Internal components:} For the three modules performing computations, i.e., the diagnosis engine, the query generation, and the query selection, the objective is to maximize their time and memory performance as well as the quality of their output, where the definition of quality might depend on, e.g., the user's preferences, the meta information, the given diagnostic scenario, or the used algorithms. In addition, diagnosis-specific gains can also be achieved by leveraging modifications of the stored data, i.e., the meta information and DPI. Concerning the latter, since there are always multiple possible ways of modeling one and the same thing, one goal could be to find a representation of the diagnosed system that optimally balances diagnostic accuracy (reasonable number of generated diagnoses) with diagnostic efficiency (reasonable diagnosis computation time). Regarding the meta information, possible goals are, e.g., to find optimal configurations of the algorithms involved in the diagnosis process, optimal settings that suit a particular (group of) user(s) best, optimal heuristics (e.g., for query selection) to employ based on the given diagnosis scenario, or the optimal usage of the given (and possibly misleading \cite{rodler_rio_2013}) fault information in the process. \vspace{4pt} 


\noindent\emph{Summary:} These considerations boil down 
to the following research objectives (cf.\ red text in Fig.~\ref{fig:generic_diagnosis_system_goals}):
\begin{itemize}[noitemsep,topsep=0pt]
	\item \emph{Diagnosis engine:} Maximize the performance wrt.\ time, memory and output quality.
	\item \emph{Query generation:} Maximize the performance wrt.\ time, memory and output quality.
	\item \emph{Query selection:} Maximize the performance wrt.\ time, memory and output quality.
	\item \emph{DPI:} Optimize the system model wrt.\ diagnostic accuracy and efficiency.
	\item \emph{Meta information:} Optimize the provided data for best diagnostic performance.
	\item \emph{Logical reasoner:} Minimize the expense of communications with the reasoner.
	\item \emph{Oracle:} Minimize the expense for the oracle wrt.\ time, effort, and other costs.	 
\end{itemize}
The research works included in this thesis consider all these research objectives, i.e., they cover all modules and aspects of the discussed generic diagnosis system. The contributions of these works will be explored in more detail in the upcoming chapters.

\newpage\null\thispagestyle{empty}\newpage

\chapter{Our Research: Overview}
\label{chap:research_overview}
The aim of this chapter is to give the reader a concise overview of the conducted research in which the author of this thesis was involved. First, we approach this task in two orthogonal ways, by synopsizing the research works \emph{(i)}~in Sec.~\ref{sec:research_based_on_diagnosis_system_aspect}, from the perspective of the aspects of 
a generic diagnosis system 
that are addressed, 
as well as \emph{(ii)}~in Sec.~\ref{sec:research_based_on_type_of_paper}, regarding their type (e.g., whether the respective paper is theoretical, empirical, application-related, etc.). 
Finally, in Sec.~\ref{sec:research_works_incl_in_this_thesis_in_a_nutshell}, we focus our attention on the works included in this thesis and outline them from the two perspectives (i) and (ii), along with other relevant information.

\begin{table}[t]
	\footnotesize
	\centering
	\rowcolors{2}{white}{gray!20}
	\begin{tabular}{@{}>{\columncolor{white}[0pt][\tabcolsep]}l|p{10.7cm}>{\columncolor{white}[\tabcolsep][0pt]}l@{}}
		\rowcolor{gray!70}
		\textbf{Aspect of Diagnosis System} & \textbf{Research Works where Aspect is Addressed} \\ \hline
		Diagnosis Engine           &  \cite{hofer_modeling_2022,rodler_appendix_2022,rodler_dynamichs_2022,rodler_formal_2022,rodler_how_2022,rodler_memory-limited_2022,rodler_random_2022,rodler_appendix_2021,rodler_linear-space_2021,rodler_randomized_2021,rodler_dynamichs_2020,rodler_reuse_2020,rodler_sound_2020,rodler_too_2020,rodler_we_2020,rodler_scheduling_2020,rodler_towards_2019,rodler_are_2019-1,rodler_reducing_2018,rodler_statichs_2018,schekotihin_ontodebug_2018,rodler_theory_2016,rodler_interactive_2015,shchekotykhin_direct_2014,shchekotykhin_sequential_2014,shchekotykhin_direct_2012}              \\
		Query Generation           &  \cite{rodler_appendix_2022,rodler_one_2022,rodler_sequential_2022,rodler_expert_2020,rodler_how_2019,rodler_new_2019,rodler_usefulness_2019,rodler_active_2017,rodler_generally_2017,rodler_inexpensive_2017,rodler_theory_2016,rodler_towards_2016,rodler_interactive_2015,shchekotykhin_interactive_2012}              \\
		Query Selection            &  \cite{rodler_appendix_2022,rodler_one_2022,rodler_random_2022,rodler_sequential_2022,rodler_we_2020,rodler_new_2019,rodler_comparing_2018,rodler_impact_2018,rodler_active_2017,rodler_generally_2017,rodler_inexpensive_2017,rodler_towards_2016,rodler_interactive_2015,shchekotykhin_interactive_2014,shchekotykhin_sequential_2014,rodler_rio_2013,rodler_rio_2012,shchekotykhin_interactive_2012,shchekotykhin_direct_2012}               \\
		DPI                        &   \cite{hofer_modeling_2022,rodler_dynamichs_2022,rodler_appendix_2021,rodler_reuse_2020,rodler_are_2019-1,rodler_statichs_2018,schekotihin_ontodebug_2018,schekotihin_protege_2018,schekotihin_test-driven_2018,rodler_reducing_2017}             \\
		Meta Information           &  \cite{rodler_appendix_2022,rodler_dynamichs_2022,rodler_how_2022,rodler_memory-limited_2022,rodler_one_2022,rodler_random_2022,rodler_rbf-hs_2022,rodler_sequential_2022,rodler_appendix_2021,rodler_we_2020,rodler_are_2019-1,rodler_usefulness_2019,rodler_comparing_2018,rodler_impact_2018,schekotihin_ontodebug_2018,rodler_active_2017,rodler_interactive_2015,shchekotykhin_interactive_2014,shchekotykhin_sequential_2014,rodler_rio_2013,rodler_rio_2012,shchekotykhin_interactive_2012,shchekotykhin_direct_2012,rodler_balancing_2011}               \\
		Logical Reasoner           &  \cite{rodler_dynamichs_2022,rodler_formal_2022,rodler_how_2022,rodler_appendix_2021,rodler_randomized_2021,rodler_reuse_2020,rodler_scheduling_2020}              \\
		Oracle                     &  \cite{rodler_one_2022,rodler_how_2019,rodler_new_2019,rodler_usefulness_2019,rodler_are_2019-1,schekotihin_ontodebug_2018,schekotihin_protege_2018,schekotihin_test-driven_2018}             
	\end{tabular}
	\caption[Research works grouped based on the addressed aspect(s) of a generic diagnosis system]{Research works grouped based on the addressed aspect(s) of a generic diagnosis system.
	Per group (i.e., table row), works are arranged chronologically and sorted by author name(s). All works with year tokens larger or equal to ``16'' have been carried out during the author's habilitation period; 
	we also include earlier works in the lists in order to give a more complete, reasonable and illustrative picture of the author's research.}
	\label{tab:research_based_on_system_aspect_addressed}
\end{table}

\section{From the Diagnosis System Perspective}
\label{sec:research_based_on_diagnosis_system_aspect}
Recalling the generic diagnosis system explicated in Sec.~\ref{sec:generic_diagnosis_system} and the associated research problems discussed in Sec.~\ref{sec:research_challenges+goals}, Tab.~\ref{tab:research_based_on_system_aspect_addressed} enumerates our research works grouped by the diagnosis system aspect (and related research objective) that was addressed in the respective paper. Note that several papers consider multiple system aspects and thus occur in different groups. As the table shows, we have been active along various lines of research, and we devoted our research to all research objectives identified above. This is a result of our holistic perspective on the model-based diagnosis field and of our overarching goal to advance the state-of-the-art of diagnosis systems along different dimensions.

\section{From the Type of Research Perspective}
\label{sec:research_based_on_type_of_paper}
To be able to better categorize and assess the conducted research, it can be helpful to view the works from another different angle. To this end, Tab.~\ref{tab:research_based_on_type_of_work} lists the research works grouped by the type of the respective paper. At this, we discern
\begin{itemize}[noitemsep,topsep=0pt]
	\item papers proposing novel methodologies or algorithms (\emph{Methodology/Algorithm}),
	\item application papers which are---notwithstanding the generality of the suggested approaches, cf.\ Sec.\ref{sec:application_areas}---devoted to one particular application of model-based diagnosis (\emph{Application}),
	\item explanatory papers of didactic or survey nature (\emph{Explanatory/Didactic/Survey}),
	\item empirical papers including comprehensive experimental evaluations or user studies (\emph{Empirical}),
	\item papers including substantial theoretical results such as algorithm derivations, correctness proofs or complexity analyses (\emph{Theoretical}),
	\item critical papers that fundamentally challenge the state-of-the-art (\emph{Critical}), and
	\item tool papers describing, analyzing, evaluating, and differentiating debugging tools (\emph{Tool}).
\end{itemize}

The table demonstrates that our research activities not only consider different research avenues, but have also produced a heterogeneous collection of works in terms of the nature of the publications.

\begin{table}[t]
	\footnotesize
	\centering
	\rowcolors{2}{white}{gray!20}
	\begin{tabular}{@{}>{\columncolor{white}[0pt][\tabcolsep]}l|p{10.7cm}>{\columncolor{white}[\tabcolsep][0pt]}l@{}}
		\rowcolor{gray!70}
		\textbf{Type of Work}	   & \textbf{Research Works of this Type} \\ \hline
		Methodology/Algorithm      &  \cite{hofer_modeling_2022,rodler_appendix_2022,rodler_dynamichs_2022,rodler_how_2022,rodler_memory-limited_2022,rodler_one_2022,rodler_random_2022,rodler_rbf-hs_2022,rodler_sequential_2022,rodler_appendix_2021,rodler_linear-space_2021,rodler_randomized_2021,rodler_dynamichs_2020,rodler_expert_2020,rodler_reuse_2020,rodler_sound_2020,rodler_too_2020,rodler_we_2020,rodler_scheduling_2020,rodler_towards_2019,rodler_how_2019,rodler_usefulness_2019,rodler_are_2019-1,rodler_reducing_2018,rodler_statichs_2018,rodler_active_2017,rodler_generally_2017,rodler_inexpensive_2017,rodler_towards_2016,rodler_rio_2013,rodler_rio_2012,shchekotykhin_direct_2012,rodler_balancing_2011}               \\
		Application		           &   \cite{hofer_modeling_2022,rodler_one_2022,rodler_randomized_2021,rodler_scheduling_2020,rodler_how_2019,rodler_usefulness_2019,rodler_are_2019-1,rodler_impact_2018,schekotihin_ontodebug_2018,schekotihin_protege_2018,schekotihin_test-driven_2018,rodler_rio_2012,shchekotykhin_interactive_2012,shchekotykhin_direct_2012,rodler_balancing_2011},             \\
		Explanatory/Didactic/Survey&   \cite{rodler_formal_2022,rodler_how_2022,schekotihin_ontodebug_2018,schekotihin_test-driven_2018,rodler_active_2017}             \\
		Empirical                  &    \cite{hofer_modeling_2022,rodler_appendix_2022,rodler_dynamichs_2022,rodler_memory-limited_2022,rodler_one_2022,rodler_random_2022,rodler_rbf-hs_2022,rodler_sequential_2022,rodler_appendix_2021,rodler_linear-space_2021,rodler_randomized_2021,rodler_reuse_2020,rodler_sound_2020,rodler_too_2020,rodler_we_2020,rodler_scheduling_2020,rodler_how_2019,rodler_usefulness_2019,rodler_are_2019-1,rodler_comparing_2018,rodler_reducing_2018,rodler_statichs_2018,rodler_impact_2018,rodler_active_2017,shchekotykhin_sequential_2014,rodler_rio_2013,shchekotykhin_interactive_2012,shchekotykhin_direct_2012,rodler_balancing_2011}            \\
		Theoretical		           &   \cite{hofer_modeling_2022,rodler_appendix_2022,rodler_dynamichs_2022,rodler_formal_2022,rodler_how_2022,rodler_memory-limited_2022,rodler_one_2022,rodler_rbf-hs_2022,rodler_sequential_2022,rodler_appendix_2021,rodler_randomized_2021,rodler_reuse_2020,rodler_scheduling_2020,rodler_statichs_2018,rodler_active_2017,rodler_reducing_2017,rodler_generally_2017,rodler_inexpensive_2017,rodler_theory_2016,rodler_towards_2016,rodler_interactive_2015,shchekotykhin_sequential_2014,rodler_rio_2013,shchekotykhin_interactive_2012,shchekotykhin_direct_2012,rodler_balancing_2011}             \\
		Critical		           &   \cite{rodler_one_2022,rodler_random_2022,rodler_are_2019-1}             \\
		Tool                       &     \cite{rodler_how_2022,rodler_are_2019-1,schekotihin_ontodebug_2018,schekotihin_protege_2018,schekotihin_test-driven_2018}          
	\end{tabular}
\caption[Research works grouped by type of paper]{Research works grouped by type of paper. Per group (i.e., table row), works are arranged chronologically and sorted by author name(s). All works with year tokens larger or equal to ``16'' have been carried out during the author's habilitation period; we also include earlier works in the lists in order to give a more complete, reasonable and illustrative picture of the author's research.}
\label{tab:research_based_on_type_of_work}
\end{table}

\section{Works Included in this Thesis: In a Nutshell}
\label{sec:research_works_incl_in_this_thesis_in_a_nutshell}
Let us now have a look at the summary of the research works included in this thesis given by Tab.~\ref{tab:research_works_incl_in_thesis_summary}. The table can be read row-wise to learn about the main focus and type of a particular paper, and column-wise to find the papers that address a particular system aspect or are of a particular type. Note that many of the papers consider more than the system aspects mentioned in the table, but for clarity the table indicates only the topics that constitute the papers' emphasis (please see Tab.~\ref{tab:research_based_on_system_aspect_addressed} and 
for an overview that considers also secondary matters investigated by the papers).

\setlength{\tabcolsep}{5pt}

\begin{table}[t!]
	\scriptsize
	\centering
	\rowcolors{2}{white}{gray!20}
	\begin{tabular}{@{}>{\columncolor{white}[0pt][\tabcolsep]}lclll|lllllll|lllllll>{\columncolor{white}[\tabcolsep][0pt]}l@{}}
		\hline
		\rowcolor{gray!70}
		\multicolumn{5}{c|}{\bfseries Research Work Included in Thesis}                                                     & \multicolumn{7}{c|}{\bfseries Diagnosis System Aspects in Focus}                                                                                                & \multicolumn{7}{c}{\bfseries Type of Work}                                                                                                             \\ \hline
		\rowcolor{gray!40}
		{\bfseries Paper} & {\bfseries Cat.} & {\bfseries Venue} & {\bfseries CA} & {\bfseries SA}                                                                & \tabrotate{\bfseries Diagnosis Engine} & \tabrotate{\bfseries Query Generation} & \tabrotate{\bfseries Query Selection} & \tabrotate{\bfseries DPI}          & \tabrotate{\bfseries Meta Information} & \tabrotate{\bfseries Logical Reasoner} & \tabrotate{\bfseries Oracle}       & \tabrotate{\bfseries Methodology/Algo.} & \tabrotate{\bfseries Application}  & \tabrotate{\bfseries Explan./Didact./Survey} & \tabrotate{\bfseries Empirical} & \tabrotate{\bfseries Theoretical}  & \tabrotate{\bfseries Critical}     & \tabrotate{\bfseries Tool}         \\ 
		\hline
		\cite{rodler_memory-limited_2022} & A & Artificial Intelligence  & $\checkmark$ & $\checkmark$                                             & $\checkmark$                                  &                  &                 &              &                  &                  &              & $\checkmark$                &              &                             &  $\checkmark$                    & $\checkmark$ &              &              \\ 
		\cite{rodler_one_2022} & A & Knowledge-Based Systems   & $\checkmark$ & $\checkmark$                                             &                                               & $\checkmark$     & $\checkmark$    &              &                  &                  & $\checkmark$ & $\checkmark$                & $\checkmark$ &                             & $\checkmark$                     & $\checkmark$ & $\checkmark$ &              \\
		\cite{rodler_dynamichs_2022} & A & Information Sciences    & $\checkmark$ & $\checkmark$                                               & $\checkmark$                                  &                  &                 &  $\checkmark$            &                 & $\checkmark$                 &              & $\checkmark$                &              &                             & $\checkmark$                     & $\checkmark$ &              &              \\
		\cite{rodler_formal_2022} & A & Artificial Intelligence Review   & $\checkmark$ & $\checkmark$                                      & $\checkmark$                                  &                  &                 &              &                  & $\checkmark$     &              &                             &              & $\checkmark$                &                      & $\checkmark$ &              &              \\
		\cite{rodler_sequential_2022} & (A) & Artificial Intelligence (*) & $\checkmark$ & $\checkmark$                                           &                                               & $\checkmark$     & $\checkmark$    &              &                  &                  &              & $\checkmark$                &              &                             &  $\checkmark$                    & $\checkmark$ &              &              \\
		\cite{hofer_modeling_2022} & (A) & Artificial Intelligence (**)      & $\checkmark$ &                                      & $\checkmark$                                              &                  &                 & $\checkmark$ &                  &                  &              & $\checkmark$                & $\checkmark$ &                             & $\checkmark$         & $\checkmark$ &              &              \\
		\cite{rodler_random_2022} & A & AAAI      & $\checkmark$ & $\checkmark$                     & $\checkmark$                                  &                  &                 &              &  $\checkmark$                &                  &              & $\checkmark$                &              &                             & $\checkmark$         &              & $\checkmark$ &              \\
		\cite{rodler_how_2022} & C & DX    & $\checkmark$ & $\checkmark$                     & $\checkmark$                                  &                  &                 &              &                  &                  &              &  $\checkmark$                           &              & $\checkmark$                &                      &  $\checkmark$            &              &  $\checkmark$            \\
		\cite{rodler_randomized_2021} & A & KR  & $\checkmark$ &            & $\checkmark$                                  &                  &                 &              &                  & $\checkmark$     &              & $\checkmark$                & $\checkmark$ &                             &  $\checkmark$                    &  $\checkmark$            &              &              \\
		\cite{rodler_reuse_2020} & A & ECAI  & $\checkmark$ & $\checkmark$                     &  $\checkmark$                                             &                  &                 &  $\checkmark$            &                  &  $\checkmark$                &              & $\checkmark$                &              &                             & $\checkmark$                     & $\checkmark$             &              &              \\
		\cite{rodler_are_2019-1} & A & Knowledge-Based Systems    & $\checkmark$ &                                             & $\checkmark$                                  & $\checkmark$     & $\checkmark$    & $\checkmark$ & $\checkmark$     &                  & $\checkmark$ & $\checkmark$                & $\checkmark$ &                             & $\checkmark$         &              & $\checkmark$ & $\checkmark$ \\
		\cite{rodler_statichs_2018} & A & SoCS        & $\checkmark$ &                      & $\checkmark$                                  &                  &                 & $\checkmark$ &                  &                  &              & $\checkmark$                &              &                             &  $\checkmark$                    & $\checkmark$ &              &              \\
		\cite{rodler_impact_2018} & A & RuleML+RR      & $\checkmark$ &              &                                               &      & $\checkmark$    &  & $\checkmark$     &                  &              &                             &              &                             & $\checkmark$         &              &              &              \\
		\cite{schekotihin_ontodebug_2018} & B & FoIKS  & $\checkmark$ &  & $\checkmark$                                  & $\checkmark$     & $\checkmark$    & $\checkmark$ & $\checkmark$     &                  & $\checkmark$             &                             & $\checkmark$ & $\checkmark$                &                      &              &              & $\checkmark$ \\
		\cite{rodler_active_2017} & C & DX  & $\checkmark$ & $\checkmark$                       &                                               &                  & $\checkmark$    &              & $\checkmark$     &                  &              & $\checkmark$                &              & $\checkmark$                &   $\checkmark$                   & $\checkmark$ &              &              \\
		\cite{rodler_reducing_2017} & C & DX  & $\checkmark$ &                        &                                               &                  &                 & $\checkmark$ &                  &                  &              &                             &              &                             &                      & $\checkmark$ &              &     \\
		\hline        
	\end{tabular}
\caption[Overview of the research works included in this thesis]{Overview of the research works included in this thesis (\emph{left table sector}), based on 
the aspect of a generic diagnosis system \emph{mainly} addressed in the paper (\emph{middle table sector}) and the paper type (\emph{right table sector}). Meaning of the columns:\\
	(\emph{Venue}) quotes the publication venue of the respective paper, where journal names are given in full while conference names are abbreviated to keep the table size manageable (the reader is kindly asked to refer to the bibliography for the full names); a single asterisk \emph{(*)} means that the work is currently under revision, and the double asterisk \emph{(**)} signifies that the work is currently under review. 	\\
	(\emph{Cat.}) gives the rating of the publication venue of the paper (according to its classification in the expert-curated
	public research database ``FoDok'' of the University of Klagenfurt), where ``A'' refers to top-tier journals (Q1 on Scimago Journal \& Country Rank) or top-tier international conferences (acceptance rate below $0.33$), ``B'' means good journals (Q2 on Scimago Journal \& Country Rank) or good international conferences (acceptance rate below $0.66$), while all works not falling in categories A or B are classified by ``C''; the rating is parenthesized for not yet published works.\\
	(\emph{CA}) indicates whether the author of this thesis is a corresponding author of the respective paper.\\
	(\emph{SA}) marks whether the author of this thesis is the single author of the respective work.}
\label{tab:research_works_incl_in_thesis_summary}
\end{table}

\clearpage
\newpage\null\thispagestyle{empty}\newpage

\chapter{Our Research: Details}
\label{chap:research_details}

\begin{figure}[!tp]
	\centering
	\includegraphics[trim={6cm 0 0 0},clip,width=0.9\linewidth]{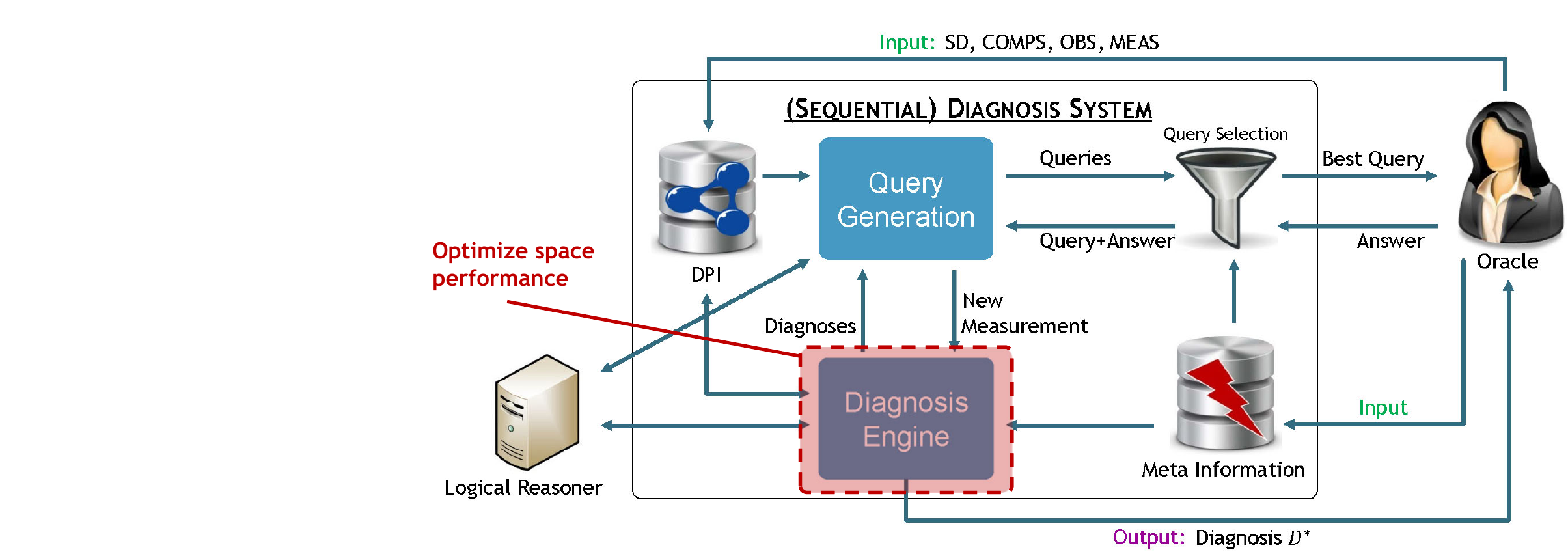}
	\caption[Memory-limited diagnosis computation]{Memory-limited diagnosis computation: Addressed system aspects and goals.}
	\label{fig:memory-limited}
\end{figure}

\begin{figure}
	\centering
	\includegraphics[trim={5.7cm 0 0 0},clip,width=0.9\linewidth]{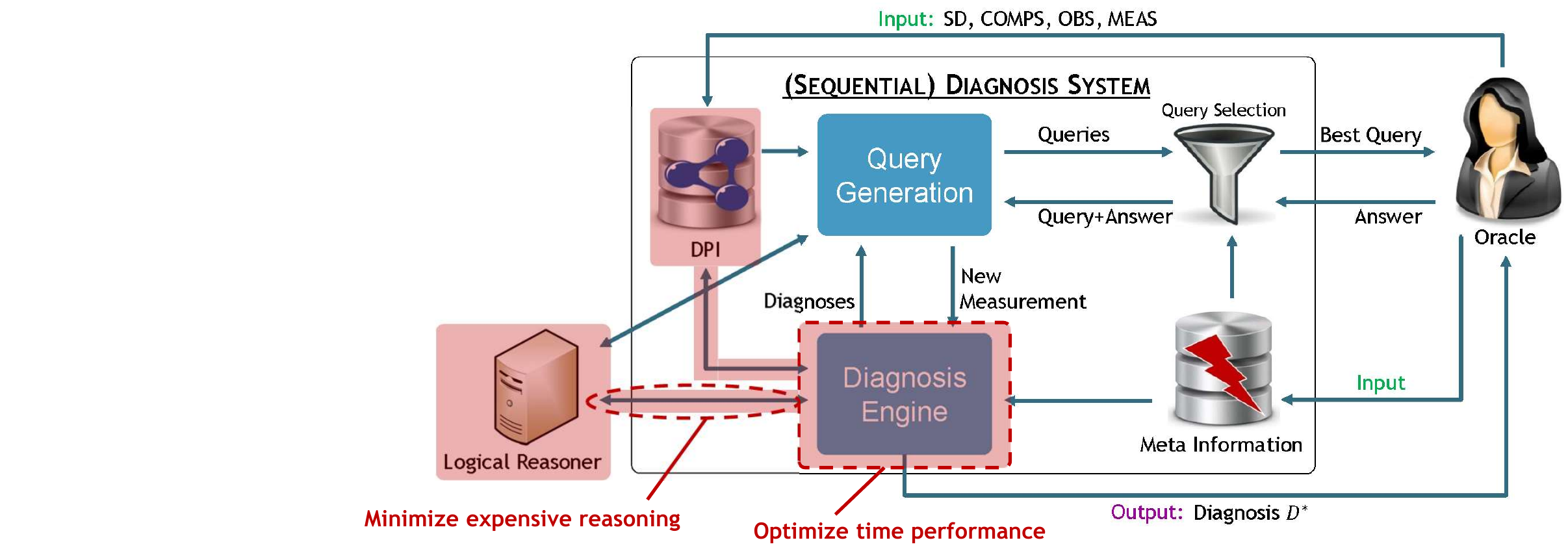}
	\caption[DynamicHS---Streamlining Reiter's hitting-set tree for sequential diagnosis]{DynamicHS---Streamlining Reiter's hitting-set tree for sequential diagnosis: Addressed system aspects and goals.}
	\label{fig:dynHS}
\end{figure}

\begin{figure}
	\centering
	\includegraphics[trim={6cm 0 0 0},clip,width=0.9\linewidth]{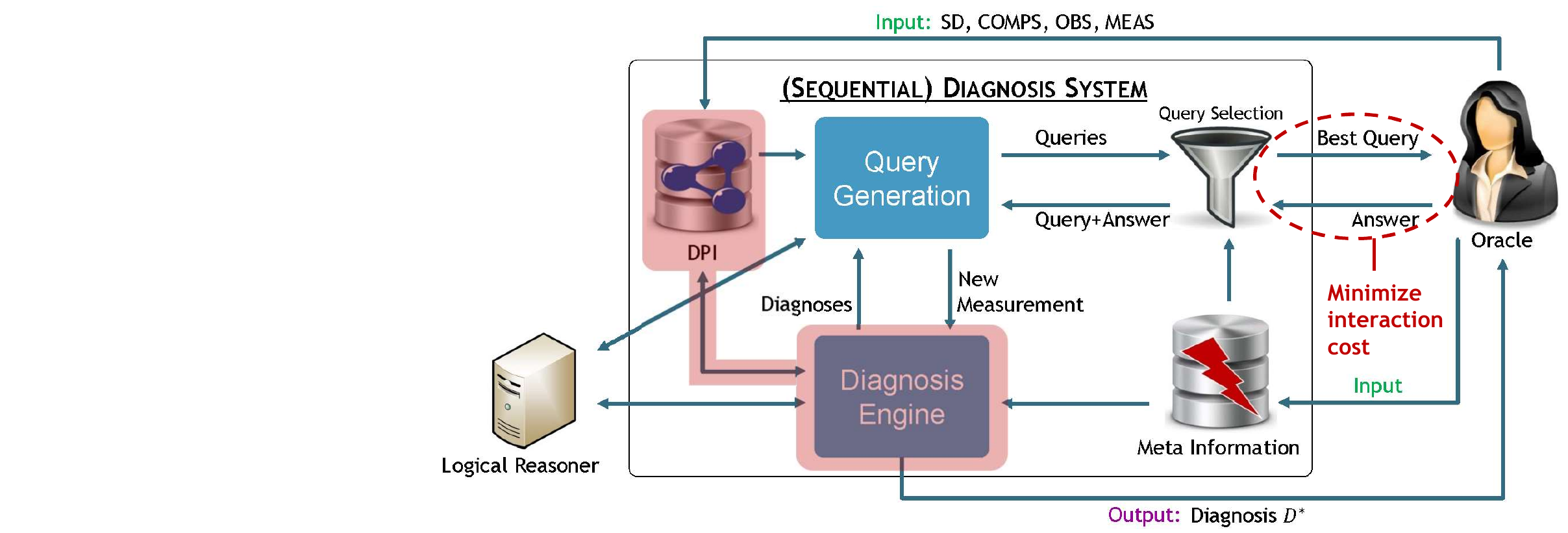}
	\caption[StaticHS---Reducing costs by solving a generalized sequential diagnosis problem]{StaticHS---Reducing user interaction costs by solving a generalized sequential diagnosis problem: Addressed system aspects and goals.}
	\label{fig:statHS}
\end{figure}

\begin{figure}
	\centering
	\includegraphics[trim={5.7cm 0 0 0},clip,width=0.9\linewidth]{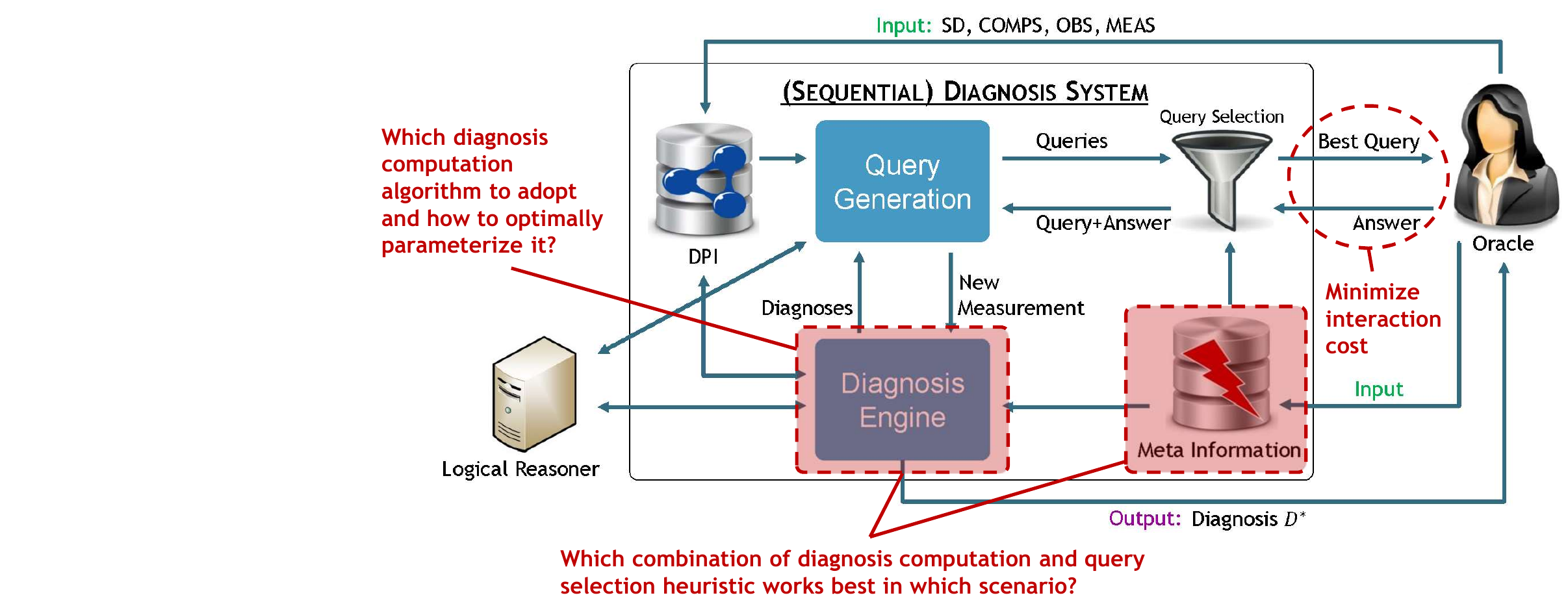}
	\caption[On the impact of diagnosis computation strategies on diagnostic decision making]{On the impact of diagnosis computation strategies on diagnostic decision making: Addressed system aspects and goals.}
	\label{fig:random_vs_bestfirst}
\end{figure}

\begin{figure}[!t]
	\centering
	\includegraphics[trim={4.8cm 0 0 0},clip,width=0.9\linewidth]{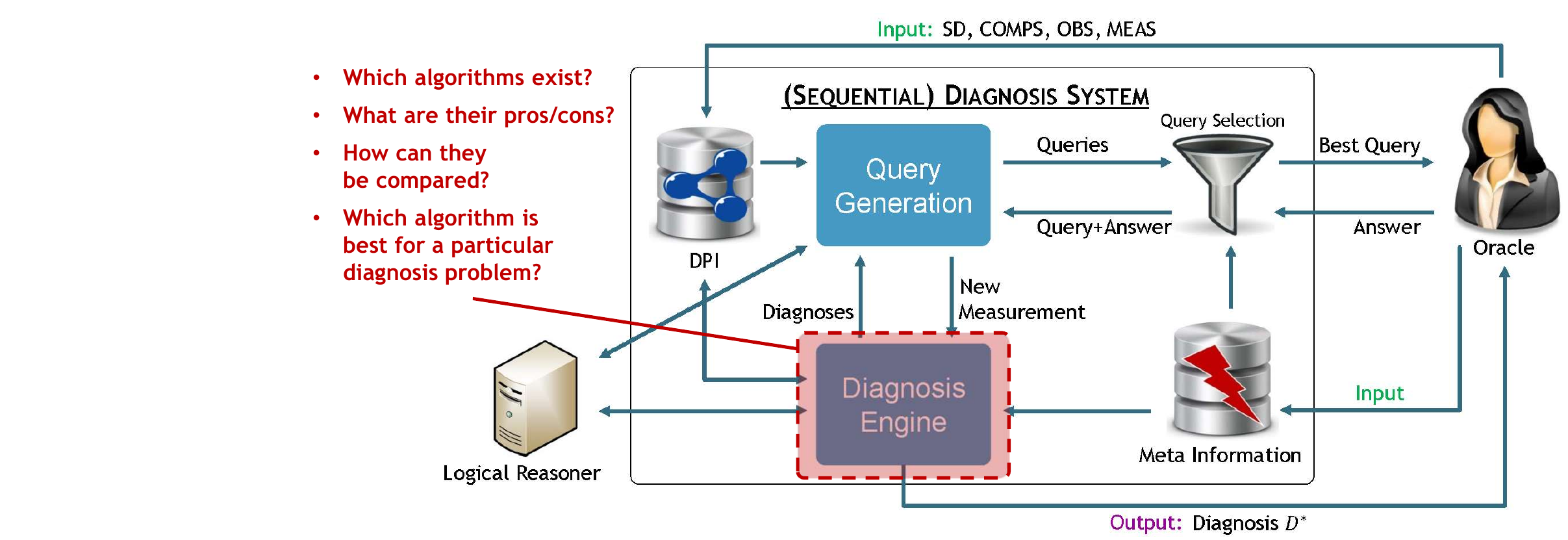}
	\caption[A taxonomy and classification of diagnosis computation algorithms]{A taxonomy and classification of diagnosis computation algorithms: Addressed system aspects and goals.}
	\label{fig:how_should_compute_candidates}
\end{figure}

\begin{figure}
	\centering
	\includegraphics[trim={5cm 0 0 0},clip,width=0.9\linewidth]{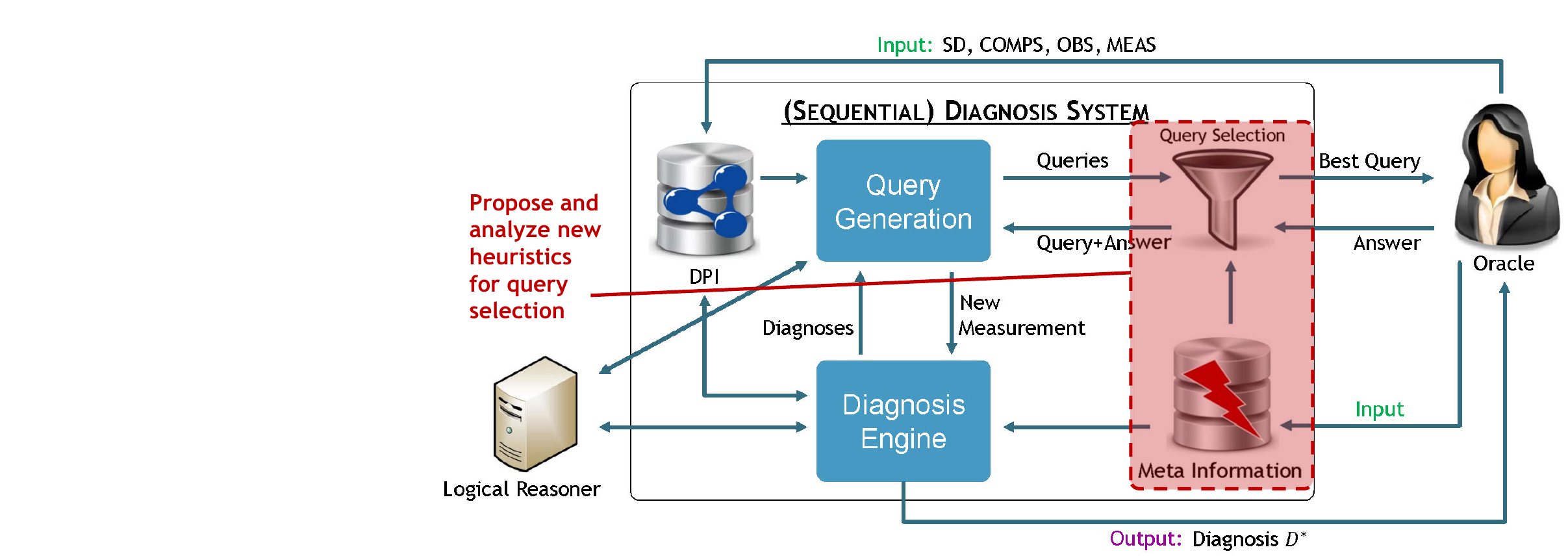}
	\caption[Proposal and theoretical analysis of query selection heuristics]{Proposal and theoretical analysis of query selection heuristics: Addressed system aspects and goals.}
	\label{fig:active_learning_in_SD}
\end{figure}

\begin{figure}
	\centering
	\includegraphics[trim={6.3cm 0 0 0},clip,width=0.85\linewidth]{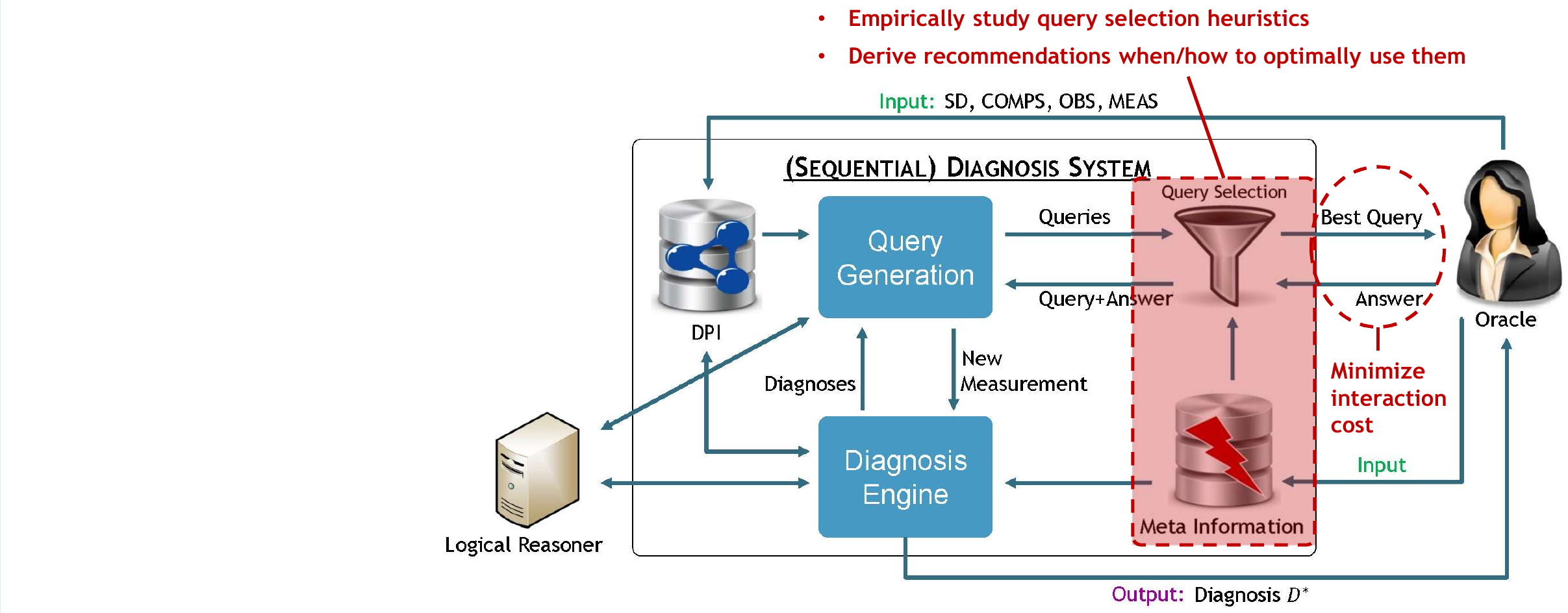}
	\caption[Empirical analysis of query selection heuristics]{Empirical analysis of query selection heuristics: Addressed system aspects and goals.}
	\label{fig:impact_proper_use_of_heuristics}
\end{figure}

\begin{figure}
	\centering
	\includegraphics[trim={6.3cm 0 0 0},clip,width=0.85\linewidth]{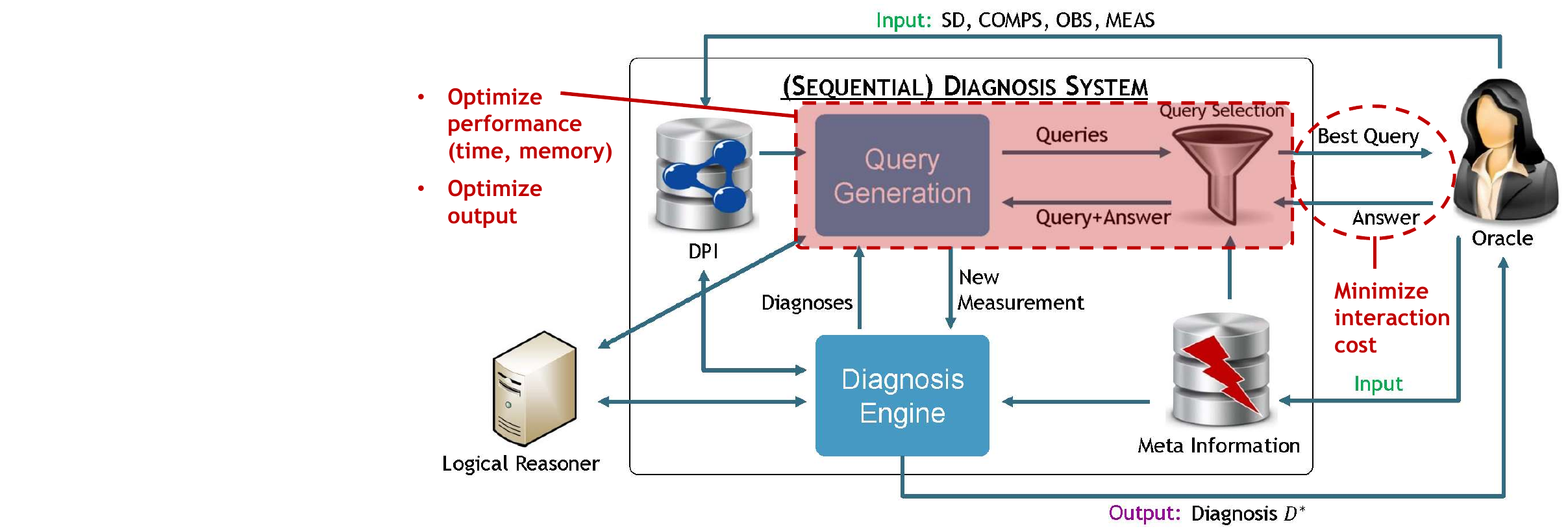}
	\caption[Efficient query computation and selection by systematic search]{Efficient query computation and selection by systematic search: Addressed system aspects and goals.}
	\label{fig:seq_diag_by_systematic_search}
\end{figure}

\begin{figure}[!tp]
	\centering
	\includegraphics[trim={5cm 0 0 0},clip,width=0.9\linewidth]{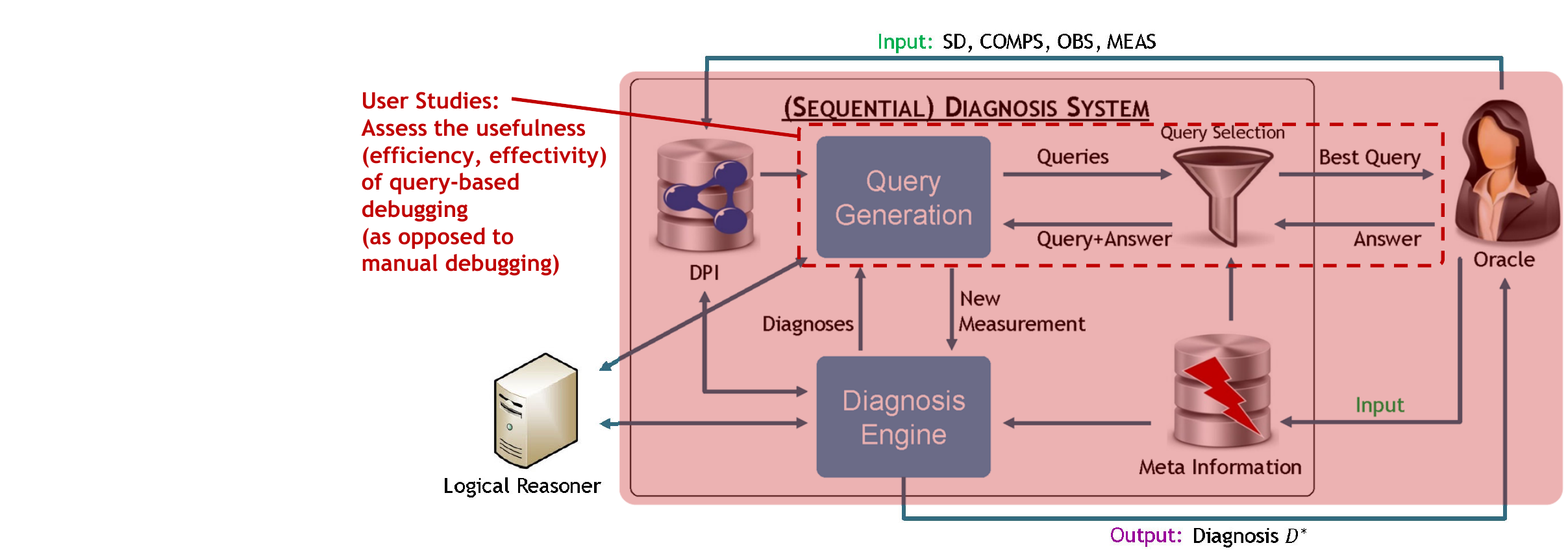}
	\caption[User study]{User study: Addressed system aspects and goals.}
	\label{fig:user_study}
\end{figure}

\begin{figure}[!tp]
	\centering
	\includegraphics[trim={5cm 0 0 0},clip,width=0.9\linewidth]{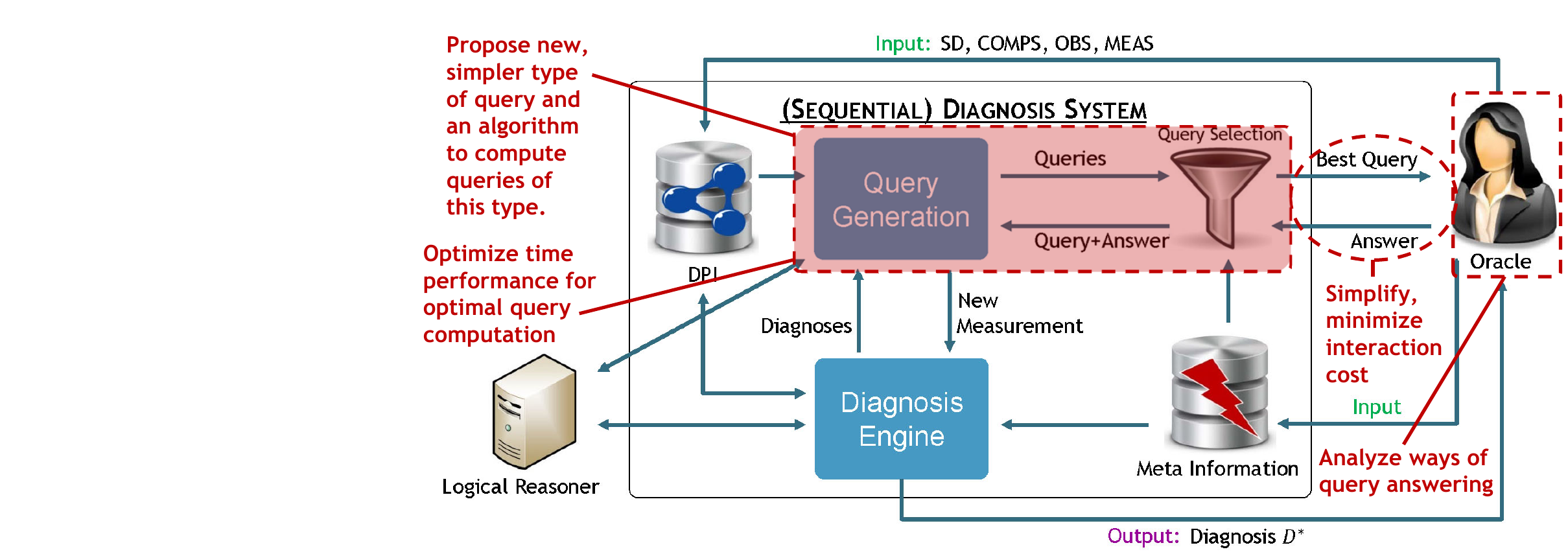}
	\caption[How to make query-based debugging simpler and more efficient]{On user types, assumptions, optimization criteria, and how to make query-based debugging simpler and more efficient: Addressed system aspects and goals.}
	\label{fig:one_step_at_a_time}
\end{figure}

\begin{figure}
	\centering
	\includegraphics[trim={5cm 0 0 0},clip,width=0.9\linewidth]{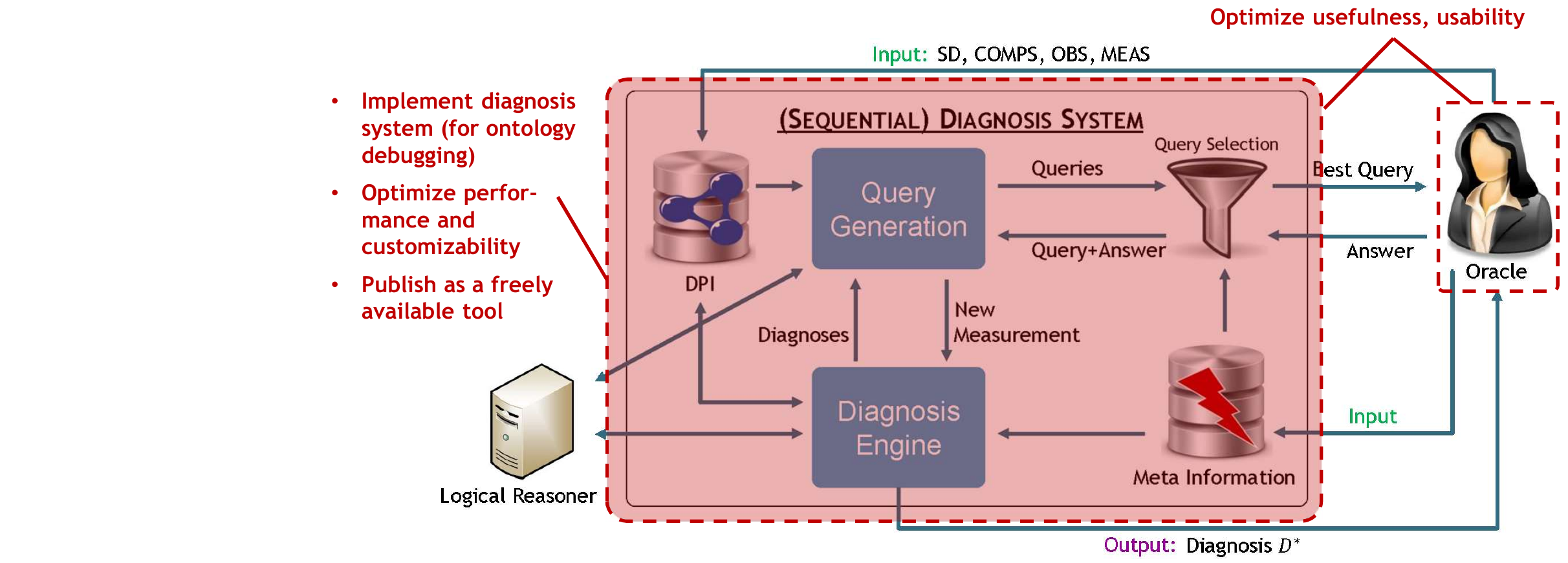}
	\caption[OntoDebug---A full-fledged ontology debugging tool]{OntoDebug---A full-fledged ontology debugging tool: Addressed system aspects and goals.}
	\label{fig:ontodebug}
\end{figure}

\begin{figure}
	\centering
	\includegraphics[trim={5cm 0 0 0},clip,width=0.9\linewidth]{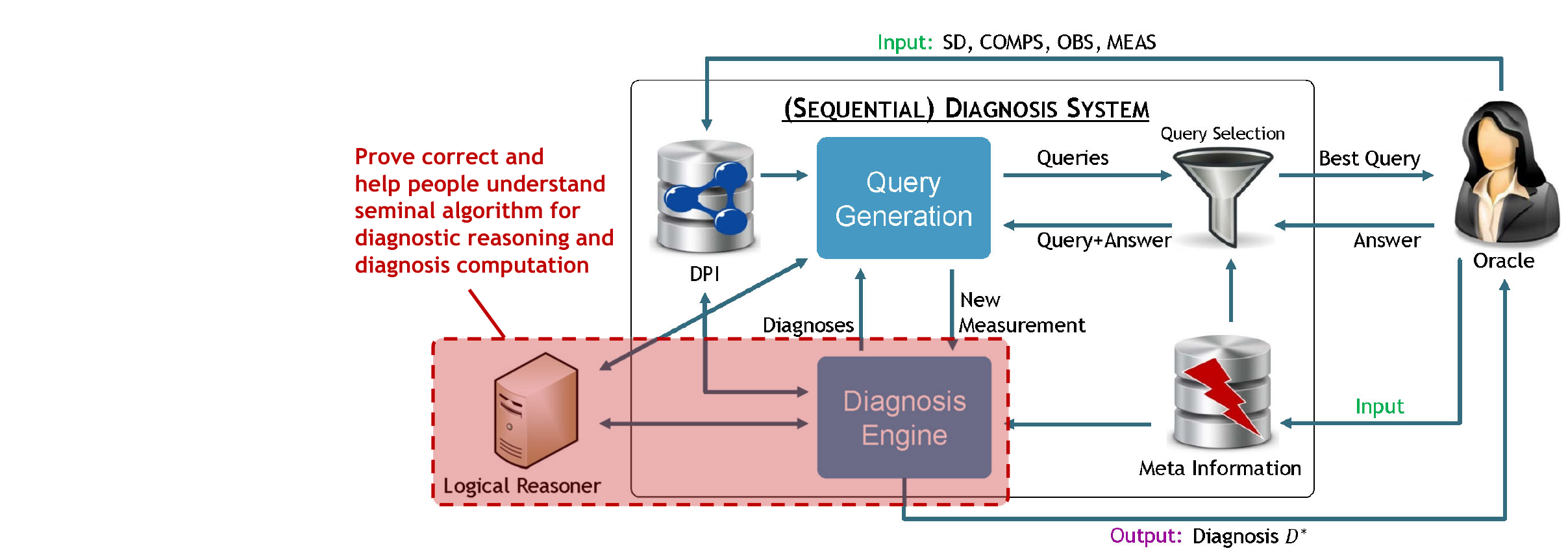}
	\caption[A formal proof and simple explanation of the QuickXplain algorithm]{A formal proof and simple explanation of the QuickXplain algorithm: Addressed system aspects and goals.}
	\label{fig:QX}
\end{figure}

\begin{figure}
	\centering
	\includegraphics[trim={5cm 0 0 0},clip,width=0.9\linewidth]{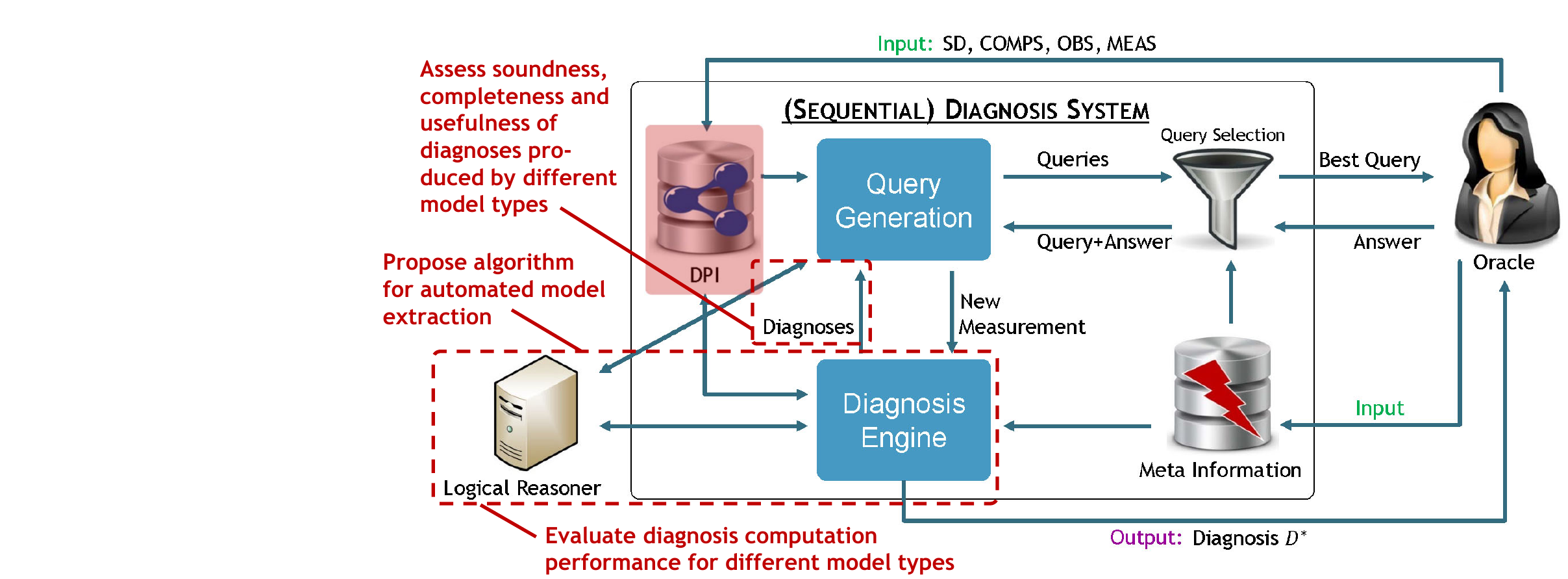}
	\caption[On modeling techniques for spreadsheet debugging]{On modeling techniques for spreadsheet debugging: Addressed system aspects and goals.}
	\label{fig:spreadsheet_debugging}
\end{figure}

\begin{figure}
	\centering
	\includegraphics[trim={5cm 0 0 0},clip,width=0.9\linewidth]{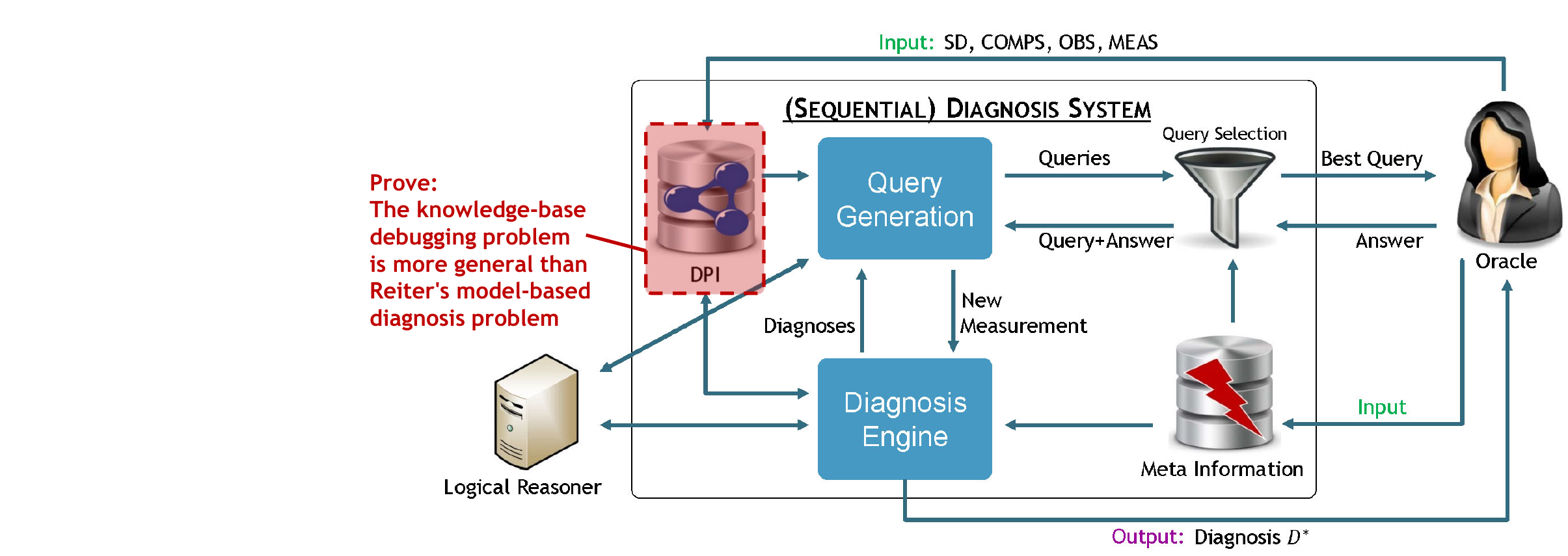}
	\caption[Reducing model-based diagnosis to knowledge base debugging]{Reducing model-based diagnosis to knowledge base debugging: Addressed system aspects and goals.}
	\label{fig:reducing_MBD_to_KBD}
\end{figure}

\begin{minipage}{0.75\textwidth}
In this chapter, we give some more details about the research papers included in this thesis. We thematically subdivide the discussion into five topics, each discussed in one of the Sections~\ref{chap:diag_computation}--\ref{chap:diagnostic_modeling}. Each of these sections is organized into subsections, one for each publication. The description of each paper is structured into three parts. 
First, we provide one figure for each publication which illustrates the aspects of the generic diagnosis system addressed 
as well as the research goals aimed at (cf.\ Figures~\ref{fig:memory-limited}--\ref{fig:reducing_MBD_to_KBD}). The legend pertinent to these figures is shown in Fig.~\ref{fig:legend}. 
Second, we treat its contents, where we briefly review the paper's highlights 
by sketching the paper's motivation, contributions as well as results.\footnotemark{} Third, we provide some meta information about each work including the publication venue, its scientific classification\footnotemark{} (cf.\ Tab.~\ref{tab:research_works_incl_in_thesis_summary}), the contributions of the author of this thesis to the paper as well as other publications on the topic where the author of this thesis was involved.  
All papers covered in this chapter can be found in Appendix~\ref{apx:papers_incl_in_thesis}.
\end{minipage}
\hfill
\begin{minipage}{0.22\textwidth}
	\centering 
	\includegraphics[width=0.9\textwidth]{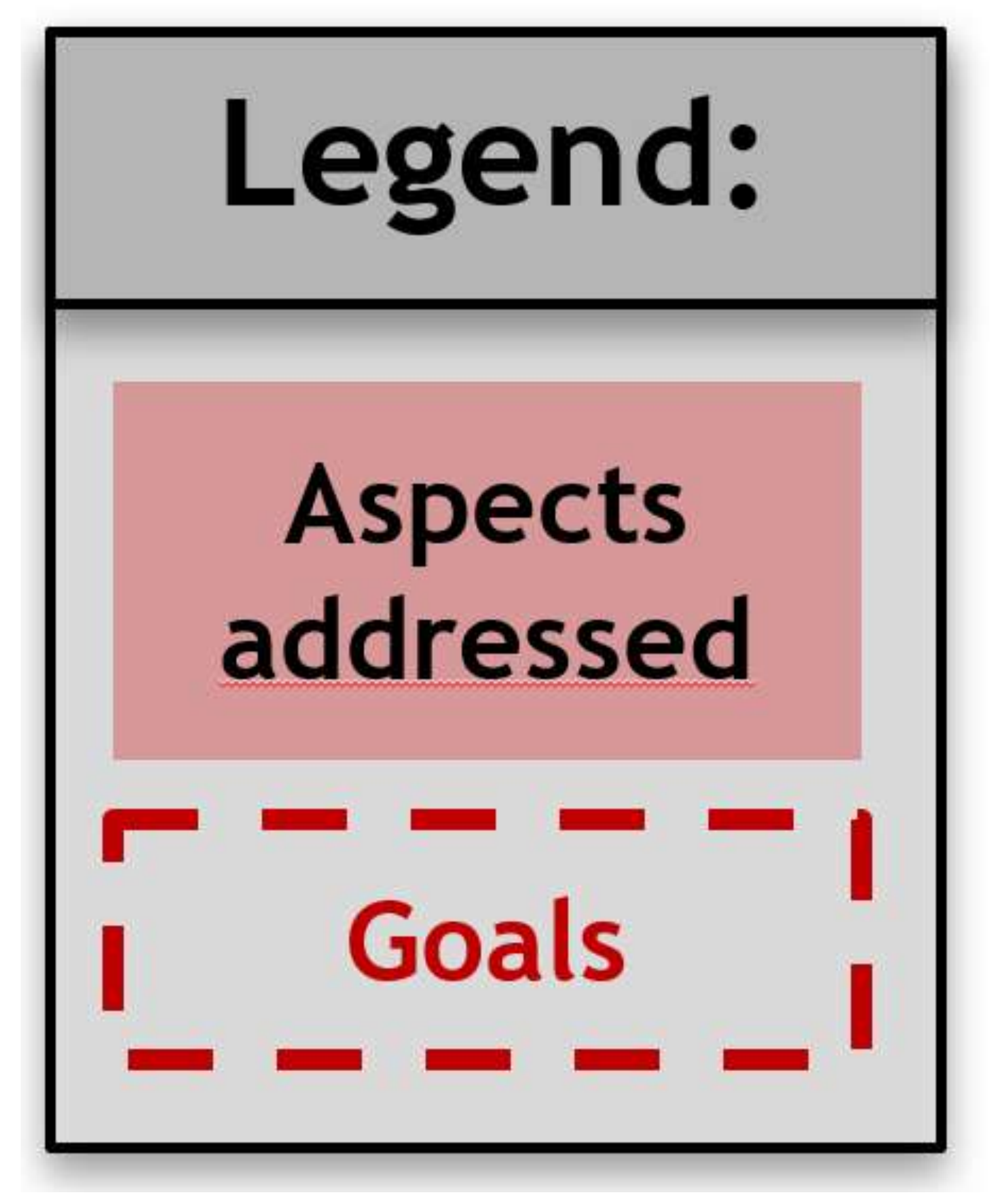}
	\captionof{figure}{Meaning of the accentuations used in Figures~\ref{fig:memory-limited}--\ref{fig:reducing_MBD_to_KBD}.}
	\label{fig:legend} 
\end{minipage} 
\footnotetext[6]{Please find animated presentations of some of the discussed works on \url{http://isbi.aau.at/ontodebug/publications}.}
\footnotetext{For each journal publication, we provide the journal's quartile in the Scimago Journal \& Country Rank as well as its Scimago Journal Rank Indicator (cf.\ \url{https://www.scimagojr.com/}). To generate the ranking (for the year 2021), we used ``Computer Science'' as subject area, and ``Artificial Intelligence'' as subject category.}

\section{Diagnosis Computation}
\label{chap:diag_computation}

\subsection{Memory-Limited Diagnosis Computation}
\label{sec:memory-limited_MBD}

\begin{description}[noitemsep,topsep=0pt]
	\item[Publication: System Aspects and Goals Addressed] See Fig.~\ref{fig:memory-limited}.
\end{description}
\vspace{-7pt}

\subsubsection*{Publication: Contents in a Nutshell}


\noindent\emph{Motivation:} 
Various model-based diagnosis scenarios require the computation of the most preferred minimal diagnoses. These diagnoses can be helpful, e.g., for a reasonable decision making and a well-founded early termination in the course of a sequential diagnosis process \cite{rodler_random_2022}. Existing algorithms that are sound (i.e., output only minimal diagnoses) and complete (i.e., can return all minimal diagnoses), however, require exponential space to achieve this task. This can prevent the application of these techniques to diagnosis settings which, e.g., pose substantial memory requirements on the diagnostic methods, or suffer from too little memory.\vspace{4pt}

\noindent\emph{Contributions:}
To enable successful diagnosis both on memory-restricted devices and for memory-intensive problem cases, we propose the RBF-HS algorithm for diagnosis computation which combines two seminal search methods, Reiter's hitting set computation strategy HS-Tree \cite{reiter_theory_1987} and Korf's path-finding technique Recursive Best-First Search (RBFS) \cite{korf_linear-space_1993}. 
RBF-HS is the \emph{first} diagnosis search that is sound, complete, best-first, generally applicable (to any diagnosis problem as per Reiter's theory \cite{reiter_theory_1987}), \emph{and linear-space}. In addition, we present 
HBF-HS, a hybrid between RBF-HS and HS-Tree, the idea behind which is to find a trade-off between runtime optimization and a restricted space consumption that does not exceed the available memory. Notably, both suggested approaches are not restricted to diagnosis problems, but applicable to best-first hitting set computation in general, which is relevant to various research and application domains beyond the frontiers of model-based diagnosis (cf., e.g., \cite{slaney_set-theoretic_2014}).\vspace{4pt}
%


\noindent\emph{Results:}
We conducted comprehensive experiments on real-world benchmarks where we compared RBF-HS and HBF-HS to HS-Tree, the most popular algorithm with the same properties (soundness, completeness, best-first property, general applicability). First, for the task of computing minimum-cardinality diagnoses, we found that
%
\begin{itemize}[noitemsep,topsep=0pt]
	\item in most cases, RBF-HS substantially reduces memory requirements by up to several orders of magnitude while exhibiting no more than marginal runtime losses, if any.  
	\item in more than a third of the cases, RBF-HS even achieved \emph{both} runtime \emph{and} memory improvements. E.g., in one case we observed $65\,\%$ runtime savings coupled with a $99.9\,\%$ memory reduction; in another case, HS-Tree ran out of memory (32\,GB) while RBF-HS never required the storage of more than the comparably negligible number of 125 tree nodes.
	\item whenever the runtime of RBF-HS was significantly higher than that of HS-Tree, the use of HBF-HS could level this overhead while still reasonably limiting the used memory.
	\item RBF-HS scales to large numbers of computed diagnoses and to problems involving high-cardinality minimal diagnoses.
\end{itemize} 
Second, when 
computing minimal diagnoses in descending order of probability, we observe that RBF-HS tends to trade memory savings more or less one-to-one for runtime overheads (which has well-understood theoretical reasons we discuss in the paper). Again, HBF-HS proves to be a reasonable remedy to cut down the runtime while complying with practicable memory bounds.

\subsubsection*{Publication: Meta Information}

\begin{tabular}{@{}lp{9.5cm}}
	\emph{Publication included in Thesis:} & \cite{rodler_memory-limited_2022} \\
	\emph{Authors:} & Patrick Rodler
	\\
	\emph{Title:} &  Memory-limited model-based diagnosis
	\\
	\emph{Year:} & 2022
	\\
	\emph{Publication Venue (Type):} & Artificial Intelligence \quad (Journal)
	\\
	\emph{Venue Metrics:} & A \qquad\quad (Scimago: Q1, Journal Rank Indicator 1.673)
	\\
	\emph{Own Contributions:} & Single-author publication
	\\
	\emph{Other Publications on the Topic:} & \cite{rodler_linear-space_2021} (Int'l Symposium on Combinatorial Search) \newline 
	\cite{rodler_sound_2020} (Int'l Workshop on Principles of Diagnosis) \newline
	\cite{rodler_rbf-hs_2022} (Technical report)
\end{tabular}

\subsection{DynamicHS:\ Optimizing Reiter's Hitting-Set Tree for Sequential Diagnosis}
\label{sec:dynHS}

\begin{description}[noitemsep,topsep=0pt]
	\item[Publication: System Aspects and Goals Addressed] See Fig.~\ref{fig:dynHS}.
\end{description}
\vspace{-7pt}

\subsubsection*{Publication: Contents in a Nutshell}


\noindent\emph{Motivation:}
To reason about the best next query, sequential diagnosis methods usually require a sample of diagnoses at each step of the iterative diagnostic process. The computation of this sample can be accomplished by various diagnostic search algorithms \cite{rodler_how_2022}. Among those, Reiter's HS-Tree \cite{reiter_theory_1987} is one of the most popular due to its desirable properties and general applicability. However, HS-Tree per-se does not encompass any specific provisions for being employed in an iterative way. Usually, it is used in a stateless fashion throughout the diagnosis process to (re)compute a sample of diagnoses per iteration, each time given the latest (updated) system knowledge including all so-far collected measurements. At this, the built search tree is discarded between two iterations, albeit often large parts of the tree have to be rebuilt in the next iteration, involving redundant operations and calls to costly reasoning services. Already Raymond Reiter, in his seminal\footnote{On Google Scholar, the paper boasts almost 4500 citations as of December 2022.} paper \cite{reiter_theory_1987} from 1987, pointed to the research problem of devising a reasonable strategy to reuse an existing hitting-set search tree to compute diagnoses after new system information is obtained. This very issue remained unanswered for decades.\vspace{4pt}  

\noindent\emph{Contributions:} Addressing Reiter's longstanding open question, we propose DynamicHS, a variant of HS-Tree that maintains state throughout the diagnostic session and embraces special strategies to minimize the number of expensive reasoner invocations.
The main objective of this new algorithm is to allow for more efficient computations in sequential diagnosis than HS-Tree while maintaining \emph{all} the advantages (general applicability, soundness, completeness, best-first property) of the latter.\vspace{4pt}

\noindent\emph{Results:} In extensive experiments, we used a corpus of real-world benchmarks to compare DynamicHS with HS-Tree in various sequential diagnosis scenarios (considering different numbers of diagnoses to be computed per iteration, and various employed query selection heuristics). Te main insights were:
\begin{itemize}[noitemsep,topsep=0pt]
	\item DynamicHS is superior to HS-Tree in terms of computation time in $99.4\,\%$ of the investigated diagnostic scenarios. Roughly, these savings are achieved by trading less time (fewer redundant operations and reasoner calls) for more space (statefulness), where the additional memory required by DynamicHS
	was reasonable in the vast majority of the scenarios; and, whenever HS-Tree was applicable in our experiments in terms of memory requirements, DynamicHS was so as well.
	\item The average runtime savings over HS-Tree achieved by DynamicHS are substantial and statistically significant in most scenarios, and reach median and maximal values of $52\,\%$ and $75\,\%$. That is, HS-Tree requires up to an average of four times the computation time of DynamicHS in the tested diagnosis scenarios. 
	\item In single diagnosis sessions, we observed that it took HS-Tree up to more than nine times as much time as DynamicHS---notably, while both methods always compute the same diagnostic solutions.
	\item The median runtime savings of DynamicHS per scenario appear to be neither dependent on the number of diagnoses computed nor on the measurement selection heuristic used.
	\item Considering the hardest cases per diagnosis scenario, which were up to one order of magnitude harder than the average cases, the time savings obtained by means of DynamicHS are even more substantial than on average, and reach median and maximal values of $64\,\%$ and $89\,\%$, respectively.
\end{itemize}

\subsubsection*{Publication: Meta Information}

\begin{tabular}{@{}lp{9.5cm}}
	\emph{Publication included in Thesis:} & \cite{rodler_dynamichs_2022} \\
	\emph{Authors:} & Patrick Rodler
	\\
	\emph{Title:} &  DynamicHS: Streamlining Reiter's hitting-set tree for \newline sequential diagnosis
	\\
	\emph{Year:} & 2022
	\\
	\emph{Publication Venue (Type):} & Information Sciences \quad (Journal)
	\\
	\emph{Venue Metrics:} & A \qquad\quad (Scimago: Q1, Journal Rank Indicator 2.290)
	\\
	\emph{Own Contributions:} & Single-author publication
	\\
	\emph{Other Publications on the Topic:} & \cite{rodler_reuse_2020} (European Conference on Artificial Intelligence) \newline
	\cite{rodler_too_2020} (Int'l Symposium on Combinatorial Search) \newline
	\cite{rodler_dynamichs_2020} (Int'l Workshop on Principles of Diagnosis) \newline
	\cite{rodler_appendix_2021} (Technical report)
\end{tabular}

\subsection[StaticHS: Reducing Costs by Solving a Generalized Sequential Diagnosis Problem]{StaticHS: Reducing User Interaction Costs by Solving a Generalized Sequential Diagnosis Problem}
\label{sec:staticHS}

\begin{description}[noitemsep,topsep=0pt]
	\item[Publication: System Aspects and Goals Addressed] See Fig.~\ref{fig:statHS}.
\end{description}
\vspace{-7pt}

\subsubsection*{Publication: Contents in a Nutshell}


\noindent\emph{Motivation:}
In sequential diagnosis, new measurements are successively added to an initially given DPI. Each such acquisition of new information leads to a change of the solution space of minimal diagnoses. In fact, as we touched upon in Example~\ref{ex:circuit_seqDiag}, each minimal diagnosis for the newly obtained DPI is
either equal to or a proper superset of some minimal diagnosis for the original DPI. In other words, any ``new'' minimal diagnosis emerging throughout the sequential diagnosis process assumes faulty strictly more components than some initial minimal diagnosis. 
In many real-world applications such as physical devices, however, components are usually much more likely to be normal than at fault (at a given point in time). Thus, there is a high chance of the actual diagnosis (pinpointing the actually faulty components) being among the minimal diagnoses for the original input DPI. In such scenarios, one might---as a first step---only want to explore the initial diagnosis solution space, and neglect all ``new'' minimal diagnoses arising after DPI transitions.\vspace{4pt}

\noindent\emph{Contributions:} We suggest an alternative interpretation of the sequential diagnosis problem, called StatSD, where new measurements are not used to formulate a new DPI (as in the standard interpretation that we refer to as DynSD for disambiguation in this section), but only as constraints on the initial solution space of minimal diagnoses. The goal is to continue gathering information until only one (sufficiently likely) minimal diagnosis remains among the solutions of the initial DPI (and not the current DPI including all collected measurements as in DynSD).
As a study of the literature revealed, existing sequential diagnosis methods focus only on DynSD. 

As a consequence, we propose StaticHS, a novel sound and complete diagnosis search that can generate minimal diagnoses in best-first order, is suitable to solve both StatSD and DynSD, and is as generally applicable as Reiter's HS-Tree \cite{reiter_theory_1987}. 
We theoretically prove that, if the actual diagnosis is among the minimal diagnoses for the originally given DPI, then for any set of oracle queries required to solve the DynSD problem (e.g., by means of HS-Tree), a subset of these queries already suffices to locate the actual diagnosis when the StatSD problem is considered and solved (e.g., by means of StaticHS) instead.
Hence, if we could ensure that the same queries are asked regardless of whether StatSD or DynSD is solved (which we generally cannot since the diagnoses available to guide query selection may be different for both problems), then solving StatSD would be proven more efficient than DynSD given the preconditions mentioned above. 

To solve DynSD, StaticHS will solve a series of StatSD problems to obtain the same solution diagnosis as any standard sequential diagnosis algorithm geared towards DynSD. At this, different strategies (influencing performance, not correctness) can be taken defining which specific StatSD problems are tackled. One such strategy is to use a new DPI whenever a new measurement is obtained, in which case StaticHS will act as any standard sequential diagnosis technique. Consequently, \emph{(1)}~StaticHS is a generalization of other techniques, such as HS-Tree, in the context of sequential diagnosis, and \emph{(2)}~solving multiple StatSD problems is a more general strategy than solving one DynSD problem.\vspace{4pt}

\noindent\emph{Results:} As somewhat expected based on 
the theoretical findings, empirical examinations using real-world problems revealed that StaticHS, compared to a standard sequential diagnosis algorithm addressing DynSD (in our tests w.l.o.g.\ implemented by HS-Tree), reduces the required number of queries substantially 
both when \emph{(a)}~tackling the StatSD problem suffices (actual diagnosis is among the initial minimal diagnoses) and when \emph{(b)}~the consideration of the DynSD problem is necessary (actual diagnosis is not among the initial minimal diagnoses). In numbers, the avg.\ / max.\ saved measurement effort for the oracle in scenarios (a) and (b) amounted to $20\,\%$ / $65\,\%$ and $21\,\%$ / $65\,\%$, respectively. Considering the individual executed sequential diagnosis sessions, the significant superiority of StaticHS as opposed to a method attacking DynSD is manifested in terms of a lower / same / higher number of queries in $76\,\%$ / $21\,\%$ / $3\,\%$ of the cases. These savings are a result
of StaticHS's ability to combine a search space reduction
(StatSD) with completeness (DynSD).
As to the average algorithm reaction time (oracle waiting time), StaticHS outperformed HS-Tree for a half of the benchmarks, while it was slower for the other half. However, whenever StaticHS exhibited a higher reaction time, which in absolute terms was always less than 5 sec, this was
compensated by very few queries StaticHS required to locate the actual diagnosis.

\subsubsection*{Publication: Meta Information}

\begin{tabular}{@{}lp{9.5cm}}
	\emph{Publication included in Thesis:} & \cite{rodler_statichs_2018} \\
	\emph{Authors:} & Patrick Rodler, Manuel Herold
	\\
	\emph{Title:} &  StaticHS: A variant of Reiter's hitting set tree for \newline efficient sequential diagnosis
	\\
	\emph{Year:} & 2018
	\\
	\emph{Publication Venue (Type):} & Int'l Symposium on Combinatorial Search \quad (Conference)
	\\
	\emph{Venue Metrics:} & A
	\\
	\emph{Own Contributions (est.):} & 
	Concept/Idea: $100\,\%$ \newline
	Theory/Algorithms: $100\,\%$ \newline
	Development/Implementation: $30\,\%$ \newline
	Evaluation: $80\,\%$ \newline
	Manuscript: $100\,\%$
	\\
	\emph{Other Publications on the Topic:} & \cite{rodler_reducing_2018} (Int'l Workshop on Principles of Diagnosis)
\end{tabular}

\subsection{On the Impact of Diagnosis Computation Strategies \\ on Diagnostic Decision Making}
\label{sec:random_vs_best-first}

\begin{description}[noitemsep,topsep=0pt]
	\item[Publication: System Aspects and Goals Addressed] See Fig.~\ref{fig:random_vs_bestfirst}.
\end{description}
\vspace{-7pt}

\subsubsection*{Publication: Contents in a Nutshell}


\noindent\emph{Motivation:}
Statistical samples, in order to be representative, have to be drawn from a population in a random and unbiased way. Nevertheless, it is common practice in the field of model-based diagnosis to make estimations from (biased) best-first samples. One example is the computation of a few most probable diagnoses for a defective system and the use of these to assess which aspect of the system, if measured, would bring the highest information gain.\vspace{4pt}

\noindent\emph{Contributions:}
In this work, we scrutinize whether these statistically not well-founded conventions, that both diagnosis researchers and practitioners have adhered to for decades, are indeed reasonable. To this end, we empirically analyze six diagnosis computation methods for generating samples of minimal diagnoses: one best-first strategy yielding the most probable diagnoses, 
one outputting a random sample of diagnoses, and, as a baseline, a worst-first technique which returns the least probable diagnoses; moreover, we suggest an efficient heuristic method based on \cite{shchekotykhin_sequential_2014} that can be configured to produce approximations of the best-first, the random, and the worst-first samples. The particular focus of our evaluations is on the comparison between (statistically unfounded, but commonly used) best-first and (statistically well-founded, but not commonly adopted) random sampling approaches. 
Specifically, we
\begin{itemize}[noitemsep,topsep=0pt]
	\item study the samples' \emph{(a)}~``theoretical'' representativeness (``accuracy'') in terms of how well they allow to estimate features (eliminated diagnoses, answer probabilities) of queries taken into account by query selection heuristics to suggest the best next query,
	as well as their \emph{(b)}~``practical'' representativeness (``efficiency'') in terms of
	how many queries and how much time are needed to locate the fault in case the particular sample type is used,
	\item investigate the impact of the used sample size (number of computed diagnoses),
	\item examine the optimal trade-off between sampling efficiency (diagnosis computation time) and effectivity (effort until the actual diagnosis is located), and
	\item explore how the approximate sampling techniques (which do not guarantee a specific set of diagnoses to be computed) compare to the exact ones (which provably generate a precisely defined subset of all diagnoses).
\end{itemize}
\vspace{4pt}

\noindent\emph{Results:}
The bottom line of our investigations is: Somewhat surprisingly, the best-first samples including the most probable diagnoses commonly used in the field proved to be the best choice in a large fraction of the investigated cases. Yet, we also find that, for certain configurations of a diagnosis system, best-first samples imply drastic overheads compared to other sample types. Random samples, though enabling highly reliable estimations, often led to a worse diagnostic efficiency than biased ones. We discuss reasons for this phenomenon and make recommendations which configurations wrt.\ diagnosis computation algorithm, query selection heuristic and sample size users of diagnosis systems should adopt for best diagnostic performance. E.g., best-first samples are favorable for small sample sizes or when the information gain \cite{de_kleer_diagnosing_1987} or split-in-half \cite{shchekotykhin_interactive_2012} heuristics are used, whereas random ones are best for larger sample sizes or when adopting the risk optimization \cite{rodler_rio_2013} or most probable singleton \cite{rodler_active_2017} techniques for query selection. Further, our results suggest a time-information trade-off in diagnosis sampling, i.e., more efficient sampling tends to imply less effective queries. Finally, we find that the suggested approximate, and often efficient, sampling technique based on the Inv-HS-Tree algorithm \cite{shchekotykhin_sequential_2014} in many cases provides a good balance between sampling efficiency and diagnostic effectivity.

\subsubsection*{Publication: Meta Information}

\begin{tabular}{@{}lp{9.5cm}}
	\emph{Publication included in Thesis:} & \cite{rodler_random_2022} \\
	\emph{Authors:} & Patrick Rodler
	\\
	\emph{Title:} &  Random vs.\ best-first: Impact of sampling strategies on \newline decision making in model-based diagnosis
	\\
	\emph{Year:} & 2022
	\\
	\emph{Publication Venue (Type):} & AAAI Conference on Artificial Intelligence \quad (Conference)
	\\
	\emph{Venue Metrics:} & A
	\\
	\emph{Own Contributions:} & Single-author publication
	\\
	\emph{Other Publications on the Topic:} & \cite{rodler_we_2020} (Int'l Workshop on Principles of Diagnosis)
\end{tabular}

\subsection{A Taxonomy and Classification of Diagnosis Computation Algorithms}
\label{sec:how_should_I_compute_my_candidates}

\begin{description}[noitemsep,topsep=0pt]
	\item[Publication: System Aspects and Goals Addressed] See Fig.~\ref{fig:how_should_compute_candidates}.
\end{description}
\vspace{-7pt}

\subsubsection*{Publication: Contents in a Nutshell}


\noindent\emph{Motivation:}
Diagnosis computation is one of the most integral tasks in
model-based diagnosis as it allows to generate fault hypotheses, which are essential for both fault localization and
repair. Due to its generality, the model-based diagnosis formalism has been used to express and tackle debugging problems in a wide diversity of application areas, as expounded in Sec.~\ref{sec:application_areas}. This has led to a remarkable multitude and heterogeneity of the diagnosis computation methods proposed in the literature, which are often motivated by and tailored for application-specific requirements and problem
cases. As a result, it is a hard task for both researchers and practitioners to
\begin{itemize}[noitemsep,topsep=0pt]
	\item get an overview of existing approaches,
	\item identify the crucial properties of diagnostic techniques,
	\item assess the methods based on these properties, and
	\item choose the appropriate approach for a research- or
	application-related diagnostic task at hand.
\end{itemize}
\vspace{4pt}

\noindent\emph{Contributions:}
To account for this, 
we present a taxonomy for diagnosis computation algorithms (under the commonly used weak fault model setting, cf.\ Sec.~\ref{sec:MBD}). Specifically, we introduce and formally define a range of features which are arguably vital for a proper understanding, comparison, selection, and use of diagnostic techniques. We explain the influence of each feature on the proper selection of a diagnosis algorithm for a diagnostic task, discuss the potential impact of different feature manifestations on the performance of diagnosis algorithms, and examine relationships among the features.\vspace{4pt} 

\noindent\emph{Results:} To demonstrate the value and application of the proposed taxonomy, we provide a multi-dimensional assessment and categorization of over $30$ important diagnostic methods in the literature.

\subsubsection*{Publication: Meta Information}

\begin{tabular}{@{}lp{9.5cm}}
	\emph{Publication included in Thesis:} & \cite{rodler_how_2022} \\
	\emph{Authors:} & Patrick Rodler
	\\
	\emph{Title:} &  How should I compute my candidates? A taxonomy and classification of diagnosis computation algorithms
	\\
	\emph{Year:} & 2022
	\\
	\emph{Publication Venue (Type):} & Int'l Workshop on Principles of Diagnosis \quad (Conference)
	\\
	\emph{Venue Metrics:} & C
	\\
	\emph{Own Contributions:} & Single-author publication
	\\
	\emph{Other Publications on the Topic:} & \cite{rodler_dynamichs_2022} (Information Sciences)
\end{tabular}

\section{Query Computation and Selection}
\label{chap:query_computation+selection}

\subsection{Proposal and Theoretical Analysis of Query Selection Heuristics}
\label{sec:active_learning}

\begin{description}[noitemsep,topsep=0pt]
	\item[Publication: System Aspects and Goals Addressed] See Fig.~\ref{fig:active_learning_in_SD}.
\end{description}
\vspace{-7pt}

\subsubsection*{Publication: Contents in a Nutshell}


\noindent\emph{Motivation:}
Research in the field of Active Learning (AL) \cite{settles_active_2010} provides a range of diverse general 
heuristics targeting the optimization of hypotheses discrimination tasks. While traditionally and very fruitfully exploited in machine learning, e.g., for efficient text classification, image retrieval, concept learning, machine translation, or natural language processing, the key idea behind AL is that a learner can achieve greater accuracy with less newly collected information if the used training data can be adaptively chosen based on its current state of knowledge. At each iteration of the learning process, the active learner can consult an oracle, e.g. a human expert, to label any query from some predefined query space. The new information in terms of the query's label is then taken into account to update the learner's current knowledge state.
As this bears close resemblance to the generic information acquisition process pursued by sequential diagnosis systems (cf.\ Sections~\ref{sec:sequential_diagnosis} and \ref{sec:generic_diagnosis_system}), many AL strategies, termed Query Selection Measures (QSMs), are basically tailored for being used in the context of sequential diagnosis. QSMs are real-valued functions quantifying the quality of queries. However, only few of these AL QSMs, such as information entropy \cite{de_kleer_diagnosing_1987}, have so far been carried over to or employed in interactive diagnosis.\vspace{4pt}

\noindent\emph{Contributions:}
Motivated by these observations, the contributions of this work are the following: We
\begin{itemize}[noitemsep,topsep=0pt]
	\item investigate the relationship between AL and sequential diagnosis and discuss commonalities of and differences between both fields,
	\item survey all popular AL QSMs and adapt them for being used as heuristics in sequential diagnosis,
	\item define a plausible general discrimination preference order (DPO) on queries (formalizing the no-
	tion of the ``diagnoses discrimination power'' of queries),
	\item figure out superiority relationships between QSMs based on their compliance with the DPO, which suggests a preference order on QSMs, 
	\item harness 
	this preference order to make recommendations regarding the selection of a suitable QSM for a given diagnostic scenario,
	\item derive improved (parameterized) versions from some QSMs to overcome unveiled deficits (e.g., objectively suboptimal decisions) arising from idiosyncrasies of model-based diagnosis, 
	\item give equivalence classes of QSMs under various conditions (query spaces, QSM parametrizations),
	\item theoretically analyze QSM functions regarding their global optima and determine qualitative properties of optimal queries,
	\item show how these properties can be used to formulate heuristics and pruning techniques that allow us to design efficient systematic search procedures for 
	optimal queries wrt.\ any given QSM, 
	\item based on the systematic search technique devised, suggest a paradigm shift in sequential diagnosis from the commonly adopted \emph{pool-based query selection} strategy (where the best query is filtered out from a potentially large set of candidates) towards a \emph{query synthesis} approach (whose aim is to find a sufficiently good query while 
	exploring a minimal fraction of the query space in a goal-directed way based on heuristics). This novel approach can be especially promising for diagnostic problem cases (e.g., in knowledge base debugging \cite{rodler_interactive_2015}) where a pool of queries is neither explicitly given nor efficiently computable.
\end{itemize}
\vspace{4pt}

\noindent\emph{Results:} 
In experiments using a benchmark dataset of real-world diagnosis problems, we compared the suggested query synthesis (QS) against the commonly used pool-based query selection (PB) approach. The main insights (from preliminary results) were:
\begin{itemize}[noitemsep,topsep=0pt]
	\item On average, over all cases, it took QS for all investigated QSMs less than $1$ sec to achieve a query optimality (compared to the theoretically optimal QSM-value) of more than $99.999\,\%$ while 
	exploring only minor fractions of less than $4\,\%$ the complete query space.
	\item PB could not handle scenarios where much diagnostic evidence (i.e., a large sample of diagnoses, cf.\ Sec.~\ref{sec:random_vs_best-first}) is available for query selection. More precisely, while QS could easily and efficiently deal with up to $80$ given minimal diagnoses, PB could in most cases only process no more than $10$ ones within a timeout of an hour.
	\item PB consumed substantially more time than QS. Specifically, the former required minimally / on average / maximally $27$ / $787$ / $2528$ (!) times the query computation time of the latter. 
	By absolute numbers, the minimal / average / maximal PB execution time amounted to $6$ / $137$ / $566$ sec whereas QS never required more than $0.6$ sec for the same cases.   
\end{itemize}
Overall, these observations indicate both a very high efficiency and a very high query quality achieved by the proposed novel query computation paradigm, and let QS appear to be the method of choice (at least) for model-based diagnosis problems with query spaces of large size or implicit nature (where query candidates need to be expensively computed, e.g., by means of logical reasoning \cite{shchekotykhin_interactive_2012}).

\subsubsection*{Publication: Meta Information}

\begin{tabular}{@{}lp{9.5cm}}
	\emph{Publication included in Thesis:} & \cite{rodler_active_2017} \\
	\emph{Authors:} & Patrick Rodler
	\\
	\emph{Title:} &  On active learning strategies for sequential diagnosis
	\\
	\emph{Year:} & 2017
	\\
	\emph{Publication Venue (Type):} & Int'l Workshop on Principles of Diagnosis \quad (Conference)
	\\
	\emph{Venue Metrics:} & C
	\\
	\emph{Own Contributions:} & Single-author publication
	\\
	\emph{Other Publications on the Topic:} & \cite{rodler_towards_2016} (Technical report)
\end{tabular}

\subsection{Empirical Analysis of Query Selection Heuristics}
\label{sec:impact_of_heuristics_in_ontology_debugging}

\begin{description}[noitemsep,topsep=0pt]
	\item[Publication: System Aspects and Goals Addressed] See Fig.~\ref{fig:impact_proper_use_of_heuristics}.
\end{description}
\vspace{-7pt}

\subsubsection*{Publication: Contents in a Nutshell}


\noindent\emph{Motivation:}
A variety of query selection heuristics have been suggested for sequential diagnosis (cf.\ Sec.~\ref{sec:active_learning}), but it is unclear which heuristic to opt for given a particular diagnostic scenario. Questions of interest are, e.g.: Is some heuristic always superior to all others? On which factors does the 
performance of the particular heuristics depend? Under which circumstances should one use which heuristic?\vspace{4pt} 

\noindent\emph{Contributions:}
To give well-founded guidance to users of debugging systems and to bring light to these and other questions, 
we conduct comprehensive empirical evaluations of the query selection heuristics that have been proposed in model-based diagnosis literature. Using real-world benchmark diagnosis problems, 
we investigate the heuristics under varying conditions regarding the \emph{(a)}~diagnosis probability distributions, \emph{(b)}~quality (meaningfulness) of the probabilities, \emph{(c)}~available diagnostic evidence (size of the computed diagnosis sample) for query computation, and \emph{(d)}~diagnostic structure (i.e., number of system components; number and cardinality of diagnoses; reasoning complexity).\vspace{4pt}

\noindent\emph{Results:}
Our main findings based on experiments using real-world problem cases are:
\begin{itemize}[noitemsep,topsep=0pt]
	\item Using an appropriate heuristic is essential. For inadequate choices of heuristics, we observe overheads in terms of oracle effort of avg.\ / max.\ $100\,\%$ / $250\,\%$.
	\item The one and only (generally) best heuristic does not exist, or has not yet been found.
	\item The main factors affecting the number of queries required for fault localization 
	are the number and size of the given diagnoses as well as the degree of bias in and quality of the given fault information.
	\item In different diagnosis scenarios, higher or lower numbers of computed diagnoses can be appropriate in combination with a used heuristic. We make recommendations in this regard.
	\item Different heuristics prevail in different scenarios. Specifically, for 
	\begin{enumerate}[label=\emph{(\alph*)},noitemsep,topsep=0pt]
		\item misleading fault information (e.g., derived from historical data which does not apply to the current problem), the random query selection strategy is overall preferable to all others,
		\item vague fault information (e.g., where the probabilities are 	partially reasonable, and partially not), the MPS 
		heuristic \cite{rodler_active_2017}, which selects a query that maximizes the probability of an elimination of a maximal number of diagnoses, turns out to be best approach, and for
		\item reasonable fault information, the 
		BME 
		heuristic \cite{rodler_active_2017}, which proposes a query that maximizes the number of diagnoses that can be eliminated with a probability of more than $50\,\%$, is most favorable (closely followed by the well-known information entropy heuristic \cite{de_kleer_diagnosing_1987} and the EMCb heuristic \cite{rodler_active_2017}, which suggests a query that maximizes the expected diagnosis elimination rate).
	\end{enumerate}
\end{itemize}

\subsubsection*{Publication: Meta Information}

\begin{tabular}{@{}lp{9.5cm}}
	\emph{Publication included in Thesis:} & \cite{rodler_impact_2018} \\
	\emph{Authors:} & Patrick Rodler, Wolfgang Schmid
	\\
	\emph{Title:} &  On the impact and proper use of heuristics in \newline test-driven ontology debugging
	\\
	\emph{Year:} & 2018
	\\
	\emph{Publication Venue (Type):} & Int'l Joint Conference on Rules and Reasoning \quad (Conference)
	\\
	\emph{Venue Metrics:} & A
	\\
	\emph{Own Contributions (est.):} & 
	Concept/Idea: $100\,\%$ \newline
	Theory/Algorithms: $100\,\%$ \newline
	Development/Implementation: $40\,\%$ \newline
	Evaluation: $80\,\%$ \newline
	Manuscript: $100\,\%$
	\\
	\emph{Other Publications on the Topic:} & \cite{rodler_comparing_2018} (Int'l Workshop on Principles of Diagnosis) \newline
	\cite{rodler_evaluating_2018} (Technical report)
\end{tabular}

\subsection{Efficient Query Computation and Selection by Systematic Search}
\label{sec:SD_by_systematic_search}

\begin{description}[noitemsep,topsep=0pt]
	\item[Publication: System Aspects and Goals Addressed] See Fig.~\ref{fig:seq_diag_by_systematic_search}.
\end{description}
\vspace{-7pt}

\subsubsection*{Publication: Contents in a Nutshell}


\noindent\emph{Motivation:}
As the determination of optimal queries is NP-hard \cite{pattipati_application_1990}, sequential diagnosis methods often have to rely on myopic strategies to gauge the gain of different query candidates (cf.\ Sec.~\ref{sec:sequential_diagnosis}). One approach is to evaluate the quality of
queries in terms of the utility of the expected situation after knowing their answer. Such one-step-lookahead analysis
has proven to constitute a particularly favorable trade-off between computational efficiency and diagnostic effectivity,
and is today state-of-the-art in sequential diagnosis. However, this solves only a part of the problem, as various other
sources of complexity remain. Examples include the strong reliance of sequential diagnosis tools on costly reasoning
services, highly expressive knowledge representation languages used for system modeling, large numbers of query
candidates, and that queries might neither be explicitly given nor efficiently computable (cf., e.g., \cite{rodler_interactive_2015}). 

Existing sequential diagnosis methods often assume a particular \emph{(i)}~type of diagnosed system, \emph{(ii)}~formalism to describe the system, \emph{(iii)}~inference engine, \emph{(iv)}~type of query to be of interest, \emph{(v)}~query quality criterion to be adopted,
or \emph{(vi)}~diagnosis computation algorithm to be employed. Further, they often cannot deal with large or implicit query
spaces or with expressive logics, or require inputs (e.g., certain inferences) that cannot always be provided.\vspace{4pt}

\noindent\emph{Contributions:}
As a remedy, we propose a novel one-step lookahead query computation technique for sequential diagnosis that
overcomes the mentioned issues of existing methods. More speciﬁcally, our approach \emph{(1)}~is well-founded and based
on a solid theory, \emph{(2)}~involves a systematic search for optimal queries, \emph{(3)}~can operate on implicit and huge query spaces, \emph{(4)}~allows for a two-stage optimization of queries (wrt.\ both their number and their cost), \emph{(5)}~is designed to reduce the required logical inferences to a minimum, and \emph{(6)}~is generally applicable in that it can deal with any type of diagnosis problem as per Reiter's theory \cite{reiter_theory_1987}, is applicable with any monotonic knowledge representation language, can interact with a multitude of diagnosis engines and logical reasoners, and allows for a quality optimization of queries based on any of the common heuristics in literature (cf.\ Sections~\ref{sec:active_learning} and \ref{sec:impact_of_heuristics_in_ontology_debugging}).\vspace{4pt}

\noindent\emph{Results:}
Using a benchmark of real-world diagnosis problems, we extensively studied the performance of the novel technique (N) and compared it with the state-of-the-art method (S). The gained insights are: 
\begin{itemize}[noitemsep,topsep=0pt]
	\item N is complete, i.e., it can explore all queries (as opposed to S),
	\item N is non-redundant, i.e., it considers each query at most once (as opposed to S),
	\item N can compute queries without any expensive reasoner calls (as opposed to S), 
	\item N gives theoretical guarantees about the achieved query quality (as opposed to S),
	\item N is by orders of magnitude faster than S,
	\item N \emph{always} returns as good or better queries than S, and
	\item N scales to input sizes of up to hundreds of diagnoses while S can handle only single-digit numbers.
\end{itemize}
As a conclusion, given this new approach, query computation and selection is no longer a bottleneck in sequential diagnosis, even in particularly challenging scenarios where queries are neither explicitly given nor efficiently extractable from the system model.

\subsubsection*{Publication: Meta Information}

\begin{tabular}{@{}lp{9.5cm}}
	\emph{Publication included in Thesis:} & \cite{rodler_sequential_2022} \\
	\emph{Authors:} & Patrick Rodler
	\\
	\emph{Title:} &  Sequential diagnosis by systematic search
	\\
	\emph{Year:} & 2022
	\\
	\emph{Publication Venue (Type):} & Artificial Intelligence {\small(under revision)} \quad (Journal)
	\\
	\emph{Venue Metrics:} & A \qquad\quad (Scimago: Q1, Journal Rank Indicator 1.673)
	\\
	\emph{Own Contributions:} & Single-author publication
	\\
	\emph{Other Publications on the Topic:} & \cite{rodler_inexpensive_2017} (Int'l Workshop on Principles of Diagnosis) \newline
	\cite{rodler_towards_2016} (Technical report) \newline
	\cite{rodler_generally_2017} (Technical report) \newline
	\cite{rodler_appendix_2022} (Technical report)
\end{tabular}

\section{User Studies and Debugging Tool}
\label{chap:user_studies+debugging_tool}

\subsection{User Study: Are Query-Based Debuggers Really Helping Users?}
\label{sec:are_query-based_debuggers_really_helpful}

\begin{description}[noitemsep,topsep=0pt]
	\item[Publication: System Aspects and Goals Addressed] See Fig.~\ref{fig:user_study}.
\end{description}
\vspace{-7pt}

\subsubsection*{Publication: Contents in a Nutshell}


\noindent\emph{Motivation:}
Interactive debugging based on queries that are automatically generated by the diagnosis system is an attractive methodology as tools implementing this technique can interactively guide their users to the true cause of the observed problem. This query-based approach has been successfully adopted, e.g., for ontology debugging problems, an equally challenging as topical application domain in the era of the Semantic Web \cite{berners-lee_semantic_2001}. It has proven to be a 
feasible and efficient approach in various evaluations based on computational experiments (e.g., \cite{shchekotykhin_interactive_2012,rodler_impact_2018}).
While such simulation-based evaluations can assess and compare the performance of debugging algorithms in terms of 
various interesting aspects such as the time and space complexity, the number of reasoner calls, 
the theoretical number of required user interactions, or the precision of the fault localization process, they also have 
certain limitations and often cannot fully inform us about a method's true usefulness. E.g., they cannot be used to determine if 
certain 
assumptions made by an evaluated debugging method hold for actual users, or to gauge the user acceptance. This can be accomplished by an evaluation approach based on user studies where the performance and behavior of experts while using a debugger is observed and analyzed. 
No user studies with a focus on the investigation of query-based ontology debugging have 
been published so far.
\vspace{4pt} 

\noindent\emph{Contributions:}
We conducted user studies in the form of testing and debugging exercises that were specifically designed to evaluate if query-based debugging is truly favorable over a previous debugging approach based on test cases, where users specify and add measurements (logical sentences) to the DPI manually in order to prune the diagnosis space. The studies were performed in the context of ontology debugging, where we used a version of the OntoDebug tool \cite{schekotihin_ontodebug_2018} (cf.\ Sec.~\ref{sec:ontodebug}) for the experiments, which i.a.\ involved fault localization tasks given ontologies with injected faults. The study participants were computer science students who had a certain level of education in the development and debugging of ontologies and who received some initial training with the tool. The examined research questions were related to \emph{(i)}~the efficiency and effectiveness of query-based debugging (i.e., do experts need less time, do they find more faults?), \emph{(ii)}~the cognitive ability of users to find out which of the identified diagnoses is the correct one, and \emph{(iii)}~the difficulty of answering system-generated queries for experts.\vspace{4pt}

\noindent\emph{Results:}
Our studies revealed the following:
\begin{itemize}[noitemsep,topsep=0pt]
	\item A query-based approach can make the debugging process more efficient (wrt.\ user time and effort), without leading to a loss in effectiveness (wrt.\ 
	the rate of 
	identified faulty ontology axioms). 
	\item The measured overhead using the manual approach amounts to an average of $37\,\%$ wrt.\ time and $117\,\%$ wrt.\ effort (captured by the number of required mouse clicks).
	\item The capability of a debugger to rank the actual diagnosis higher in a list of candidates visible to the users does not translate into a more effective debugging process. That is, users generally do not have a ``perfect bug understanding'', a result in line with similar studies conducted in the software debugging domain \cite{parnin_are_2011}.
	\item The experiments revealed that ``oracle errors'', where users provide faulty inputs, are relatively frequent. Indeed, at least one fault occurred to a quarter of the study participants.
\end{itemize}
As unveiled by a survey of the literature, oracle errors are a largely open issue, as algorithmic testing or debugging methods usually do not assume, handle or examine oracle errors. As a first step towards approaching this topic, 
we thus propose a prediction model for oracle errors based on insights from our studies. An assessment of the model evinces that queries it estimates to be hard in fact 
\emph{(1)}~lead to a higher failure rate, 
\emph{(2)}~are perceived to be harder, and 
\emph{(3)}~result in a lower confidence of users in their answers, and vice versa.

%

\subsubsection*{Publication: Meta Information}

\begin{tabular}{@{}lp{9.5cm}}
	\emph{Publication included in Thesis:} & \cite{rodler_are_2019-1} \\
	\emph{Authors:} & Patrick Rodler, Dietmar Jannach, Konstantin Schekotihin, \newline Philipp Fleiss
	\\
	\emph{Title:} &  Are query-based ontology debuggers really helping \newline knowledge engineers?
	\\
	\emph{Year:} & 2019
	\\
	\emph{Publication Venue (Type):} & Knowledge-Based Systems \quad (Journal)
	\\
	\emph{Venue Metrics:} & A  \qquad\quad (Scimago: Q1, Journal Rank Indicator 2.192)
	\\
	\emph{Own Contributions (est.):} & 
	Concept/Idea: $40\,\%$ \newline
	Theory/Algorithms: $50\,\%$ \newline
	Development/Implementation: $50\,\%$ \newline
	Evaluation: $70\,\%$ \newline
	Manuscript: $60\,\%$
	\\
	\emph{Other Publications on the Topic:} & None
\end{tabular}

\subsection{On User Types, Assumptions, Optimization Criteria, and How to Make Query-Based Debugging Simpler and More Efficient}
\label{sec:one_step_at_a_time}

\begin{description}[noitemsep,topsep=0pt]
	\item[Publication: System Aspects and Goals Addressed] See Fig.~\ref{fig:one_step_at_a_time}.
\end{description}
\vspace{-7pt}

\subsubsection*{Publication: Contents in a Nutshell}


\noindent\emph{Motivation:}
To suggest as informative queries as possible, existing sequential diagnosis methods draw on various algorithmic optimizations as well as heuristics. However, these computations are often based on certain assumptions about the interacting expert and its behavior, which often have not been fully validated.\vspace{4pt}

\noindent\emph{Contributions:}
In this work, we study the reasonability of assumptions about users (experts), optimization techniques and criteria made by state-of-the-art interactive debugging systems in the application context of ontology debugging. We demonstrate that the made assumptions might not always be adequate and discuss consequences of their violations. In particular, we characterize a range of expert types with different query answering behavior and show that existing approaches are far from achieving optimal efficiency for all of them. In addition, we find that the cost metric adopted by state-of-the-art techniques might not always be realistic and that a change of metric can have a decisive impact on the best choice of query answering strategy. As a remedy, based on the insights of our investigations, we suggest a new---and simpler---type of expert question that leads to a stable fault localization performance for all analyzed expert types and effort metrics, and has a range of further advantages over existing techniques, e.g., smaller query search spaces. Moreover, we present a polynomial-time algorithm for computing optimal queries of the suggested type which is fully compatible with existing concepts (e.g., query selection heuristics \cite{rodler_active_2017}) and infrastructure (e.g., debugging user interfaces \cite{schekotihin_ontodebug_2018}) in the field.\vspace{4pt} 

\noindent\emph{Results:}
Comprehensive experiments encompassing more than $23\,000$ fault localization sessions on faulty real-world problem cases attest that the new querying method is substantially and statistically significantly superior to existing techniques both in terms of the number of necessary expert interactions and in terms of the query computation time. We find that relying on the new querying method is better than drawing on existing techniques in $98\,\%$ of the investigated diagnosis scenarios, and can save an interacting expert more than $80\,\%$ of their work. In addition, the new approach can reduce the expert's waiting time for the next query by more than three orders of magnitude, or, in other words, avoid an overhead of more than $120\,000\,\%$. 
E.g., for one problem case we recorded computation times for the optimal normal query by the state-of-the-art approach of avg.\ / max.\ $150$ sec / $430$ sec, whereas an optimal query of the new type was generated in $0.49$ sec / $1.06$ sec by the novel algorithm, which never required more than $1.1$ sec to find a best next query throughout all conducted experiments. Beside these findings, we demonstrate that the efficiency of existing query-based tools can be significantly boosted by suggesting an appropriate query answering strategy to an expert; we also make recommendations in this regard. Further, we suggest optimal configurations of a debugger for situations where the new type of query is used.


\subsubsection*{Publication: Meta Information}

\begin{tabular}{@{}lp{9.5cm}}
	\emph{Publication included in Thesis:} & \cite{rodler_one_2022} \\
	\emph{Authors:} & Patrick Rodler 
	\\
	\emph{Title:} &  One step at a time: An efficient approach to query-based \newline ontology debugging
	\\
	\emph{Year:} & 2022
	\\
	\emph{Publication Venue (Type):} & Knowledge-Based Systems \quad (Journal)
	\\
	\emph{Venue Metrics:} & A  \qquad\quad (Scimago: Q1, Journal Rank Indicator 2.192)
	\\
	\emph{Own Contributions:} & Single-author publication
	\\
	\emph{Other Publications on the Topic:} & \cite{rodler_usefulness_2019} (Int'l Conference on Industrial Engineering and \newline Other Applications of Applied Intelligent Systems) \newline
	\cite{rodler_how_2019} (Joint Ontology Workshops) \newline
	\cite{rodler_new_2019} (Technical report) \newline
	\cite{rodler_expert_2020} (Technical report)
\end{tabular}

\subsection{OntoDebug: A Full-Fledged Ontology Debugging Tool}
\label{sec:ontodebug}

\begin{description}[noitemsep,topsep=0pt]
	\item[Publication: System Aspects and Goals Addressed] See Fig.~\ref{fig:ontodebug}.
\end{description}
\vspace{-7pt}

\subsubsection*{Publication: Contents in a Nutshell}


\noindent\emph{Motivation:}
Applications of semantic systems require their users to design ontologies that correctly formalize knowledge about a domain. One example of a vital field extensively adopting ontologies for highly critical applications is biomedicine, cf., e.g., the OBO project\footnote{http://obo.sourceforge.net} or the NCI-Thesaurus\footnote{http://ncit.nci.nih.gov}. In many cases, factors such as an insufficient understanding of a knowledge representation language, problems concerning modeling techniques and granularity, 
collaborative development processes, large and complex domains to be described, or the inability to foresee all implications of formulated axioms result in faulty ontologies. Possible deficiencies of ontologies are, e.g., logical contradictions or wrong entailments, which can have severe consequences, e.g., when a wrong therapy is suggested for a patient in health-related applications. What additionally exacerbates the problem is that the root cause of such errors can be extremely hard to identify for humans.

Debugging tools help to localize faults in ontologies by finding explanations of discrepancies between the actual ontology and the intended one. Existing ontology development environments, such as Protégé \cite{musen_protege_2015}, Swoop \cite{kalyanpur_swoop_2006}, or ORE \cite{lehmann_ore_2010}, implement various expedient 
methodologies to assist users in 
debugging tasks. 
However,
they are mainly based on the principle of presenting collections of conflicts (also referred to as justifications in this context) to the interacting user, who may then need to execute manual analyses in order
to figure out the actually faulty axioms and how to appropriately fix them. 

Since, in practice, it is often the case that the initially given information is insufficient to unambiguously localize the faulty axioms, 
there may exist a possibly large number of minimal conflicts and minimal diagnoses, which can render a manual analysis tedious at best, and error-prone or sheerly infeasible at worst. In such scenarios, tool support in the form of interactive debuggers, which semi-automatically guide users to the actual fault by asking a series of questions (whether certain axioms must or must not be entailments of the intended ontology), can greatly alleviate the load on the user, both cognitively and in terms of time and effort (cf., Sec.~\ref{sec:are_query-based_debuggers_really_helpful} and \cite{rodler_are_2019-1}). However, to date no such interactive debugging tools have been proposed for ontology debugging.\vspace{4pt}


\noindent\emph{Contributions:}
We present OntoDebug, an interactive ontology debugger integrated in Protégé, the most popular open-source ontology editor in the world that is used for the maintenance, development, and quality assurance of OBO ontologies and the NCI-Thesaurus (see above), and for numerous other 
critical applications. OntoDebug is free, open-source, and publicly available in the standard plug-ins repository of Protégé. Its development and implementation was a yearslong endeavor, underpinned and considerably shaped by well-founded research results of our group over more than a decade.\vspace{4pt}

\noindent\emph{Results:} OntoDebug builds upon the ideas of the existing tools discussed above and improves on them i.a.\ as follows:
\begin{itemize}[noitemsep,topsep=0pt]
	\item The tool aids the user wrt.\ all three principal model-based diagnosis tasks (cf. Sec.~\ref{sec:diagnostic_tasks}), i.e., in terms of fault detection, localization, and repair.  
	\item The plug-in supports an interactive query-based ontology debugging. By using the techniques suggested in \cite{shchekotykhin_interactive_2012,rodler_interactive_2015} and further improved, e.g., in \cite{rodler_active_2017,rodler_impact_2018,rodler_inexpensive_2017}, the tool is user-centric and based on a query-answer dialogue where the user can rely on their domain knowledge to answer questions about axioms of the intended ontology in order to successively restrict the set of diagnoses. In this vein, the system guides the user throughout the fault localization process and, as opposed to existing approaches, the user is not required to manually specify, analyze or compare various fault explanations in terms of diagnoses and conflicts. 
	\item The plug-in enables a test-driven ontology development, which is closely related and based on the same principles as the familiar test-driven software development \cite{beck_test-driven_2003}. It supports the specification of test cases, which capture requirements 
	on the intended ontology. In particular, the user can define axioms which must be entailments or non-entailments of the intended ontology. These are then automatically verified 
	and a debugging session is started if at least one of them is violated.
	\item The tool offers a repair interface enabling non-intrusive modifications of the ontology. In the process towards defining a correct repair, a user can introduce and test various modifications of faulty axioms on a virtual copy of the ontology. All modifications are only applied to the productive ontology after the user is satisfied with the obtained result.
	\item OntoDebug provides the opportunity of a versatile parametrization, configuration, individualization, and combination of various algorithms. 
	Such a proper selection and fine-tuning of used techniques, e.g., based on research results, can help to streamline the performance of OntoDebug based on a given diagnosis problem setting.
\end{itemize}
As of now, the tool boasts a tally of more than 50K downloads.

\subsubsection*{Publication: Meta Information}

\begin{tabular}{@{}lp{9.5cm}}
	\emph{Publication included in Thesis:} & \cite{schekotihin_ontodebug_2018} \\
	\emph{Authors:} & Konstantin Schekotihin, Patrick Rodler, Wolfgang Schmid 
	\\
	\emph{Title:} &  OntoDebug: Interactive ontology debugging plug-in for Protégé
	\\
	\emph{Year:} & 2018
	\\
	\emph{Publication Venue (Type):} & Int'l Symposium on Foundations of Information and \newline Knowledge Systems \quad (Conference)
	\\
	\emph{Venue Metrics:} & B
	\\
	\emph{Own Contributions (est.):} & 
	Concept/Idea: $50\,\%$ \newline
	Theory/Algorithms: $60\,\%$ \newline
	Development/Implementation: $25\,\%$ \newline
	Evaluation: n.a. \newline
	Manuscript: $50\,\%$
	\\
	\emph{Other Publications on the Topic:} & \cite{schekotihin_protege_2018} (Int'l Conference on Biological Ontology) \newline
	\cite{schekotihin_test-driven_2018} (Int'l Conference on Biological Ontology)
\end{tabular}

\section{Diagnostic Reasoning}
\label{chap:diagnostic_reasoning}

\begin{figure}
	\centering
	\includegraphics[trim={5cm 0 0 0},clip,width=0.9\linewidth]{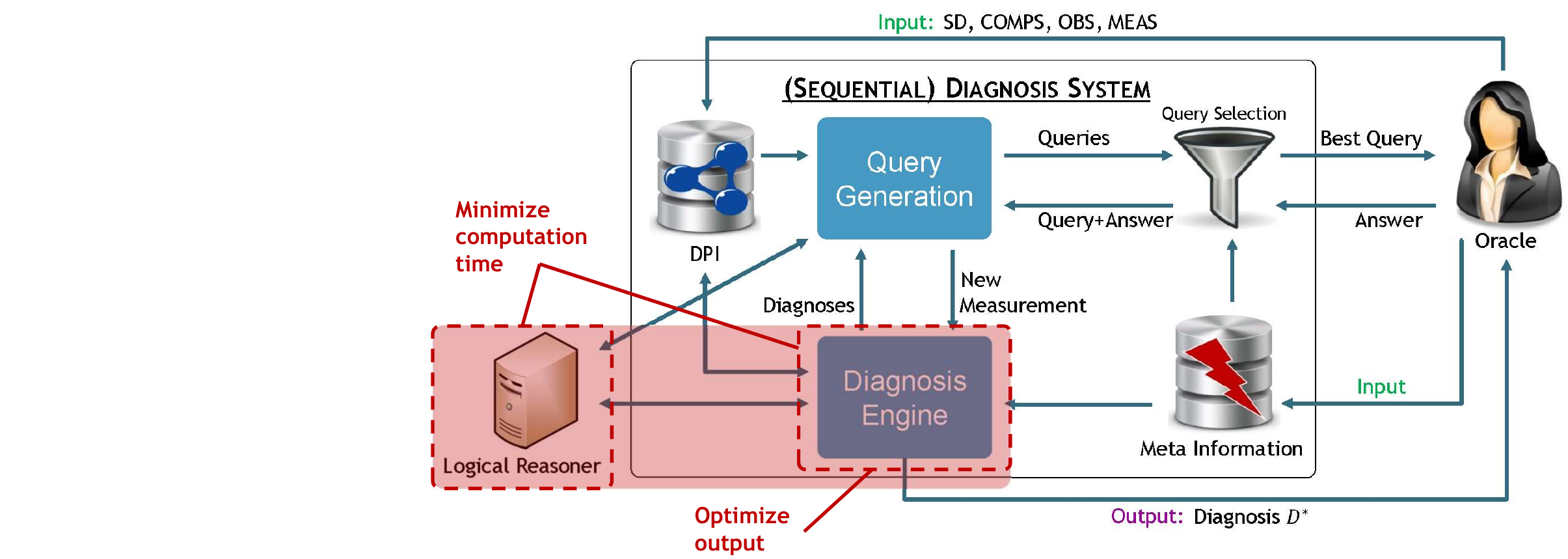}
	\caption[Optimization by randomization for hard diagnostic problem cases]{Optimization by randomization for hard diagnostic problem cases: Addressed system aspects and goals.}
	\label{fig:problem-relax-solving}
\end{figure}

\subsection{Optimization by Randomization for Hard Diagnostic Problem Cases}
\label{sec:randomized_problem_relaxation_solving}

\begin{description}[noitemsep,topsep=0pt]
	\item[Publication: System Aspects and Goals Addressed] See Fig.~\ref{fig:problem-relax-solving}.
\end{description}
\vspace{-7pt}

\subsubsection*{Publication: Contents in a Nutshell}

\noindent\emph{Motivation:}
As discussed in Sec.~\ref{sec:MBD}, the logical reasoning is in many cases the most computationally costly operation in the course of a model-based diagnosis task. Thus, the (expected) time required per theorem prover call also has an influence on the selection of the algorithms employed for debugging. In particular, given theorem proving is cheap (taking time, say, in the range of milliseconds per call of an inference engine), a wider range of algorithms will be applicable to a problem instance than in case reasoning operations are expensive. One particularly challenging domain in this regard is given by, e.g., over-constrained scheduling problems involving too many jobs (products) that cannot all be accomplished (produced) until a given deadline, e.g., due to seasonal order fluctuations, unforeseen machine breakdowns, or incoming high-priority orders. In this diagnosis use case, times per reasoner call, even if executed by a highly performant cutting-edge solver, can reach up to several minutes. 
When the objective in such a scenario is the computation of a preferred diagnosis, e.g., a minimal-cost or minimum-cardinality set of jobs to be dropped from the schedule in a way the remaining jobs can be serviced in time, we often measure times in the range of multiple hours per diagnosis computation. Such long waiting times however might not be practical in today's highly dynamical production regimes. 
For such demanding diagnostic conditions, a more efficient diagnosis computation and optimization approach 
is required to achieve a maximal customer satisfaction (e.g., by postponing the fewest 
products 
as a result of computing a preferred diagnosis) while maintaining a higher decision flexibility (through a reasonable diagnosis computation time), which can be an essential factor towards giving the company a competitive edge.\vspace{4pt}

\noindent\emph{Contributions:}
Catering for the described diagnosis use cases where the inefficiency of logical inference operations becomes too severe a bottleneck for standard diagnosis techniques to be put into action, we suggest a novel algorithm for the computation of preferred (e.g., minimum-cardinality or most probable) diagnoses. The method is based on the principle of optimization by randomization, where random minimal diagnoses are successively generated until a (sufficiently) optimal diagnosis is found. The underlying rationale is to trade one hard optimization problem 
for a range of easier decision problems 
by explicitly solving (multiple instances of) the subset-minimality problem implicit in the optimization problem.
Applied to the mentioned production industry domain, given an over-constrained scheduling task stated as a constraint optimization problem, this means to forgo the (\emph{very inefficient}) direct 
determination of a preferred diagnosis by means of a state-of-the-art constraint solver. Instead, our approach involves 
calling an MSMP algorithm such as QuickXplain \cite{junker_quickxplain_2004,rodler_formal_2022} (cf.\ Sections~\ref{sec:MBD}, \ref{sec:related_fields} and \ref{sec:qx_proof}) multiple times with a randomly modified input (cf.\ \cite{rodler_random_2022}) to generate a random sample of minimal diagnoses, where each such execution of the MSMP algorithm requires a linear number of (\emph{comparably much more efficient}) consistency checks. We note that this suggested separation between the optimization problem to be solved and its inherent relaxed subset-minimality problem is usually not supported by modern constraint solvers, which is a confirmation that alternative approaches like the one introduced can be useful.

The proposed algorithm is efficiently parallelizable since no information exchange is necessary between successive diagnosis computations, and it can use and benefit from the most suitable and performant algorithms for the involved sub-problems (random number generation, theorem proving, and 
MSMP problem) as all modules of the method are viewed as black-boxes.\vspace{4pt}


\noindent\emph{Results:}
In extensive tests on popular benchmark instances, we compared the performance of the novel optimization-by-randomization technique to a direct approach to solving the optimization problem drawing on a 
world's leading solver for scheduling problems. The results showed that, first, the new method manifested significant time improvements. In fact, it was \emph{in all cases} able to find better diagnoses in less than half the time required by the direct approach. Second, the proposed strategy could achieve substantial 
quality enhancements of the computed diagnoses in that 
over $25\,\%$ / $60\,\%$ fewer product orders needed to be canceled or postponed on average / maximally.

\subsubsection*{Publication: Meta Information}

\begin{tabular}{@{}lp{9.5cm}}
	\emph{Publication included in Thesis:} & \cite{rodler_randomized_2021} \\
	\emph{Authors:} & Patrick Rodler, Erich Teppan, Dietmar Jannach 
	\\
	\emph{Title:} &  Randomized problem-relaxation solving \newline for over-constrained schedules
	\\
	\emph{Year:} & 2021
	\\
	\emph{Publication Venue (Type):} & Int'l Conference on Principles of Knowledge Representation \newline and Reasoning \quad (Conference)
	\\
	\emph{Venue Metrics:} & A
	\\
	\emph{Own Contributions (est.):} & 
	Concept/Idea: $50\,\%$ \newline
	Theory/Algorithms: $60\,\%$ \newline
	Development/Implementation: $0\,\%$ \newline
	Evaluation: $50\,\%$ \newline
	Manuscript: $70\,\%$
	\\
	\emph{Other Publications on the Topic:} & \cite{rodler_scheduling_2020} (Int'l Workshop on Principles of Diagnosis)
\end{tabular}

\subsection[A Formal Proof and Simple Explanation of the Seminal QuickXplain Algorithm]{A Formal Proof and Simple Explanation of the QuickXplain Algorithm}
\label{sec:qx_proof}

\begin{description}[noitemsep,topsep=0pt]
	\item[Publication: System Aspects and Goals Addressed] See Fig.~\ref{fig:QX}.
\end{description}
\vspace{-7pt}

\subsubsection*{Publication: Contents in a Nutshell}

\noindent\emph{Motivation:}
In his seminal\footnote{On Google Scholar, the paper boasts more than 550 citations as of December 2022.} paper \cite{junker_quickxplain_2004} of 2004, Ulrich Junker proposed the QuickXplain algorithm, which provides a divide-and-conquer computation strategy to find within a given set an irreducible subset with a particular (monotone) property. Beside its original application in the domain of constraint satisfaction problems, the algorithm has since then found widespread adoption in areas as different as model-based diagnosis, recommender systems, verification, or the Semantic Web. This popularity is due to the frequent occurrence of the problem of finding irreducible subsets (cf.\ Sec.~\ref{sec:related_fields}) on the one hand, and to QuickXplain’s general applicability and favorable computational complexity on the other hand. However, although (we regularly experience and have conducted a study to confirm that) people are having a hard time understanding QuickXplain and seeing why it works correctly, a proof of correctness of the algorithm has never been published.\vspace{4pt}

\noindent\emph{Contributions:}
This is what we account for in this work, by explaining QuickXplain in a novel tried and tested way, based on a simple, accessible ``flat'' notation instead of a tree (cf.\ Example~\ref{ex:conflict_computation}), and by presenting a first formal correctness proof of it. The latter is expounded as a ``proof to explain'', where a particular focus is put on its intelligibility by constructing and explaining it in a way to illuminate the workings and to foster the understandability and intuition of the algorithm.\vspace{4pt}

\noindent\emph{Results:}
Apart from showing the correctness of the algorithm and excluding the later detection of errors (\emph{proof and trust effect}), the added value of the availability of a formal proof is, e.g., \emph{(i)} that the workings of the algorithm often become completely clear only after studying, verifying and comprehending the proof (\emph{didactic effect}), \emph{(ii)}~that the shown proof methodology can be used as a guidance for proving other recursive algorithms (\emph{transfer effect}), and \emph{(iii)}~the possibility of providing ``gapless'' correctness proofs of systems that rely on (results computed by) QuickXplain, such as numerous model-based debuggers (\emph{completeness effect}).

\subsubsection*{Publication: Meta Information}

\begin{tabular}{@{}lp{9.5cm}}
	\emph{Publication included in Thesis:} & \cite{rodler_formal_2022} 
	\\
	\emph{Authors:} & Patrick Rodler 
	\\
	\emph{Title:} &  A formal proof and simple explanation of \newline the QuickXplain algorithm
	\\
	\emph{Year:} & 2022
	\\
	\emph{Publication Venue (Type):} & Artificial Intelligence Review \quad (Journal)
	\\
	\emph{Venue Metrics:} & A  \qquad\quad (Scimago: Q1, Journal Rank Indicator 2.180)
	\\
	\emph{Own Contributions:} & Single-author publication
	\\
	\emph{Other Publications on the Topic:} & 
		\cite{rodler_interactive_2015} (PhD thesis) \newline
	\cite{rodler_understanding_2020} (Technical report) 

\end{tabular}

\section{Diagnostic Modeling}
\label{chap:diagnostic_modeling}

\subsection{On Modeling Techniques for Spreadsheet Debugging: \\ A Theoretical and Empirical Analysis}
\label{sec:modeling_for_spreadsheet_debugging}

\begin{description}[noitemsep,topsep=0pt]
	\item[Publication: System Aspects and Goals Addressed] See Fig.~\ref{fig:spreadsheet_debugging}.
\end{description}
\vspace{-7pt}

\subsubsection*{Publication: Contents in a Nutshell}

\noindent\emph{Motivation:}
Based on a formal description of the system, model-based diagnosis allows to precisely and deterministically reason about potential faults responsible for an observed system misbehavior. Under certain circumstances such a formal model can even be extracted from the buggy system in a fully automatic way, such as often in case of software. Since logical reasoning is a core building block of the diagnosis process, the performance of model-based debuggers is largely affected by the reasoning efficiency. The latter in turn depends on the complexity and expressivity of the system description. As highly detailed system descriptions capturing the exact semantics of the system are often beyond the reach of state-of-the-art reasoning techniques, researchers have proposed to reduce the level of detail by introducing more abstract ways to describe the system. When relying on such non-exact system models, it is of interest in how far the reduced detail in terms of modeling translates to an increased efficiency of computations based on the model, and which effects it has on the informativeness of diagnostic inference.\vspace{4pt}

\noindent\emph{Contributions:}
In this work, we thoroughly analyze different system modeling techniques with a focus on fault localization in spreadsheets, one of the most popular end-user programming paradigms used by millions in a wide range of areas. Specifically, we present three constraint model types to characterize a spreadsheet at various levels of abstraction, show how they can be extracted from a buggy spreadsheet automatically, and provide theoretical and empirical investigations of the implications of the abstraction level on the computed diagnostic solutions as well as on the computation performance. The model types under study are \emph{(1)}~an exact value-based model that captures the precise semantics of the spreadsheet, \emph{(2)}~a functional dependency model that abstracts from the concrete numerical values and allows to reason only in terms of whether spreadsheet cells are correct or faulty, and \emph{(3)}~a newly proposed qualitative deviation model 
whose level of abstractness lies between the two other model types in that it allows to express whether (numerical) cell values are loo low, exact, or too high.\vspace{4pt}

\noindent\emph{Results:}
The main theoretical findings are that
\begin{itemize}[noitemsep,topsep=0pt]
	\item all model types are	diagnostically complete in that no diagnosis can be missed, i.e., the localization of the actually faulty cells is facilitated by each model type,
	\item the non-exact abstract models may be unsound in that they may generate spurious diagnoses that are in fact no valid fault explanations, and 
	\item in general, a higher degree of abstraction implies an equal or lower diagnostic accuracy (equally many or more generated diagnoses) and equally many or more spurious diagnoses.
\end{itemize} 
Empirically, based on comprehensive experiments using real-world and synthetic benchmarks of faulty spreadsheets, the key lessons learned are that 
\begin{itemize}[noitemsep,topsep=0pt]
	\item the exact model produces significantly fewer diagnoses than the abstract models on average, albeit the number of returned diagnoses is always within one order of magnitude for all model types,
	\item the exact model is often not (efficiently) applicable, thus abstract models are well motivated,
	\item diagnosis computation performance is very similar for both abstract models and by at least one and up to more than four orders of magnitude faster on average than for the value-based model,
	\item the proposed qualitative deviation model is as good regarding diagnostic accuracy as the exact model in a third of all examined cases, and up to more than half of the cases per diagnostic 
	scenario,
	\item in general, when the exact model exhibits poor performance, then abstract models can be a powerful and efficient surrogate.
\end{itemize}

\subsubsection*{Publication: Meta Information}

\begin{tabular}{@{}lp{9.5cm}}
	\emph{Publication included in Thesis:} & \cite{hofer_modeling_2022} 
	\\
	\emph{Authors:} & Birgit Hofer, Dietmar Jannach, Iulia Nica, \newline Patrick Rodler, Franz Wotawa 
	\\
	\emph{Title:} &  On modeling techniques for spreadsheet debugging: \newline A theoretical and empirical analysis
	\\
	\emph{Year:} & 2022
	\\
	\emph{Publication Venue (Type):} & Artificial Intelligence {\small(to be submitted shortly)} \quad (Journal)
	\\
	\emph{Venue Metrics:} & A  \qquad\quad (Scimago: Q1, Journal Rank Indicator 1.673)
	\\
	\emph{Own Contributions (est.):} & 
	Concept/Idea: $40\,\%$ \newline
	Theory/Algorithms: $65\,\%$ \newline
	Development/Implementation: $0\,\%$ \newline
	Evaluation: $70\,\%$ \newline
	Manuscript: $85\,\%$
	\\
	\emph{Other Publications on the Topic:} & None
\end{tabular}

\subsection{Reducing Model-Based Diagnosis to Knowledge Base Debugging}
\label{sec:reducing_MBD_to_KBD}

\begin{description}[noitemsep,topsep=0pt]
	\item[Publication: System Aspects and Goals Addressed] See Fig.~\ref{fig:reducing_MBD_to_KBD}.
\end{description}
\vspace{-7pt}

\subsubsection*{Publication: Contents in a Nutshell}


\noindent\emph{Motivation:}
In the literature, we have witnessed a range of ways to formalize the model-based diagnosis (MBD) problem. This variety originates, e.g., from different application areas of interest to be addressed, different notational conventions or preferences, or the desire to represent aspects not captured by Reiter's original formulation of the problem \cite{reiter_theory_1987}. In particular, we have in our research intensively dealt with the knowledge base debugging (KBD) problem \cite{rodler_interactive_2015}, which draws on concepts from model-based diagnosis to find faults in a monotonic knowledge base. However, on a very precise theoretical level, the exact relationship between Reiter's account and ours was for long not fully clear and subject of recurring discussions between researchers.\vspace{4pt}      

\noindent\emph{Contributions:}
This paper brings light into the darkness regarding this matter. To this end, we theoretically analyze and compare the two views on the debugging problem.\vspace{4pt}

\noindent\emph{Results:}
We find that KBD is a generalization of MBD in that any MBD problem can be reduced to a KBD problem and solutions of the former can be directly extracted from solutions of the latter. Moreover, our studies reveal that the sequential MBD problem (cf.\ Sec.~\ref{sec:sequential_diagnosis}) is a special case of the sequential KBD problem (cf.\ \cite{rodler_interactive_2015}) in that the latter allows a user to provide more types of measurements. As a consequence of these results, KBD approaches can be applied to all systems amenable to MBD. So, research on KBD is relevant also to MBD.
The particular consequence of 
this
is that our line of research, which has often considered the KBD formulation, is, without any restrictions, generally applicable to the model-based diagnosis problem (cf.\ Sec.~\ref{sec:application_areas}).

\subsubsection*{Publication: Meta Information}

\begin{tabular}{@{}lp{9.5cm}}
	\emph{Publication included in Thesis:} & \cite{rodler_reducing_2017} \\
	\emph{Authors:} & Patrick Rodler, Konstantin Schekotihin 
	\\
	\emph{Title:} & Reducing model-based diagnosis to knowledge base debugging 
	\\
	\emph{Year:} & 2017
	\\
	\emph{Publication Venue (Type):} & Int'l Workshop on Principles of Diagnosis \quad (Conference)
	\\
	\emph{Venue Metrics:} & C
	\\
	\emph{Own Contributions (est.):} & 
	Concept/Idea: $30\,\%$ \newline
	Theory/Algorithms: $70\,\%$ \newline
	Development/Implementation: n.a. \newline
	Evaluation: n.a. \newline
	Manuscript: $95\,\%$ 
	\\
	\emph{Other Publications on the Topic:} & None
\end{tabular}

\newpage\null\thispagestyle{empty}\newpage

\newcommand{\etalchar}[1]{$^{#1}$}

\newpage\null\thispagestyle{empty}\newpage

\appendix

\chapter{Papers Included in the Habilitation Thesis}
\label{apx:papers_incl_in_thesis}
Please find in the attachment the papers included in this thesis. The order of the papers is the same as adhered to in the discussion of the papers in Chap.~\ref{chap:research_details}.\\

\noindent\textcolor{red}{Note: For copyright reasons, the papers are omitted in this version of the thesis. To access these works, please refer to their original sources as listed below and in the bibliography. \\
}

\noindent The included works are the following:
\begin{description}
	\item[\cite{rodler_memory-limited_2022}] Patrick Rodler.
	Memory-limited model-based diagnosis.
	\emph{Artificial Intelligence},
	305:103681, 2022.
	\item[\cite{rodler_dynamichs_2022}] Patrick Rodler. DynamicHS: Streamlining Reiter’s hitting-set tree for sequential diagnosis. \emph{Information Sciences}, 627:251--279, 2022. 
	\item[\cite{rodler_reuse_2020}] Patrick Rodler. Reuse, Reduce and Recycle: Optimizing Reiter’s HS-Tree for Sequential
	Diagnosis. In \emph{European Conference on Artificial Intelligence}, pages 873–880, 2020.
	\item[\cite{rodler_statichs_2018}] Patrick Rodler and Manuel Herold. StaticHS: A Variant of Reiter’s Hitting Set Tree for
	Efficient Sequential Diagnosis. In \emph{Int’l Symposium on Combinatorial Search}, 2018.
	\item[\cite{rodler_random_2022}] Patrick Rodler. Random vs. Best-First: Impact of Sampling Strategies on Decision Making
	in Model-Based Diagnosis. In \emph{AAAI Conference on Artificial Intelligence}, 2022.
	\item[\cite{rodler_how_2022}] Patrick Rodler. How should I compute my candidates? A taxonomy and classification of
	diagnosis computation algorithms. In \emph{Int’l Workshop on Principles of Diagnosis}, 2022.
	\item[\cite{rodler_active_2017}] Patrick Rodler. On Active Learning Strategies for Sequential Diagnosis. In \emph{Int’l Workshop
	on Principles of Diagnosis}, pages 264–283, 2017.
	\item[\cite{rodler_evaluating_2018}] Patrick Rodler and Wolfgang Schmid. On the Impact and Proper Use of Heuristics in Test-Driven Ontology Debugging. In \emph{Int’l Joint Conference Rules and Reasoning}, pages 164–
	184, 2018.
	\item[\cite{rodler_sequential_2022}] Patrick Rodler. Sequential Diagnosis by Systematic Search. \emph{Artificial Intelligence (under
	revision)}, 2022.
	\item[\cite{rodler_are_2019-1}] Patrick Rodler, Dietmar Jannach, Konstantin Schekotihin, and Philipp Fleiss. Are query-
	based ontology debuggers really helping knowledge engineers? \emph{Knowledge-Based Systems},
	179:92–107, 2019.
	\item[\cite{rodler_one_2022}] Patrick Rodler. One step at a time: An efficient approach to query-based ontology debugging.\emph{ Knowledge-Based Systems}, 251:108987, 2022.
	\item[\cite{schekotihin_ontodebug_2018}] Konstantin Schekotihin, Patrick Rodler, and Wolfgang Schmid. OntoDebug: Interactive Ontology Debugging Plug-in for Protégé. In \emph{Int’l Symposium on Foundations of Information and Knowledge Systems}, pages 340–359, 2018.
	\item[\cite{rodler_randomized_2021}] Patrick Rodler, Erich Teppan, and Dietmar Jannach. Randomized Problem-Relaxation Solving for Over-Constrained Schedules. In \emph{Int’l Conference on Principles of Knowledge Representation and Reasoning}, pages 696–701, 2021.
	\item[\cite{rodler_formal_2022}] Patrick Rodler. A formal proof and simple explanation of the QuickXplain algorithm. \emph{Artificial Intelligence Review}, 55(8):6185–6206, 2022.
	\item[\cite{hofer_modeling_2022}] Birgit Hofer, Dietmar Jannach, Julia Nica, Patrick Rodler, and Franz Wotawa. On Modeling
	Techniques for Spreadsheet Debugging: A Theoretical and Empirical Analysis.\emph{ Artificial
	Intelligence (under review)}, 2022.
	\item[\cite{rodler_reducing_2017}] Patrick Rodler and Konstantin Schekotihin. Reducing Model-Based Diagnosis to Knowledge Base Debugging. In \emph{Int’l Workshop on Principles of Diagnosis}, pages 284–296, 2017.
\end{description}


\end{document}